\documentclass{article} 
\usepackage[accepted]{icml2023}

\usepackage{hyperref}
\usepackage{url}
\usepackage{graphics}
\usepackage{graphicx}
\usepackage{booktabs}  
\usepackage{amsmath}
\usepackage{amsfonts}
\usepackage{amsthm}
\usepackage{amssymb}
\usepackage{amsthm}
\usepackage{mathtools}
\usepackage{bm}
\usepackage{dsfont}
\usepackage{enumitem}
\usepackage{hyperref}
\usepackage{nicefrac}
\usepackage{multirow}
\usepackage{subcaption}
\usepackage{amsmath,amsfonts,bm}

\DeclareMathOperator*{\argmin}{arg\,min}

\newcommand{\R}{\mathbb{R}}

\def\floor#1{\lfloor #1 \rfloor}
\def\1{\bm{1}}

\newcommand{\objective}[1]{f(#1)}
\newcommand{\quantilemodel}[1]{\hat{q}_{#1}}
\newcommand{\quantileloss}{\mathcal{L}}

\newcommand{\numberobservation}[1]{{n_{#1}}}

\newcommand{\confidenceinterval}[1]{\mathcal{C}_{#1}}
\newcommand{\conformalcorrection}[1]{\gamma_{#1}}

\newcommand{\numseed}{30}
\newcommand{\numsteps}{50}
\newcommand{\sample}[1]{\tilde{#1}}

\newcommand{\numquantiles}{m}

\newcommand{\numevaluations}{n}
\newcommand{\numcandidates}{N}
\newcommand{\N}{\mathcal{N}}
\newcommand{\configspace}{\mathcal{X}}

\newcommand{\observations}{\mathcal{D}}
\newcommand{\trainset}{\observations_\text{train}}
\newcommand{\valset}{\observations_\text{val}}
\newcommand{\hp}{x}
\newcommand{\fidelity}{r}
\newcommand{\minresource}{r_{\text{min}}}
\newcommand{\maxresource}{r_{\text{max}}}
\newcommand{\redfactor}{\eta}

\newcommand{\RS}{RS}
\newcommand{\TPE}{TPE}
\newcommand{\GP}{GP}
\newcommand{\BORE}{BORE}
\newcommand{\REA}{REA}
\newcommand{\BOHB}{BOHB}
\newcommand{\ASHA}{ASHA}
\newcommand{\MOB}{MOB}
\newcommand{\HT}{HT}

\newcommand{\QR}{QR}
\newcommand{\CQR}{CQR}

\newcommand{\QRMF}{QR+MF}
\newcommand{\CQRMF}{CQR+MF}

\newcommand{\ST}{Syne Tune}

\newcommand{\FCNet}{FCNet}
\newcommand{\NASBench}{NAS201}
\newcommand{\NASSurr}{NAS301}
\newcommand{\LCBench}{LCBench}

\icmltitlerunning{Optimizing Hyperparameters with Conformal Quantile Regression}

\begin{document}

\twocolumn[
\icmltitle{Optimizing Hyperparameters with Conformal Quantile Regression}



\icmlsetsymbol{equal}{*}

\begin{icmlauthorlist}
\icmlauthor{David Salinas}{AWS}
\icmlauthor{Jacek Golebiowski}{AWS}
\icmlauthor{Aaron Klein}{AWS}
\icmlauthor{Matthias Seeger}{AWS}
\icmlauthor{Cedric Archambeau}{AWS}
\end{icmlauthorlist}

\icmlaffiliation{AWS}{Amazon Web Services}
\icmlcorrespondingauthor{David Salinas}{david.salinas.pro@gmail.com}

\icmlkeywords{Machine Learning, ICML}

\vskip 0.3in
]



\printAffiliationsAndNotice{\icmlEqualContribution} 

\begin{abstract}
Many state-of-the-art hyperparameter optimization (HPO) algorithms rely on model-based optimizers that learn surrogate models of the target function to guide the search. Gaussian processes are the de facto surrogate model due to their ability to capture uncertainty but they make strong assumptions about the observation noise, which might not be warranted in practice. In this work, we propose to leverage conformalized quantile regression which makes minimal assumptions about the observation noise and, as a result, models the target function in a more realistic and robust fashion which translates to quicker HPO convergence on empirical benchmarks. To apply our method in a multi-fidelity setting, we propose a simple, yet effective, technique that aggregates observed results across different resource levels and outperforms conventional methods across many empirical tasks.
\end{abstract}

\section{Introduction}

Hyperparameters play a vital role in the machine learning (ML) workflow. They control the speed of the optimization process (e.g., learning rate), the 
capacity of the underlying statistical model (e.g., number of units) or the generalization quality through regularization (e.g., weight decay). While virtually all ML pipelines benefit from having their hyperparameters tuned, this can be tedious and expensive to do, and practitioners tend to leave many of them at their default values.

Hyperparameter optimization~\citep{feurer-automlbook18a} is a powerful framework to tune hyperparameters automatically with a large body of work to tackle this optimization problem, spanning from simple heuristics to more complex model-based methods. Random search-based approaches~\citep{bergstra-jmlr12a} define a uniform distribution over the configuration space and repeatedly sample new configurations until an total budget is exhausted. Evolutionary algorithms modify a population of hyperparameter configurations by hand-designed mutations. Finally, model-based methods, such as Bayesian optimization~\citep{snoek-nips12a} 
use the collected data points to fit a surrogate model of the target objective, which informs the sampling distribution so that more promising configurations are selected over time.

Unfortunately, solving the HPO problem can become computationally expensive, as each function evaluation incurs the cost of fully training and validating the underlying machine learning model.
Multi-fidelity HPO~\citep{karnin13,li-iclr17} accelerates this process by terminating the evaluation of under-performing configurations early and only training promising configurations to the end. Early stopping allows algorithms to explore more configurations within the total search budget. Baseline multi-fidelity methods can be made more efficient by relying on model-based sampling, exploiting past observations to select more promising configurations over time.

Model-based multi-fidelity methods constitute the state-of-the-art in HPO today, but still face some limitations. Concretely, we identify two major gaps with the current approaches. 
First, Bayesian optimization  
usually assume that the observation noise is homoskedastic and distributed as a Gaussian. This allows for an analytic expression of the likelihood for Gaussian processes, the most popular probabilistic model for Bayesian optimization~\citep{snoek-nips12a}.  However, most HPO problems 
 exhibit heteroskedastic noise that can span several orders of magnitude \cite{salinas20a,cowen2020hebo}. For instance, the validation error of a model trained with SGD can behave very sporadically for large leaning rate and not taking this heteroskedasticity into account can severely hinder the performance of methods with Gaussian assumptions. 
 
Second, it is difficult to model probabilistically the target metric both across configurations and resources (e.g., number of epochs trained), while at the same time retaining the simplicity of Gaussian processes. Previous work maintains separate models of the target at different resource levels~\citep{Falkner:18, Yang2022}, or use a single joint model~\citep{Klein:mobster, Swersky:14}. The former does not take into account dependencies across resource levels, even though these clearly exist. The latter has to account for the non-stationarity of the 
target function, and requires strong modeling assumptions in order to remain tractable.

With this paper, we propose a conformalized quantile regression surrogate model that can be used in the presence of any noise, possibly non Gaussian or heteroskedastic. We further propose a new way to expand any model-based single-fidelity HPO method to the multi-fidelity setting, by aggregating observed results across different resource levels. This strategy can be used with most single-fidelity method and greatly simplifies the model based multi-fidelity setup. We show that, in spite of its simplicity, our framework offers competitive results across most common single-fidelity methods and significant improvements over baselines when paired with conformalized quantile regression. Our main contributions are the following:

\begin{itemize}
    \item We introduce a novel surrogate method for HPO based on conformalized quantile regression which can handle heteroskedastic and non-Gaussian distributed noise.
    \item We propose a simple way to extend any single-fidelity method into a multi-fidelity method, by using only the last observed datapoint for each hyperparameter configuration to train the function surrogate. 
    \item We run empirical evaluations on a large set of benchmarks, demonstrating that quantile regression surrogates achieve a more robust performance compared to state-of-the-art methods in the single-fidelity case
    \item We show that our new multi-fidelity framework outperforms state-of-the-art methods across multiple single-fidelity surrogates.
\end{itemize}

The paper first reviews related work and then discuss our proposed method for single-fidelity optimization leveraging conformal quantile prediction. We then describe an extension to the multi-fidelity case, before evaluating the method on an extensive set of benchmarks.

\section{Related work}

Bayesian optimization is one of the most successful strategies for hyperparameter optimization (HPO) \citep{shahriari-ieee16a}. 
Based on a probabilistic model of the objective function, it iteratively samples new candidates by optimizing an acquisition function, that balances between exploration and exploitation when searching the space.
Typically, Gaussian processes are used as the probabilistic surrogate model
~\citep{snoek-nips12a}, but other methods, such as random forests~\citep{hutter-lion11a} or Bayesian neural networks~\citep{springenberg-nips16,snoek-icml15} are possible. 
Alternatively, instead of modeling the objective function, previous work~\citep{bergstra-nips11a,tiao2021bore} estimate the acquisition function directly by the density ratio of well and poorly performing configurations.

Despite its sample efficiency, Bayesian optimization still requires tens to hundreds of function evaluations to converge to well-performing configurations. To accelerate the optimization process, multi-fidelity optimization exploits cheap-to-evaluate fidelities of the objective function such as training epochs \citep{Swersky:14}.  \citet{jamieson-aistats16} proposed to use successive halving~\citep{karnin13} for multi-fidelity hyperparameter optimization which trains a set of configurations for a small budget and then only let the the top half configurations continue for twice as many resources. Hyperband~\citep{li-iclr17} calls successive halving as a sub-routine with varying minimum resources level, to avoid that configurations are terminated too early.
\citet{Falkner:18} combined Hyperband with Bayesian optimization to replace the inefficient random sampling of configuration by Bayesian optimization with kernel density estimators~\citep{bergstra-nips11a}. ASHA \citep{Li:19} proposed to extend Hyperband to the asynchronous case when using multiple workers which led to significant improvements and \citet{Klein:mobster,Yang2022} later combined this method with Gaussian process based Bayesian optimization. Instead of relying on a model-based approach, \citet{awad-ijcai21} instead proposed to combine Hyperband with evolution algorithms.

An orthogonal line of work models the learning curves of machine learning algorithms directly; see \citet{mohr-arxiv22} for an overview.
Previous work by \citet{domhan:15} fits an ensemble of parametric basis functions to the learning curve of a neural network. This method can be plugged into any HPO approach such that evaluation of an network is stopped if it is unlikely to outperform previous configurations and the prediction of the model is returned to the optimizer.
\citet{klein-iclr17} used a Bayesian neural networks to predict the parameters of these basis functions which is able to model the correlation across hyperparameter configurations.
\citet{wistuba20a} proposed neural networks architectures that ranks learning curves across different tasks.

To avoid requiring Gaussian homoskedastic noise, several papers considered the use of quantile regression for HPO \cite{Picheny13,salinas20a,moriconi2020high} but those approaches do not ensure that the provided uncertainties are well calibrated. Conformal prediction has been gaining traction recently in ML applications due to its ability to provide well calibrated uncertainty with widely applicable assumptions, in particular not requiring the presence of a Gaussian homoskedastic distribution \cite{Shafer07}. To the best of our knowledge, conformal prediction has only been considered for single-fidelity HPO by \citep{Stanton22} and \citet{doyle-arxiv22}. The former applies conformal correction to standard GP posteriors in order to improve model calibration on non-Gaussian noise, whereas we build our method on quantile regression which is already robust to non-Gaussian and heteroskedastic noise. The latter conducted a preliminary study showing that conformal predictors can outperform random-search on four datasets. The key difference to our method is that we utilize the framework of conformal quantile prediction from \cite{romano} which leverages quantile regression allowing to bypass the need to fit an additional model for the variance. In both cases, our work differs as we consider the asynchronous multi-fidelity case which allows the method to perform much better in presence of hundreds of observations.

\section{Single-fidelity Hyperparameter Optimization}

In the single-fidelity hyperparameter optimization setting, we are interested in finding the hyperparameter minimizing of a blackbox function $f$:
$$x^*=\argmin_{\hp \in \configspace} \objective{\hp}$$ where $\objective{\hp}\in \configspace$ denotes the validation error obtained for a hyperparameter configuration $x$. Hyperparameters may include the learning rate, number of layers and number of hidden dimensions of a transformer or a convolutional neural network. Given that evaluating $f$ is typically expensive and gradients are not readily available, we look for gradient-free and sample efficient optimization methods.

Bayesian Optimization is arguably one of the most popular approaches owing to its ability to efficiently trade-off exploration and exploitation when searching the configuration space. In each iteration $\numevaluations$, a probabilistic model of the objective function $f$ is fitted on the $\numevaluations$ previous observations $\observations = \{(\hp_1, y_1), \dots , (\hp_{\numevaluations}, y_{\numevaluations})\}$; the first initial configurations are typically drawn at random. To select the next point to evaluate, an acquisition function is then optimized to select the most promising candidate based on the probabilistic surrogate, for instance by picking the configuration that maximizes the expected improvement \cite{jones1998efficient}.

The standard approach to model the objective function uses Gaussian Processes due its computational tractability and theoretical guarantees ~\citep{Srinivas2012}.
However, this approach assumes that the observation noise is Gaussian and homoskedastic. We next describe an approach to lift this restriction by leveraging quantile regression with conformal prediction as a probabilistic model for $f$.

\section{Conformal Quantile Regression}

\paragraph{Preliminaries.} 

For a real-valued random variable $Y$, we denote by $g(y)$ its probability density function and by $F_Y(y) = \int_{-\infty}^y g(t) dt$ its cumulative distribution function (CDF). The associated quantile function of $Y$ is then defined as follows:
$$F_Y^{-1}(\alpha) = \inf_{y \in \R}\{ F_Y(y) \geq \alpha\}.$$

The quantile function allows to easily obtain confidence intervals. 
One can also easily sample from the distribution by first sampling a random quantile uniformly $\alpha \sim \mathcal{U}([0, 1])$ and then computing $y = F_Y^{-1}(\alpha)$ which provides one sample $y$ from the distribution $Y$.

\paragraph{Quantile regression.}

Given data drawn from a joint distribution $(x, y)\sim F_{(X, Y)}$, quantile regression aims to estimate a given quantile $\alpha$ of the conditional distribution of $Y$ given $X=x$, e.g. to learn the function 
$$q_\alpha (x) = F^{-1}_{Y|X=x}(\alpha)$$ which predicts the quantile $\alpha$ conditioned on $x$. This problem can be solved by minimizing the quantile loss function \cite{1982} for a given quantile $\alpha$ and some predicted value $\hat{y}$:

\begin{equation}
\quantileloss_\alpha(y, \hat{y}):= \begin{cases}\alpha(y-\hat{y}) & \text { if } y-\hat{y}>0, \\ (1-\alpha)(\hat{y}-y) & \text { otherwise. }\end{cases}
\label{eq:quantile_loss}
\end{equation}

A critical property is that minimizing the quantile loss allows to retrieve the desired quantile in the sense that
$$\argmin_{\hat{y}} \mathbb{E}_{y\sim F_Y}[
\quantileloss_\alpha(y, \hat{y})] = F_Y^{-1}(\alpha).$$

Given a set of $\numevaluations$ observations 
$\observations = \{(x_i, y_i)\}_{i=1}^\numevaluations$, one can thus estimate the quantile function by training a model $\quantilemodel{\alpha}$ with parameters $\theta$ 
that minimizes the quantile loss given by Eq. \ref{eq:quantile_loss}:
\begin{equation}
 \theta^* = \argmin_\theta \frac{1}{\numevaluations} \sum_{i=1}^\numevaluations
\quantileloss_\alpha(y_i, \quantilemodel{\alpha}(x_i)).
\label{eq:trainining-quantile}
\end{equation}
\begin{figure*}
\caption{Samples from the synthetic function $\objective{\hp} \sim \N(0, \rho(x)^2)$  to be minimized (left), illustration of the Thompson sampling procedure based on $\numquantiles=8$ predicted quantiles (middle) and acquisition function obtained for our method and a GP (right). When sampling, we sample one random quantile for each candidate and pick the best configuration obtained.
\label{fig:illustration}}
\begin{center}
\includegraphics[width=0.99\textwidth]{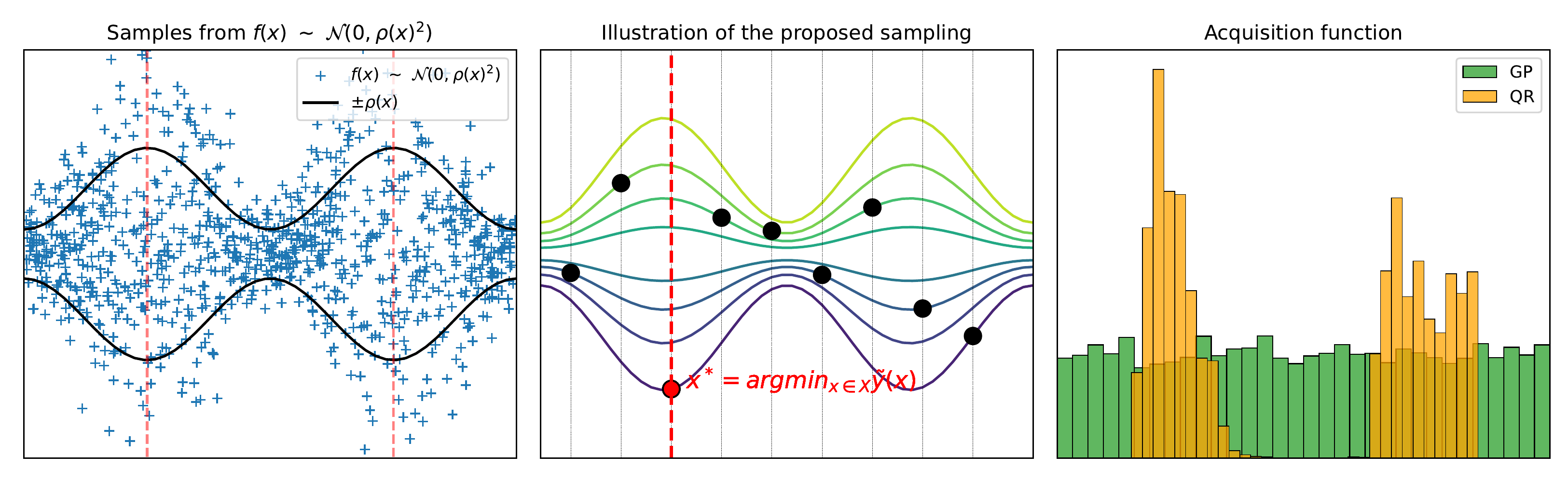}
\end{center}
\end{figure*}

\paragraph{Quantile Regression Surrogate.}

We now explain how we can leverage quantile regression to build a probabilistic surrogate for Bayesian Optimization. 

To this end, we estimate $\numquantiles$ models by minimizing Eq. \ref{eq:trainining-quantile} for equally spread-out quantiles $\{\alpha_1, \dots, \alpha_\numquantiles\}$ where $\alpha_j = j / (\numquantiles + 1)$ and where $m>1$ is an even number.
This allows to provide probabilistic predictions conditioned on any hyperparameter $x$. We then use independent Thompson sampling as the acquisition function by taking advantage that samples can easily be obtained through predicted quantiles. Indeed, one can then draw a sample $\sample{y}(x)$ from the estimated conditional distribution of $F_{Y|X}$ by simply sampling a quantile at random among the $\numquantiles$ quantiles predicted by the model $\{
\quantilemodel{\alpha_1}(x), \dots, \quantilemodel{\alpha_\numquantiles}(x)\}$.\footnote{One downside typically associated with fitting $\numquantiles$ models is that the quantiles predicted may not be monotonic \cite{gasthaus19a}, however this problem does not occur in our case since we simply use 
samples from the predicted distribution.}

To select the next configuration to evaluate, we then use independent Thompson sampling as the acquisition function. We first sample a set of $\numcandidates$ candidates $\sample{\configspace} = \{\sample{\hp}_1, \dots, \sample{\hp}_\numcandidates\}$ uniformly at random and then sample a validation performance for each of those candidates to obtain $\{\sample{y}(\sample{\hp}_1), \dots, \sample{y}(\sample{\hp}_\numcandidates)\}$. We then return the configuration that has the lowest sampled value 
$$x^* = \argmin_{x\in \{\sample{\hp}_1, \dots, \sample{\hp}_\numcandidates\}} \sample{y}(x).$$

We illustrate this procedure in Figure~\ref{fig:illustration} which shows how different quantiles are sampled in a toy example and how the next point is selected by picking the configuration with the lowest sampled predicted value. In this example, we consider $\objective{\hp} \sim \N(0, \rho(x)^2)$ as the function to minimize with $\rho(x) = \sin(x)^2 + \kappa$ where $\kappa$ is set to $\kappa=0.3$ to ensure positive variance. Samples of this function are shown in Figure~\ref{fig:illustration}, left. For this function, a standard \GP{} is not able to represent the heteroskedastic variance and as such cannot favor any part of the space as shown by its uniformly distributed acquisition. However, while the mean of the function is always zero, the optimal points to sample are both situated at $\pi/2$ and $3 \pi / 2$ where the uncertainty is the highest. While, the GP cannot model this information given its homoskedastic noise, quantile regression is able to regress the conditional quantiles and therefore correctly identify the best regions to sample as evidenced by the acquisition function which peaks at the two best values for the next candidate.

\paragraph{Conformalizing predictions.}  While quantile regression can learn the shape of any continuous distribution given enough data, the model predictions are not guaranteed to be well calibrated given insufficient data. 

More precisely, the quantiles estimated allows us to construct $\numquantiles/2$ confidence intervals $$\confidenceinterval{j}(x) = \left[\quantilemodel{\alpha_j}(x), \quantilemodel{1 - \alpha_j}(x)\right]$$ for $j\leq \numquantiles / 2$.\footnote{Note that the values chosen for the quantiles $\alpha_j = j / (\numquantiles + 1)$ ensures that the quantile $1 - \alpha_j$ belongs to the $\numquantiles$ quantiles computed since $1 - \alpha_j = \alpha_{\numquantiles - j}$.}
For each confidence interval, we would like to have a miscoverage rate $2 \alpha_j$, i.e. the predictions should have probability at least $1 - 2 \alpha_j$ of being in the confidence interval $\confidenceinterval{j}(x)$,
\begin{equation}
\mathbb{P}[Y \in \confidenceinterval{j}(x)] = 1 - 2 \alpha_j.
\label{eq:coverage-rate}
\end{equation}

In the presence of heteroskedasticity, this requires to have the length of $\confidenceinterval{j}(x)$ to depend on $x$ which is possible with the use of quantile regression as illustrated in Figure~\ref{fig:illustration}. However, the coverage statement of Eq. \ref{eq:coverage-rate} cannot be guaranteed when fitting models on finite sample size. Miscalibrated intervals can be problematic for HPO, as it may lead to a model converging early to a suboptimal solution. To address this problem, we propose using the split conformal method from \citet{romano} that allows to obtain robust coverage for each $\numquantiles / 2$ of the predicted confidence intervals that we now describe.

The method consists in applying an offset on each confidence interval, which is estimated on a validation set. We divide the dataset of available observations $\observations = \{(x_i, y_i)\}_{i=1}^\numevaluations$ into a training set $\trainset$ and a validation set $\valset$. After fitting each of the $\numquantiles$ quantile regression models $\quantilemodel{\alpha_j}$ on the training set $\trainset$, we compute conformity scores that measure how well the predicted conformal intervals fit the data for each sample in the validation set:
\begin{equation}
E_j = \left\{ 
\text{max}(\quantilemodel{\alpha_j}(x_i) - y_i, y_i - \quantilemodel{1-\alpha_j}(x_i))
\right\}_{i=1}^{|\valset|}.
\label{eq:conformity-scores}
\end{equation}

The intuition of the conformity scores is the following. First note that the sign of the score is positive when the target $y_i$ is outside of the target and negative when the target falls inside the predicted interval. This allows the score to account for both overcoverage and undercoverage cases as we want to reduce the interval in cases of overcoverage and increase it in case of undercoverage.
In addition, the score amplitude always measures the distance to the closest quantile of the confidence interval, i.e. the score amplitude of each sample is $|y_i - q_i|$ where $q_i$ is the closest quantile from $y_i$ between $\quantilemodel{\alpha_j}(x_i)$ and $\quantilemodel{1-\alpha_j}(x_i)$.

Given this score we compute a correction $\conformalcorrection{j}$ which is set to 

\begin{equation}
\conformalcorrection{j}=(1-2\alpha_j)\left(1 + \frac{1}{|\valset|}\right)\text{-th empirical quantile of }E_j.
\label{eq:gamma-correction}
\end{equation}

The conformalized prediction interval for a new data point $(x, y)$ is then given by
\begin{equation}
\hat{\confidenceinterval{j}}(x) = \left[\quantilemodel{\alpha_j} (x) - \conformalcorrection{j}, \quantilemodel{1 - \alpha_j} (x)+ \conformalcorrection{j}\right].
\label{eq:cp-correction}
\end{equation}

An important property of this procedure is that the corrected confidence intervals are guaranteed to have valid coverage, e.g. the probability that $y$ belongs the prediction interval $\hat{\confidenceinterval{j}}$ can then be shown to arbitrarily close to $1-2\alpha_j$ \cite{romano}

$$1 - 2 \alpha_j \leq \mathbb{P}[Y \in \hat{\confidenceinterval{j}}(x)] \leq  1 - 2 \alpha_j + \frac{1}{|\valset|+1}.$$

Once the confidence intervals are readjusted by offsetting quantiles, we can sample new candidates to evaluate using a protocol based on Thompson sampling discussed in the previous section while being able to guarantee coverage properties of our predicted quantiles.
The pseudo-code of the proposed algorithm to select the next configuration to evaluate is given in Algo. \ref{alg:cqr} where the key three steps are 1) fitting quantile regression models, 2) computing quantile adjustments and 3) sampling the best candidate with independent Thompson sampling.

\section{Multi-fidelity and Successful Halving}
\label{sec:multifidelity}

Single-fidelity methods steer the search towards the most promising part of the configuration space based on the observed validation performance of hyperparameter configurations, however, this does not consider leveraging other available signals, such as the loss emitted at each epoch for a neural network trained with SGD. Multi-fidelity methods consider this additional information to further accelerate the search by cutting poor configurations preemptively.

Formally, multi-fidelity optimization considers the following optimization problem:
$$x^*=\argmin_{\hp \in \configspace} \objective{\hp, \maxresource{}}$$ 
where $\objective{\hp,\maxresource{}}$ denotes the blackbox error obtained for a hyperparameter $x$ at the maximum budget $\maxresource{}$ (for instance the maximum number of epochs) and we assume that $\fidelity \in [\minresource, \maxresource]$. Typically, early values of $\objective{\hp, \fidelity}$ for $\fidelity<\maxresource{}$ are informative of $\objective{\hp, \maxresource}$, while being computationally cheaper to collect and can help us to cut poorly performing configurations.

\paragraph{Asynchronous Successive Halving.}
Asynchronous Successive Halving (ASHA) \cite{Li:19} is a multi-fidelity method that can leverage several workers to evaluate multiple configurations asynchronously while stopping poor configurations early. The method starts by evaluating a list of random configurations in parallel for a small initial budget. When a result is collected from a worker, it is continued or stopped based on its result - the evaluation of the configuration continues if it is in the top results seen so far for a given fidelity and interrupted otherwise. Stopped configurations are replaced with new candidates sampled randomly and the process is iterated until the tuning budget is exhausted.

ASHA avoids synchronization points by evaluating each configuration based on the data available at the time, which can lead to false positives in the continuation decision. Indeed, some configurations may be continued due to poor competition rather than good performance and would have been stopped if more data was available. However, the avoidance of synchronization points efficiently deals with straggler evaluation and is one of the key components of the method’s excellent performance in practice. The pseudo-code of the method is given in the appendix.

\begin{algorithm}[tb]
   \caption{{\tt CQR} candidate suggestion pseudo-code. \label{alg:sampling}}   
\begin{algorithmic}[1]
   \FUNCTION{{\tt SUGGEST}()}{}{}
   \STATE {\bfseries Input:} configuration space $\configspace$, set of observations $\observations$, number of quantiles $\numquantiles$, number of candidates $\numcandidates$  
   \STATE {\bfseries Output:} next configuration to evaluate
   \STATE{$\trainset, \valset = \text{split\_train\_val}(\observations)$}
   \FOR{$1 \leq j \leq \numquantiles$}
   \STATE Fit model $\quantilemodel{\alpha_j}$ on $\trainset$ with Eq. \ref{eq:trainining-quantile}
   \ENDFOR
   \FOR{$1 \leq j \leq \numquantiles / 2 - 1$}
   \STATE Compute conformity scores $E_j$ with Eq. \ref{eq:conformity-scores}
   \STATE Compute correction $\gamma_j$ with Eq. \ref{eq:gamma-correction} 
   \ENDFOR   
   \STATE $\sample{\configspace}$ = Sample $\numcandidates$ candidates from $\configspace$
   \FOR{$1 \leq i \leq \numcandidates$}
   \STATE Draw random quantile $j$ in $[1, \numquantiles]$
   \IF{$j < m / 2$}
   \STATE $\tilde{y}_i = \quantilemodel{\alpha_j}(x_i) - \gamma_j$
   \ELSE
   \STATE $\tilde{y}_i = \quantilemodel{\alpha_j}(x_i) + \gamma_{m-j}$
   \ENDIF
   \ENDFOR
   \STATE $i^* = \argmin_{i} \tilde{y}_i$
   \STATE \textbf{return} $x_{i^*}$
   \ENDFUNCTION{}

\end{algorithmic}
\label{alg:cqr}
\end{algorithm}

\paragraph{Model-based ASHA.}

One pitfall of ASHA is that candidates are sampled randomly at initialization and when workers become free after configurations are interrupted. However after spending some time in the tuning process, we gathered results which we would clearly like to be able to bias the search towards the most promising parts of the configuration space.

One challenge is that most single-fidelity model-based approaches regress a surrogate model $\objective{x, \maxresource{}}$ given observations at the final fidelity $\maxresource{}$. It becomes then difficult to combine model-based and multi-fidelity approaches given that when we stop a poor configuration at a resource $\fidelity < \maxresource$, we are unsure about what would have been the value at $\objective{x, \maxresource{}}$. 

\paragraph{Bridging single and multi-fidelity methods.} We propose a simple data transformation that allows to use any single fidelity method in the multi-fidelity setting. We denote the configurations and evaluations obtained at a given time as 
$$\left\{(\hp_i, \{\objective{\hp_i, \fidelity_1}, \dots, \objective{\hp_i, \fidelity_\numberobservation{\hp_i}}\})\right\}_{i=1}^n$$
where $\numevaluations{}$ denotes the number of configurations evaluated and $\numberobservation{\hp}$ denotes the number of fidelities evaluated for a configuration $\hp$.

We propose to consider the transformation that takes the last value of the time-observations of a given configuration. Namely, we propose to consider the transformation: $z = \objective{\hp, \fidelity_{n_\hp}}$ and then use a single-fidelity method rather than random-search to determine the best next configuration to evaluate given the observations $\observations = \{(\hp_i, z_i)\}_{i=1}^\numevaluations$  while using ASHA for the stopping decisions.

Relying on the last observed value $\objective{\hp, \fidelity_{n_\hp}}$ rather than all fidelities $\objective{x, r}$ for $r \leq \fidelity_{n_\hp}$ significantly simplifies the multi-fidelity setup but obscures a portion of the available signal. However, as evaluating configuration candidates longer is expected to improve their result, the data transformation is effectively pushing the poor and well performing configurations further apart. Assuming configurations are stopped with a probability inversely correlated with their performance and their results would not cross if the training were to continue, the result-based ordering of configurations remains constant regardless of whether we use the last or final observation. This means that it remains possible under those assumptions  to discriminate between promising and not-promising configurations. \footnote{The accentuated spread between bad and good configurations can be mitigated by using quantile normalization as the transformation is invariant to monotonic changes. We do not report results for this approach as it adds a layer of complexity and performed on-par with just taking the last observations in our experiments.} 

In addition to working well with the Conformal Quantile Regression that we introduced, we will also show in our experiments that this simple transformation allows to combine single-fidelity methods with ASHA while reaching the performance of state-of-the-art dedicated model-based multi-fidelity methods.

\section{Experiments}

We evaluate our method against state-of-the-art HPO algorithms on a large collection of real-world datasets on both single and multi-fidelity tasks.
The code to reproduce our results is available at \url{https://github.com/geoalgo/syne-tune/tree/icml_conformal}.

\paragraph{Benchmarks.}
Our experiments rely on 13 tasks coming from \FCNet{} \citep{Klein:19a}, \NASBench{} \citep{Dong:20} and \LCBench{} \citep{zimmer2021} benchmarks as well as \NASSurr{} \cite{nas301} using the implementation provided in \cite{pfisterer2022yahpo}. 
All methods are evaluated asynchronously with 4 workers.
Details on these benchmarks and their configuration spaces distributions are given in Appendix~\ref{sec:experiment-details}.

\paragraph{Baselines.}
For single-fidelity benchmarks, we compare our proposed method (CQR) with random-search (\RS{}) \cite{Bergstra:11}, TPE{} \cite{bergstra-nips11a}, Gaussian Process (\GP{}) \cite{snoek-nips12a}, regularized-evolution (\REA{}) \cite{real2019} and \BORE{} \cite{tiao2021bore}. For multi-fidelity benchmarks, we compare against \ASHA{} \cite{Li:19}, \BOHB{} \cite{Falkner:18}, Hyper-Tune (\HT{}) \cite{Yang2022} and Mobster (\MOB{}) \cite{Klein:mobster}. 

\paragraph{Experiment Setup.}
All tuning experiments run asynchronously with 4 workers and are stopped when $200 * \maxresource$ results were observed, which corresponds to seeing 200 different configurations for single-fidelity methods, or when the wallclock time exceeded a fixed budget. All runs are repeated with \numseed{} different random seeds, and we report mean and standard errors. We use gradient boosted trees \cite{Friedman2001} for the quantile-regression models with the same hyperparameter used for \BORE{}. We use the simulation backend provided by Syne Tune \cite{salinas2022syne} on a {\tt AWS m5.4xlarge} machine to simulate methods which allows to account for both optimizers and blackbox runtimes.

\paragraph{Metrics.}
We benchmark the performance of methods using normalized regret, ranks averaged over tasks and critical diagrams. The normalized regret, also called average distance to the minimum, is defined as $(y_t - y_\text{min}) / ( y_\text{max} - y_\text{min})$ where $y_\text{min}$ and $y_\text{max}$ denotes respectively the best and worse possible values of the blackbox $f$ and $y_t$ denotes the best value found after at a time-step $t$ for a given method. To aggregate across tasks, we report scores at \numsteps{} fractions of the total budget for each tasks. We then average normalized regret over each budget proportion across tasks and seeds. We  compute ranks for each method at all time-steps and also average those values over tasks and seeds. 
Critical diagrams show group of methods that are statistically tied together with a horizontal line using the statistical test proposed by \citep{demsar06a}. They are computed over averaged ranks of the methods obtained for a fixed budget. The performance of all methods per task is also given in the appendix in Fig. \ref{fig:single-fidelity-all-tasks}, \ref{fig:multi-fidelity-all-tasks}, \ref{fig:multi-fidelity-ablation-all-tasks}.

\begin{figure*}[t]
\center
\begin{subfigure}{.65\textwidth}
  \centering
\includegraphics[width=0.5\textwidth]{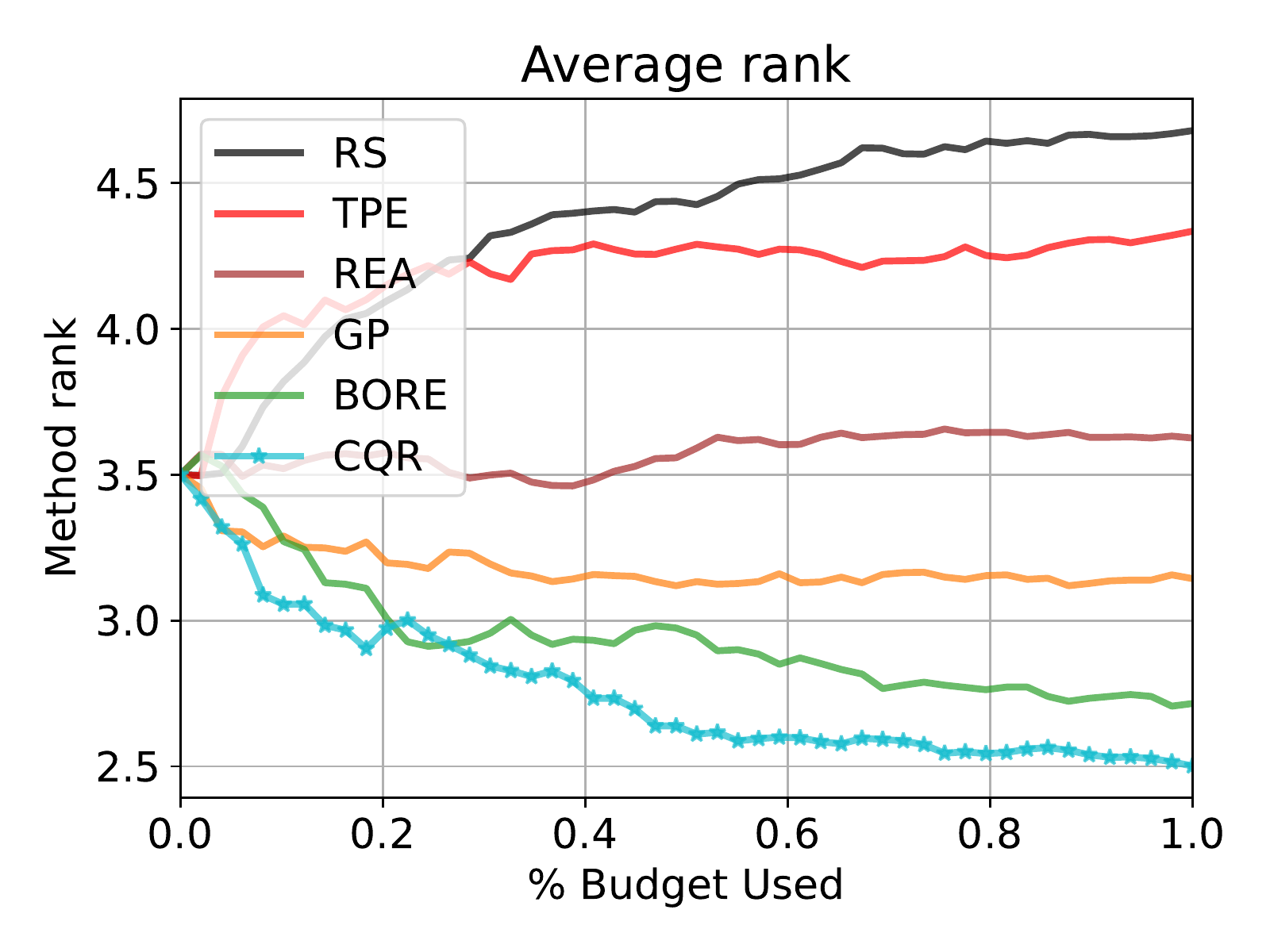}~
\includegraphics[width=0.5\textwidth]{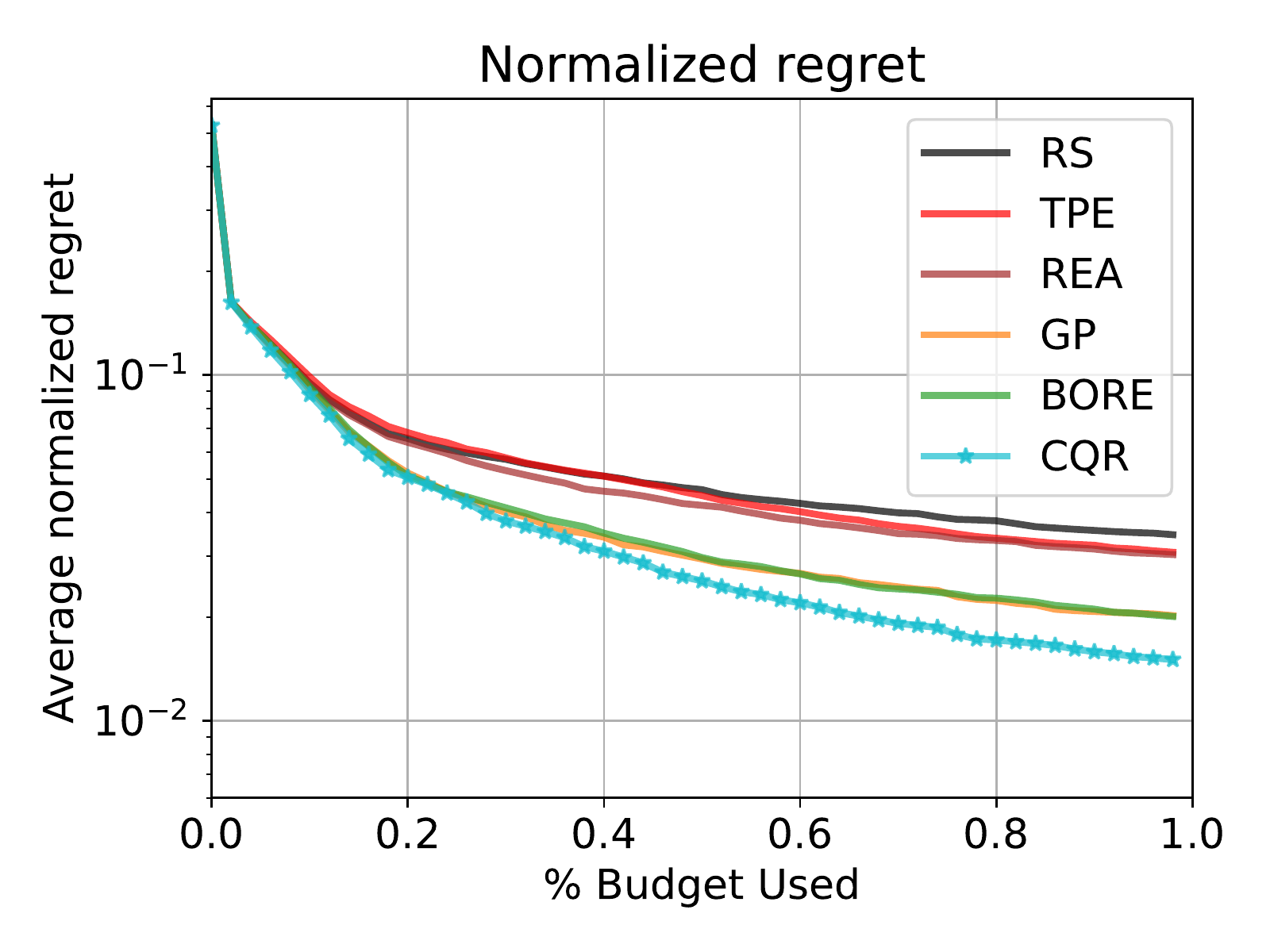} 
\end{subfigure}
\begin{subfigure}{.34\textwidth}
  \centering
\includegraphics[width=0.99\textwidth]{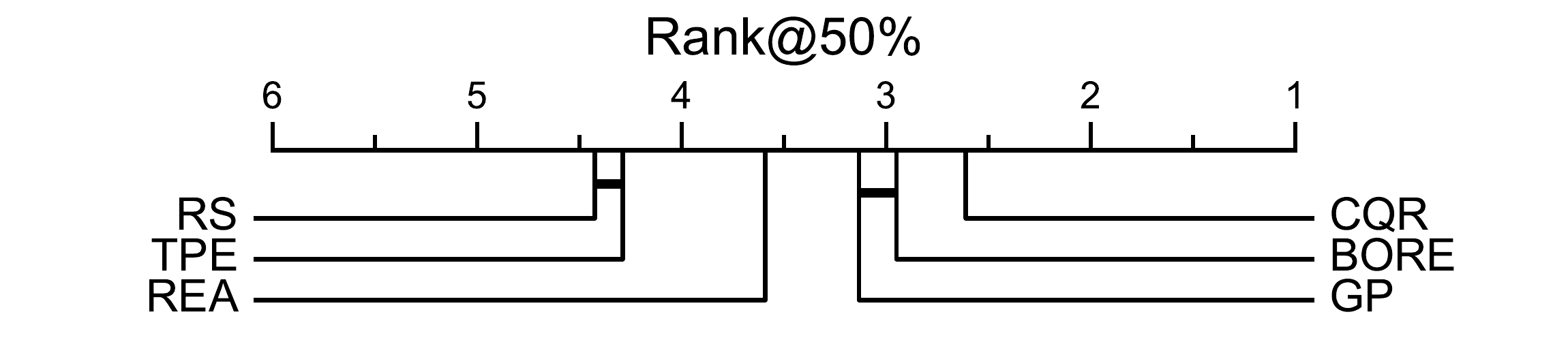}\\
\vspace{0.5cm}
\includegraphics[width=0.99\textwidth]{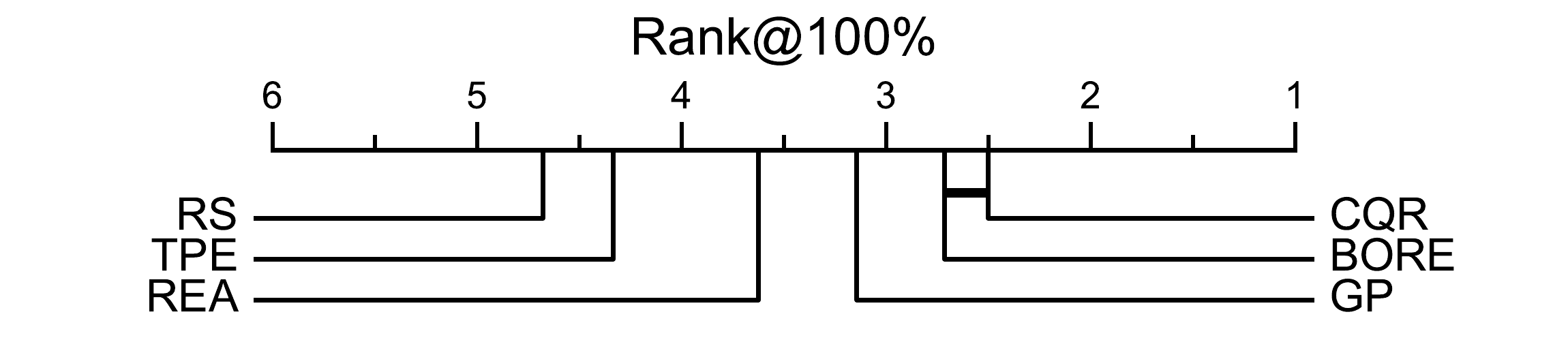}\\

\end{subfigure}
\begin{subfigure}{.65\textwidth}
  \centering
\includegraphics[width=0.5\textwidth]{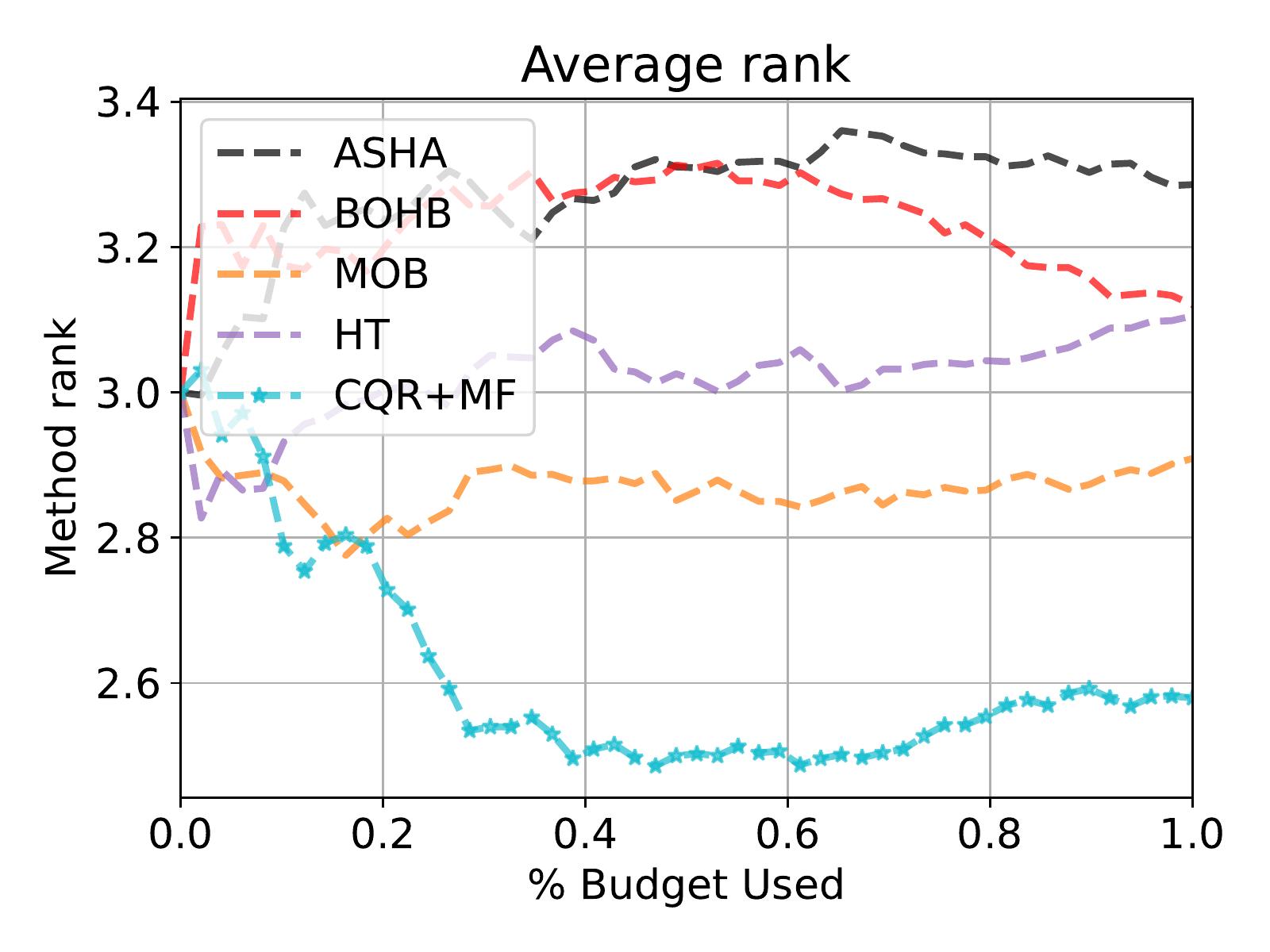}~
\includegraphics[width=0.5\textwidth]{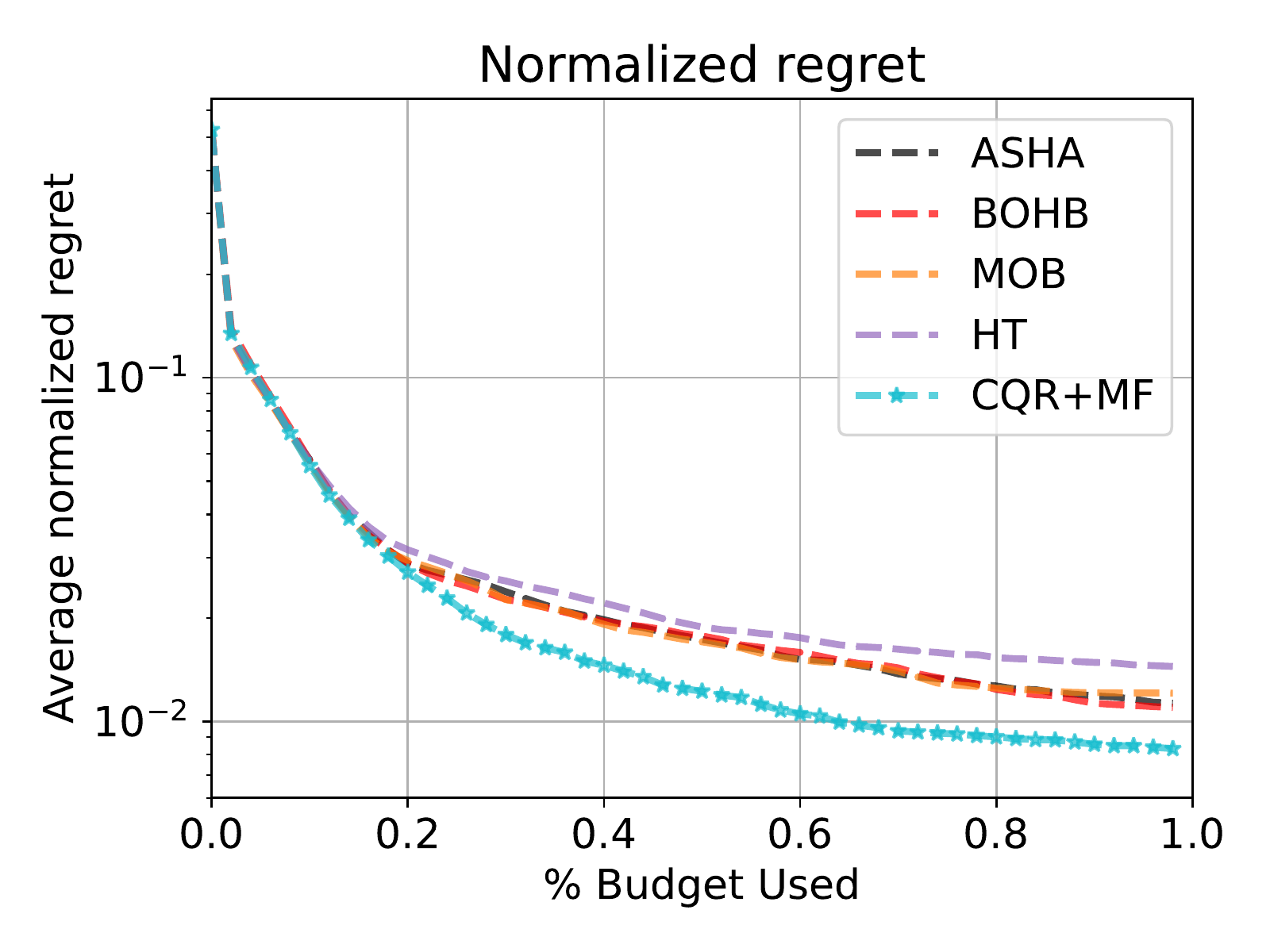} 
\end{subfigure}
\begin{subfigure}{.34\textwidth}
  \centering
\includegraphics[width=0.99\textwidth]{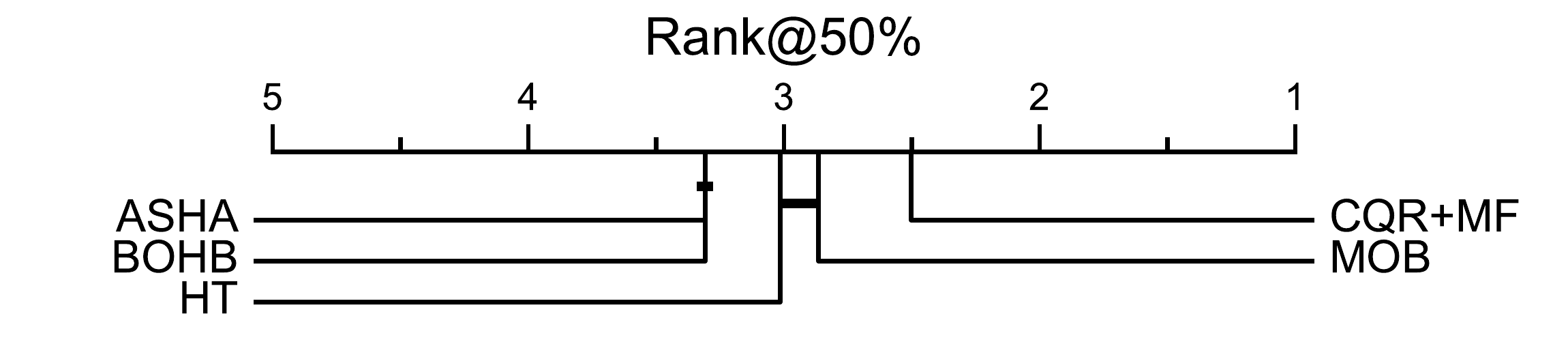}\\
\vspace{0.5cm}
\includegraphics[width=0.99\textwidth]{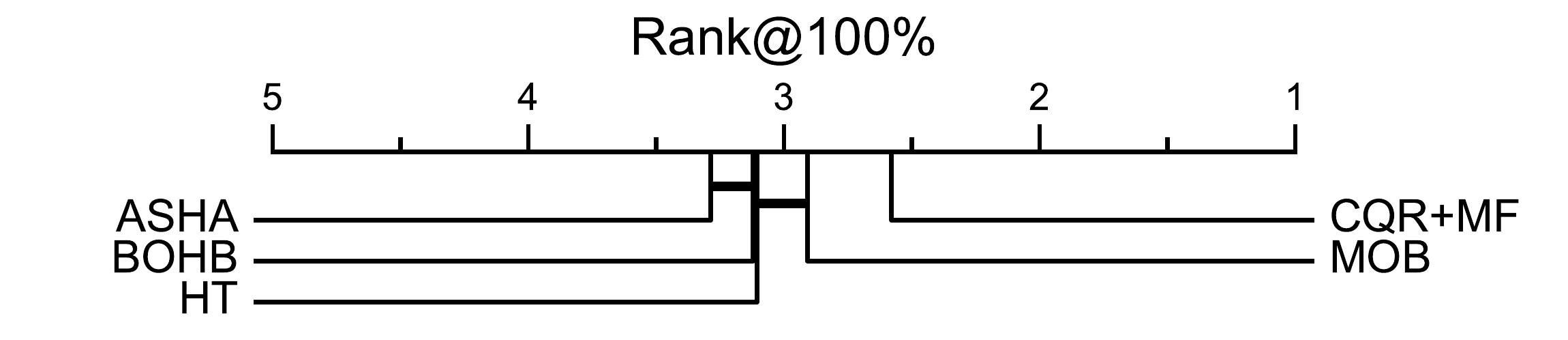}\\
\end{subfigure}

\begin{subfigure}{.65\textwidth}
  \centering
\includegraphics[width=0.5\textwidth]{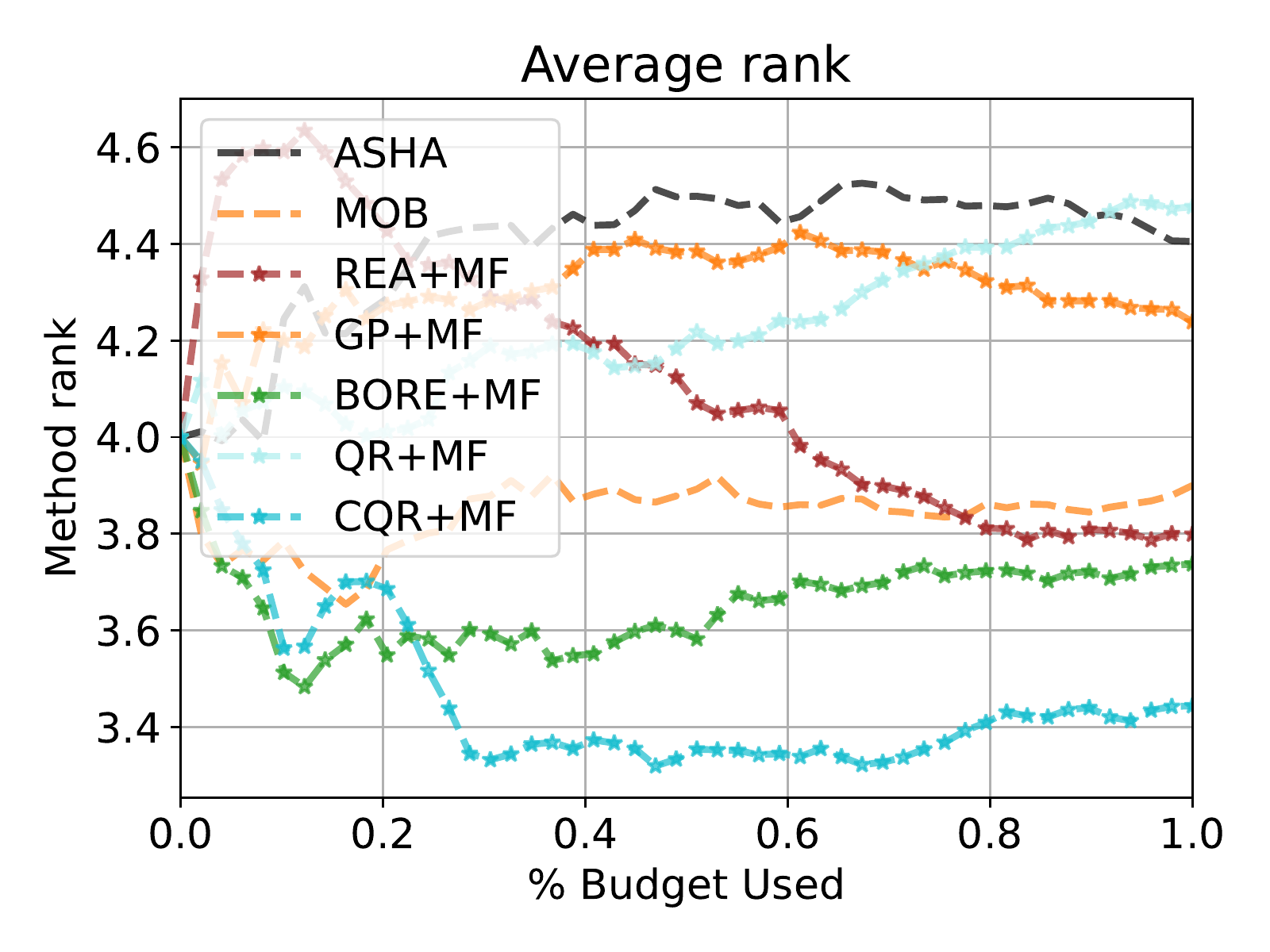}~
\includegraphics[width=0.5\textwidth]{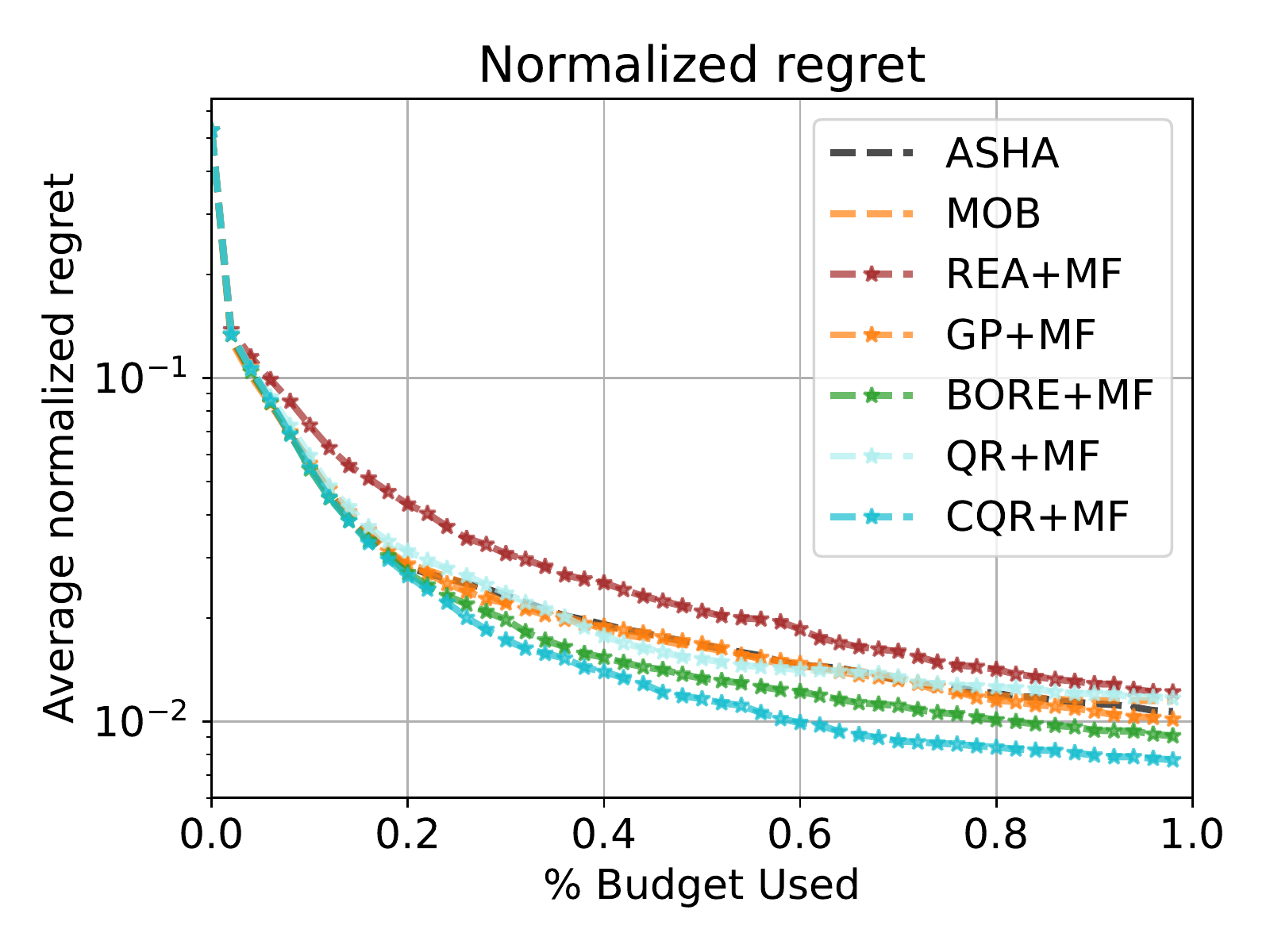} 
\end{subfigure}
\begin{subfigure}{.34\textwidth}
  \centering
\includegraphics[width=0.99\textwidth]{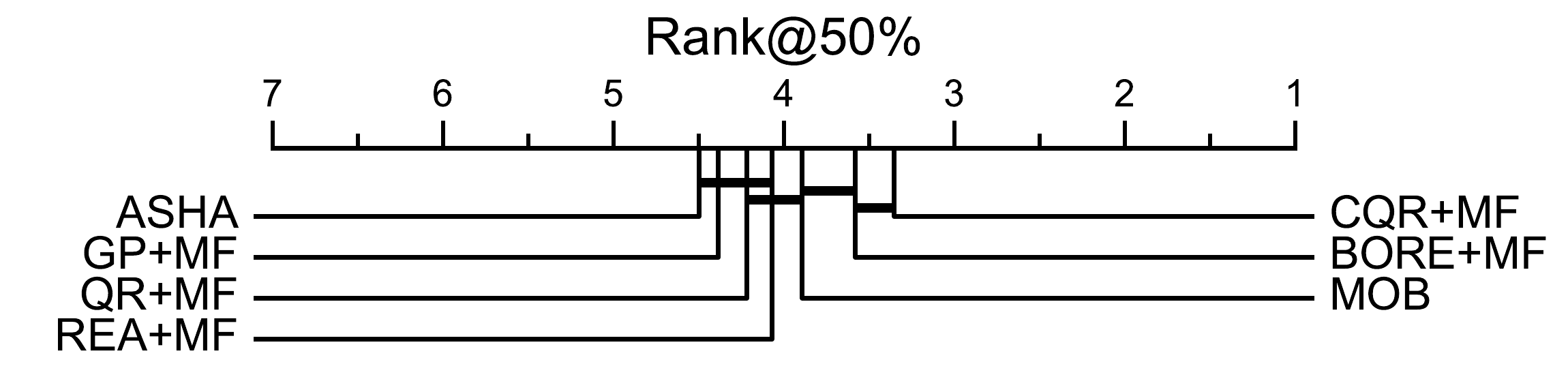}\\
\vspace{0.5cm}
\includegraphics[width=0.99\textwidth]{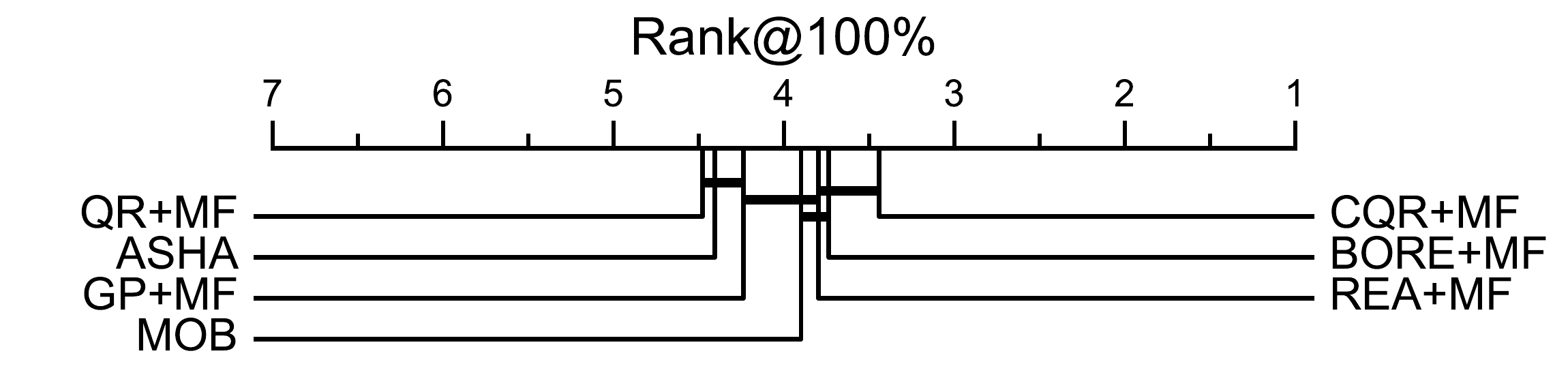}\\

\end{subfigure}

\caption{Performance for single fidelity (top) multi-fidelity (middle) and multi-fidelity variants (bottom) for average rank over all tasks (left), normalized regret (middle) and critical diagrams obtained at 50\% and 100\% of the total budget (right). \label{fig:aggregate}}
\end{figure*}

\paragraph{Results discussion for single-fidelity.}
Aggregated metrics of single-fidelity methods are shown in Figure \ref{fig:aggregate} top. Our proposed method (\CQR{}) outperforms all baselines in term of both rank and regret. In particular, critical diagrams shows that our method outperforms all baselines at 50\% of the tuning budget and is only tied statistically to (\BORE) given 100\% of the budget. Our results also show that GP-based tuning performs really well in the low sample regime where it can rely on its informative prior. After enough observations, our data-driven approach starts to outperform Bayesian competitor as it can model irregular noise better.

\paragraph{Analyzing surrogate performance.}
To better understand performance gains, we now analyze the properties of different surrogate models in more detail. In Tab. \ref{tab:surr-perf}, we compare the surrogate accuracy (RMSE), the quality of their uncertainty estimate (calibration error) and their runtime. 
In particular, we measure those metrics for different number of samples $\numevaluations$. In each case, we draw a random subset of size $\numevaluations$ to train the surrogate model and then evaluate the three metrics on remaining unseen examples. Results are averaged across seeds and benchmarks and we normalize the target with quantile normalization. Results per task are also given in the appendix as well as the definition of calibration error metric.

We compare three surrogates: the baseline \GP{}, conformalized quantile regression 
\CQR{} as well as quantile regression \QR{}. Compared to \GP{}, the RMSE of the boosted-trees surrogates is always better, which is expected as boosted-trees are known for their good performance on tabular data. However, a critical aspect in HPO is to provide good uncertainty estimates in addition to good mean predictors as uncertainty plays a critical in balancing exploration versus exploitation.

To measure the quality of uncertainty estimates, we analyze the calibration error of the different surrogates which measures how much over or under confident are each predicted quantiles of the surrogate predictive distribution. When few observations are available (e.g. when $n \leq 64$), the quality of uncertainty estimates of \GP{} is better compared to both boosted tree-methods, which is expected as \GP{} can rely on their prior in this regime whereas data-driven \QR{} and \CQR{} lack the amount of data to estimate quantile levels well enough. The lack of data also means that \CQR{} cannot adjust confidence intervals accurately given that its validation set is too small and its calibration performance just matches \QR{}. However, as the number of samples increases, the calibration of tree-based methods quickly becomes better, which underlines that quantile regression better fits the noise function observed in the benchmarks. As expected given the theoretical coverage guarantees, the calibration of \CQR{} exceeds the calibration of \QR{} given sufficient data making it a better suited surrogate for HPO.

\begin{table}
\caption{RMSE, Calibration error and runtime for different surrogates when increasing the number of samples. \label{tab:surr-perf}}
\scriptsize
\center
\addtolength{\tabcolsep}{-2.5pt}    
\begin{tabular}{l|rrr|rrr|rrr}
\toprule
 & \multicolumn{3}{c}{RMSE $\downarrow$} & \multicolumn{3}{c}{Calibration error $\downarrow$} & \multicolumn{3}{c}{Runtime $\downarrow$} \\
model & GP & QR & CQR & GP & QR & CQR & GP & QR & CQR \\
$n$ &  &  &  &  &  &  &  &  &  \\
\midrule
16 & 1.01 & 0.78 & 0.81 & 0.06 & 0.13 & 0.13 & 1.11 & 1.13 & 1.06 \\
64 & 0.92 & 0.57 & 0.58 & 0.04 & 0.10 & 0.08 & 1.71 & 1.48 & 1.43 \\
256 & 0.85 & 0.43 & 0.44 & 0.06 & 0.05 & 0.04 & 2.27 & 1.75 & 1.71 \\
1024 & 0.58 & 0.37 & 0.37 & 0.11 & 0.04 & 0.03 & 20.03 & 2.23 & 2.16 \\
\bottomrule
\end{tabular}

\addtolength{\tabcolsep}{2.5pt}
\end{table}

\paragraph{Results discussion for multi-fidelity.}
Next, we analyze the performance in the multi-fidelity setting in the middle of Fig. \ref{fig:aggregate} where we show the performance of (\CQRMF{}) which combines the single-fidelity method with the simple transformation described in section \ref{sec:multifidelity}.

In contrast to single-fidelity, multi-fidelity optimization quickly yields many hundreds of observations and the majority of the tuning process lies in the high-data regime. In this setup, (\CQRMF{}) shows significant improvement 
in term of HPO performance. In particular, while most multi-fidelity approaches are not statistically distinguishable from \ASHA{}, our proposed method (\CQRMF{}) offers statistically significant improvements over \ASHA{} at all times and over all other model-based multi-fidelity methods after spending 50\% and 100\% of the total budget. We understand the improvement mainly comes from the multi-fidelity setting which offers more observations to the HPO tuning methods which plays into the strengths of the \CQR{} surrogate illustrated in the previous paragraph in term of accuracy and calibration.

\paragraph{Ablation study.}

The surrogate analysis showed that conformal prediction improves the calibration of quantile regression but has little effect on the surrogate RMSE. To examine the benefit of this contribution on the HPO setting, we next evaluate quantile regression with and without applying conformal correction (\QRMF{}) in the bottom of Fig. \ref{fig:aggregate}. The performance of \QRMF{} is much worse than \CQRMF{} which highlights the benefit of the better uncertainty provided by conformalizing predictions.

Next, we investigate in Fig. \ref{fig:aggregate} the performance of the best single-fidelity methods \REA{}, \GP{} and \BORE{} extended to the multi-fidelity setting with our simple extension. As for \QRMF{} and \CQRMF{}, all those methods surrogates are trained using the last fidelity observed for each hyperparameter and the worst configurations are stopped with asynchronous successful halving. While those methods perform worse than \CQRMF{}, they all outperform \ASHA{} in term of average rank and regret except for \REA{} which we believe is due to the lower  performance of the method. Those simple extensions also match or improve over the performance of dedicated model-based multi-fidelity methods.

This illustrates the robustness of the proposed extension with respect to the choice of the single-fidelity method also shows the potential of future work to extend other advanced single fidelity methods - for instance multi-objective or constrained - to the multi-fidelity case.

\section{Conclusion}

We presented a new HPO approach that allows to use highly accurate tabular predictors, such as gradient boosted trees, while obtaining calibrated uncertainty estimates through conformal predictions. In addition, we showed that most single-fidelity methods can be extended to the multi-fidelity case by just using the last fidelity available while achieving good performance.

The method we proposed has a few limitations. For instance the use of Thompson Sampling may be less efficient in the presence of many hyperparameters, as such further work could consider extending the method with other acquisition functions, such as UCB. Further work could also investigate providing regret bounds or extension to support multi-objective or transfer learning scenarios.

\bibliography{conformal_icml}

\begin{thebibliography}{46}
\providecommand{\natexlab}[1]{#1}
\providecommand{\url}[1]{\texttt{#1}}
\expandafter\ifx\csname urlstyle\endcsname\relax
  \providecommand{\doi}[1]{doi: #1}\else
  \providecommand{\doi}{doi: \begingroup \urlstyle{rm}\Url}\fi

\bibitem[Awad et~al.(2021)Awad, Mallik, and Hutter]{awad-ijcai21}
Awad, N., Mallik, N., and Hutter, F.
\newblock Dehb: Evolutionary hyberband for scalable, robust and efficient
  hyperparameter optimization.
\newblock In \emph{Proceedings of the Thirtieth International Joint Conference
  on Artificial Intelligence (IJCAI'21)}, 2021.

\bibitem[Bassett \& Koenker(1982)Bassett and Koenker]{1982}
Bassett, G. and Koenker, R.
\newblock An empirical quantile function for linear models with iid errors.
\newblock \emph{Journal of the American Statistical Association}, 77\penalty0
  (378):\penalty0 407–415, Jun 1982.
\newblock ISSN 1537-274X.
\newblock \doi{10.1080/01621459.1982.10477826}.
\newblock URL \url{http://dx.doi.org/10.1080/01621459.1982.10477826}.

\bibitem[Bergstra \& Bengio(2012)Bergstra and Bengio]{bergstra-jmlr12a}
Bergstra, J. and Bengio, Y.
\newblock Random search for hyper-parameter optimization.
\newblock \emph{Journal of Machine Learning Research}, 2012.

\bibitem[Bergstra et~al.()Bergstra, Bardenet, Bengio, and Kegl]{Bergstra:11}
Bergstra, J., Bardenet, R., Bengio, Y., and Kegl, B.
\newblock Algorithms for hyperparameter optimization.
\newblock pp.\  2546--2554.

\bibitem[Bergstra et~al.(2011)Bergstra, Bardenet, Bengio, and
  K{\'e}gl]{bergstra-nips11a}
Bergstra, J., Bardenet, R., Bengio, Y., and K{\'e}gl, B.
\newblock Algorithms for hyper-parameter optimization.
\newblock In \emph{Proceedings of the 24th International Conference on Advances
  in Neural Information Processing Systems (NIPS'11)}, 2011.

\bibitem[Chen \& Guestrin(2016)Chen and Guestrin]{chen2016xgboost}
Chen, T. and Guestrin, C.
\newblock Xgboost: A scalable tree boosting system.
\newblock In \emph{Proceedings of the 22nd acm sigkdd international conference
  on knowledge discovery and data mining}, pp.\  785--794, 2016.

\bibitem[Cowen-Rivers et~al.(2020)Cowen-Rivers, Lyu, Tutunov, Wang, Grosnit,
  Griffiths, Maraval, Jianye, Wang, Peters, and Ammar]{cowen2020hebo}
Cowen-Rivers, A.~I., Lyu, W., Tutunov, R., Wang, Z., Grosnit, A., Griffiths,
  R.~R., Maraval, A.~M., Jianye, H., Wang, J., Peters, J., and Ammar, H.~B.
\newblock Hebo pushing the limits of sample-efficient hyperparameter
  optimisation, 2020.
\newblock URL \url{https://arxiv.org/abs/2012.03826}.

\bibitem[Dem{\v{s}}ar(2006)]{demsar06a}
Dem{\v{s}}ar, J.
\newblock Statistical comparisons of classifiers over multiple data sets.
\newblock \emph{Journal of Machine Learning Research}, 7\penalty0 (1):\penalty0
  1--30, 2006.
\newblock URL \url{http://jmlr.org/papers/v7/demsar06a.html}.

\bibitem[Domhan et~al.(2015)Domhan, Springenberg, and Hutter]{domhan:15}
Domhan, T., Springenberg, J.~T., and Hutter, F.
\newblock Speeding up automatic hyperparameter optimization of deep neural
  networks by extrapolation of learning curves.
\newblock In \emph{Twenty-fourth international joint conference on artificial
  intelligence}, 2015.

\bibitem[Dong \& Yang(2020)Dong and Yang]{Dong:20}
Dong, X. and Yang, Y.
\newblock {NAS-Bench-201}: Extending the scope of reproducible neural
  architecture search.
\newblock Technical Report arXiv:2001.00326 [cs.CV], 2020.

\bibitem[Doyle(2022)]{doyle-arxiv22}
Doyle, R.
\newblock Model agnostic conformal hyperparameter optimization.
\newblock \emph{arXiv:2207.03017 [cs.LG]}, 2022.

\bibitem[Falkner et~al.(2018)Falkner, Klein, and Hutter]{Falkner:18}
Falkner, S., Klein, A., and Hutter, F.
\newblock {BOHB}: Robust and efficient hyperparameter optimization at scale.
\newblock In \emph{Proceedings of the 35th International Conference on Machine
  Learning (ICML 2018)}, pp.\  1436--1445, 2018.

\bibitem[Feurer \& Hutter(2018)Feurer and Hutter]{feurer-automlbook18a}
Feurer, M. and Hutter, F.
\newblock Hyperparameter optimization.
\newblock In \emph{Automatic Machine Learning: Methods, Systems, Challenges}.
  Springer, 2018.

\bibitem[Friedman(2001)]{Friedman2001}
Friedman, J.~H.
\newblock Greedy function approximation: A gradient boosting machine.
\newblock \emph{The Annals of Statistics}, 29\penalty0 (5):\penalty0
  1189--1232, 2001.
\newblock ISSN 00905364.
\newblock URL \url{http://www.jstor.org/stable/2699986}.

\bibitem[Gasthaus et~al.(2019)Gasthaus, Benidis, Wang, Rangapuram, Salinas,
  Flunkert, and Januschowski]{gasthaus19a}
Gasthaus, J., Benidis, K., Wang, Y., Rangapuram, S.~S., Salinas, D., Flunkert,
  V., and Januschowski, T.
\newblock Probabilistic forecasting with spline quantile function rnns.
\newblock In Chaudhuri, K. and Sugiyama, M. (eds.), \emph{Proceedings of the
  Twenty-Second International Conference on Artificial Intelligence and
  Statistics}, volume~89 of \emph{Proceedings of Machine Learning Research},
  pp.\  1901--1910. PMLR, 16--18 Apr 2019.
\newblock URL \url{https://proceedings.mlr.press/v89/gasthaus19a.html}.

\bibitem[Hutter et~al.(2011)Hutter, Hoos, and Leyton-Brown]{hutter-lion11a}
Hutter, F., Hoos, H., and Leyton-Brown, K.
\newblock Sequential model-based optimization for general algorithm
  configuration.
\newblock In \emph{Proceedings of the Fifth International Conference on
  Learning and Intelligent Optimization (LION'11)}, 2011.

\bibitem[Jamieson \& Talwalkar(2016)Jamieson and Talwalkar]{jamieson-aistats16}
Jamieson, K. and Talwalkar, A.
\newblock Non-stochastic best arm identification and hyperparameter
  optimization.
\newblock In \emph{Proceedings of the 17th International Conference on
  Artificial Intelligence and Statistics (AISTATS'16)}, 2016.

\bibitem[Jones et~al.(1998)Jones, Schonlau, and Welch]{jones1998efficient}
Jones, D.~R., Schonlau, M., and Welch, W.~J.
\newblock Efficient global optimization of expensive black-box functions.
\newblock \emph{Journal of Global optimization}, 13\penalty0 (4):\penalty0 455,
  1998.

\bibitem[Karnin et~al.(2013)Karnin, Koren, and Somekh]{karnin13}
Karnin, Z., Koren, T., and Somekh, O.
\newblock Almost optimal exploration in multi-armed bandits.
\newblock In Dasgupta, S. and McAllester, D. (eds.), \emph{Proceedings of the
  30th International Conference on Machine Learning}, volume~28 of
  \emph{Proceedings of Machine Learning Research}, pp.\  1238--1246, Atlanta,
  Georgia, USA, 17--19 Jun 2013. PMLR.
\newblock URL \url{https://proceedings.mlr.press/v28/karnin13.html}.

\bibitem[Klein \& Hutter(2019)Klein and Hutter]{Klein:19a}
Klein, A. and Hutter, F.
\newblock Tabular benchmarks for joint architecture and hyperparameter
  optimization.
\newblock Technical Report arXiv:1905.04970 [cs.LG], 2019.

\bibitem[Klein et~al.(2017)Klein, Falkner, Springenberg, and
  Hutter]{klein-iclr17}
Klein, A., Falkner, S., Springenberg, J.~T., and Hutter, F.
\newblock Learning curve prediction with {Bayesian} neural networks.
\newblock In \emph{International Conference on Learning Representations
  (ICLR'17)}, 2017.

\bibitem[Klein et~al.(2020)Klein, Tiao, Lienart, Archambeau, and
  Seeger]{Klein:mobster}
Klein, A., Tiao, L., Lienart, T., Archambeau, C., and Seeger, M.
\newblock Model-based asynchronous hyperparameter and neural architecture
  search.
\newblock Number 2003.10865 [cs.LG], 2020.

\bibitem[Kuleshov et~al.(2018)Kuleshov, Fenner, and Ermon]{Kuleshov2018}
Kuleshov, V., Fenner, N., and Ermon, S.
\newblock Accurate uncertainties for deep learning using calibrated regression.
\newblock \emph{CoRR}, abs/1807.00263, 2018.
\newblock URL \url{http://arxiv.org/abs/1807.00263}.

\bibitem[Li et~al.(2017)Li, Jamieson, DeSalvo, Rostamizadeh, and
  Talwalkar]{li-iclr17}
Li, L., Jamieson, K., DeSalvo, G., Rostamizadeh, A., and Talwalkar, A.
\newblock Hyperband: Bandit-based configuration evaluation for hyperparameter
  optimization.
\newblock In \emph{International Conference on Learning Representations
  (ICLR'17)}, 2017.

\bibitem[Li et~al.(2019)Li, Jamieson, Rostamizadeh, Gonina, Hardt, Recht, and
  Talwalkar]{Li:19}
Li, L., Jamieson, K., Rostamizadeh, A., Gonina, E., Hardt, M., Recht, B., and
  Talwalkar, A.
\newblock Massively parallel hyperparameter tuning.
\newblock Technical Report 1810.05934v4 [cs.LG], 2019.

\bibitem[Li et~al.(2022)Li, Shen, Jiang, Zhang, Li, Liu, Zhang, and
  Cui]{Yang2022}
Li, Y., Shen, Y., Jiang, H., Zhang, W., Li, J., Liu, J., Zhang, C., and Cui, B.
\newblock Hyper-tune: Towards efficient hyper-parameter tuning at scale.
\newblock \emph{Proc. VLDB Endow.}, 15\penalty0 (6):\penalty0 1256–1265, jun
  2022.
\newblock ISSN 2150-8097.
\newblock \doi{10.14778/3514061.3514071}.
\newblock URL \url{https://doi.org/10.14778/3514061.3514071}.

\bibitem[Mohr \& van Rijn(2022)Mohr and van Rijn]{mohr-arxiv22}
Mohr, F. and van Rijn, J.~N.
\newblock Learning curves for decision making in supervised machine learning --
  a survey.
\newblock \emph{arXiv:2201.12150 [cs.LG]}, 2022.

\bibitem[Moriconi et~al.(2020)Moriconi, Kumar, and
  Deisenroth]{moriconi2020high}
Moriconi, R., Kumar, K.~S., and Deisenroth, M.~P.
\newblock High-dimensional bayesian optimization with projections using
  quantile gaussian processes.
\newblock \emph{Optimization Letters}, 14:\penalty0 51--64, 2020.

\bibitem[Pfisterer et~al.(2022)Pfisterer, Schneider, Moosbauer, Binder, and
  Bischl]{pfisterer2022yahpo}
Pfisterer, F., Schneider, L., Moosbauer, J., Binder, M., and Bischl, B.
\newblock Yahpo gym-an efficient multi-objective multi-fidelity benchmark for
  hyperparameter optimization.
\newblock In \emph{First Conference on Automated Machine Learning (Main
  Track)}, 2022.

\bibitem[Picheny et~al.(2013)Picheny, Ginsbourger, Richet, and
  Caplin]{Picheny13}
Picheny, V., Ginsbourger, D., Richet, Y., and Caplin, G.
\newblock Quantile-based optimization of noisy computer experiments with
  tunable precision.
\newblock \emph{Technometrics}, 55\penalty0 (1):\penalty0 2--13, 2013.
\newblock \doi{10.1080/00401706.2012.707580}.
\newblock URL \url{https://doi.org/10.1080/00401706.2012.707580}.

\bibitem[Real et~al.(2019)Real, Aggarwal, Huang, and Le]{real2019}
Real, E., Aggarwal, A., Huang, Y., and Le, Q.~V.
\newblock Regularized evolution for image classifier architecture search, 2019.

\bibitem[Romano et~al.(2019)Romano, Patterson, and Candès]{romano}
Romano, Y., Patterson, E., and Candès, E.~J.
\newblock Conformalized quantile regression.
\newblock 2019.
\newblock \doi{10.48550/ARXIV.1905.03222}.

\bibitem[Salinas et~al.(2020)Salinas, Shen, and Perrone]{salinas20a}
Salinas, D., Shen, H., and Perrone, V.
\newblock A quantile-based approach for hyperparameter transfer learning.
\newblock In III, H.~D. and Singh, A. (eds.), \emph{Proceedings of the 37th
  International Conference on Machine Learning}, volume 119 of
  \emph{Proceedings of Machine Learning Research}, pp.\  8438--8448. PMLR,
  13--18 Jul 2020.
\newblock URL \url{https://proceedings.mlr.press/v119/salinas20a.html}.

\bibitem[Salinas et~al.(2022)Salinas, Seeger, Klein, Perrone, Wistuba, and
  Archambeau]{salinas2022syne}
Salinas, D., Seeger, M., Klein, A., Perrone, V., Wistuba, M., and Archambeau,
  C.
\newblock Syne tune: A library for large scale hyperparameter tuning and
  reproducible research.
\newblock In \emph{International Conference on Automated Machine Learning},
  pp.\  16--1. PMLR, 2022.

\bibitem[Shafer \& Vovk(2007)Shafer and Vovk]{Shafer07}
Shafer, G. and Vovk, V.
\newblock A tutorial on conformal prediction.
\newblock \emph{CoRR}, abs/0706.3188, 2007.
\newblock URL \url{http://arxiv.org/abs/0706.3188}.

\bibitem[Shahriari et~al.(2016)Shahriari, Swersky, Wang, Adams, and
  de~Freitas]{shahriari-ieee16a}
Shahriari, B., Swersky, K., Wang, Z., Adams, R., and de~Freitas, N.
\newblock Taking the human out of the loop: {A} review of {B}ayesian
  optimization.
\newblock \emph{Proceedings of the {IEEE}}, 2016.

\bibitem[Siems et~al.(2020)Siems, Zimmer, Zela, Lukasik, Keuper, and
  Hutter]{nas301}
Siems, J., Zimmer, L., Zela, A., Lukasik, J., Keuper, M., and Hutter, F.
\newblock Nas-bench-301 and the case for surrogate benchmarks for neural
  architecture search.
\newblock \emph{CoRR}, abs/2008.09777, 2020.
\newblock URL \url{https://arxiv.org/abs/2008.09777}.

\bibitem[Snoek et~al.(2012)Snoek, Larochelle, and Adams]{snoek-nips12a}
Snoek, J., Larochelle, H., and Adams, R.~P.
\newblock Practical {B}ayesian optimization of machine learning algorithms.
\newblock In \emph{Proceedings of the 25th International Conference on Advances
  in Neural Information Processing Systems (NIPS'12)}, 2012.

\bibitem[Snoek et~al.(2015)Snoek, Rippel, Swersky, Kiros, Satish, Sundaram,
  Patwary, Prabhat, and Adams]{snoek-icml15}
Snoek, J., Rippel, O., Swersky, K., Kiros, R., Satish, N., Sundaram, N.,
  Patwary, M., Prabhat, and Adams, R.
\newblock Scalable {B}ayesian optimization using deep neural networks.
\newblock In \emph{Proceedings of the 32nd International Conference on Machine
  Learning (ICML'15)}, 2015.

\bibitem[Springenberg et~al.(2016)Springenberg, Klein, Falkner, and
  Hutter]{springenberg-nips16}
Springenberg, J.~T., Klein, A., Falkner, S., and Hutter, F.
\newblock Bayesian optimization with robust bayesian neural networks.
\newblock In \emph{Proceedings of the 29th International Conference on Advances
  in Neural Information Processing Systems (NIPS'16)}, 2016.

\bibitem[Srinivas et~al.(2012)Srinivas, Krause, Kakade, and
  Seeger]{Srinivas2012}
Srinivas, N., Krause, A., Kakade, S.~M., and Seeger, M.~W.
\newblock Information-theoretic regret bounds for gaussian process optimization
  in the bandit setting.
\newblock \emph{{IEEE} Transactions on Information Theory}, 58\penalty0
  (5):\penalty0 3250--3265, may 2012.
\newblock \doi{10.1109/tit.2011.2182033}.
\newblock URL \url{https://doi.org/10.1109%2Ftit.2011.2182033}.

\bibitem[Stanton et~al.(2022)Stanton, Maddox, and Wilson]{Stanton22}
Stanton, S., Maddox, W., and Wilson, A.~G.
\newblock Bayesian optimization with conformal coverage guarantees, 2022.
\newblock URL \url{https://arxiv.org/abs/2210.12496}.

\bibitem[Swersky et~al.(2014)Swersky, Snoek, and Adams]{Swersky:14}
Swersky, K., Snoek, J., and Adams, R.~P.
\newblock Freeze-thaw bayesian optimization, 2014.
\newblock URL \url{https://arxiv.org/abs/1406.3896}.

\bibitem[Tiao et~al.(2021)Tiao, Klein, Seeger, Bonilla, Archambeau, and
  Ramos]{tiao2021bore}
Tiao, L.~C., Klein, A., Seeger, M.~W., Bonilla, E.~V., Archambeau, C., and
  Ramos, F.
\newblock {BORE}: {Bayesian} optimization by density-ratio estimation.
\newblock In \emph{International Conference on Machine Learning}, pp.\
  10289--10300. PMLR, 2021.

\bibitem[Wistuba \& Pedapati(2020)Wistuba and Pedapati]{wistuba20a}
Wistuba, M. and Pedapati, T.
\newblock Learning to rank learning curves.
\newblock In III, H.~D. and Singh, A. (eds.), \emph{Proceedings of the 37th
  International Conference on Machine Learning}, volume 119 of
  \emph{Proceedings of Machine Learning Research}, pp.\  10303--10312. PMLR,
  13--18 Jul 2020.

\bibitem[Zimmer et~al.(2021)Zimmer, Lindauer, and Hutter]{zimmer2021}
Zimmer, L., Lindauer, M., and Hutter, F.
\newblock Auto-{PyTorch} {Tabular}: {Multi}-{Fidelity} {MetaLearning} for
  {Efficient} and {Robust} {AutoDL}.
\newblock \emph{arXiv:2006.13799 [cs, stat]}, April 2021.
\newblock URL \url{http://arxiv.org/abs/2006.13799}.
\newblock arXiv: 2006.13799.

\end{thebibliography}
\bibliographystyle{icml2023}

\newpage
\appendix
\onecolumn

\section{ASHA description}

We recall how the method ASHA proposed by \citep{Li:19} works. Given some positive constant $\redfactor \geq 2$, let us define a finite set of rungs $\mathcal{R} = \{\eta^0 \minresource{}, \eta^1 * \minresource{}, \eta^2 \minresource{}, ..., \maxresource{}\}$.
Successive halving~\citep{karnin13,jamieson-aistats16} starts with a set of $N=\eta^K$ initial candidates, where for simplicity, we assume that $K = \log_{\eta} \nicefrac{\maxresource{}}{\minresource{}}$.
Now, in the first iteration, successive halving collects the performance of all $N$ configurations on $\minresource{}$ and only continues the evaluation of the top $\nicefrac{1}{\redfactor}$ configurations for the next rung level.
This process is iterated until the maximum resource level $\maxresource{}$ is reached and repeated until we reach some total budget for the entire search process.

Successive halving can be trivially parallelized in the synchronous setting, however, this will require synchronization points at each rung level to wait for stragglers.
\citet{Li:19} adapted successive halving to the asynchronous setting (ASHA), where configurations are immediately promoted to the next rung level, once we observed at least $\redfactor$ configurations. While this potentially leads to the promotion of configurations that are not among the top $1/\redfactor$ configurations, it removes any synchronization overhead and has been shown to perform very well in practice. See Algo. \ref{alg:asha} for the pseudo-code of ASHA.

\begin{algorithm}
   \caption{{\tt ASHA} pseudo-code. \label{alg:asha}}   
\begin{algorithmic}[1]
   \FUNCTION{{\tt ASHA}()}{}{}
   \STATE {\bfseries Input:} minimum resource $\minresource$, maximum resource $\maxresource$, reduction factor $\redfactor$
   \REPEAT
   \FOR{each free worker}
   \STATE $(\hp, k) = {\small \texttt{get\_job}}()$
   \STATE ${\small \texttt{run\_then\_return\_val\_loss}}(\hp, \minresource\redfactor^{k})$    
   \ENDFOR
   \FOR{\upshape completed job ($\hp$, $k$) with loss $l$} 
   \STATE Update configuration $\hp$ in rung $k$ with loss $l$.
   \ENDFOR

   \UNTIL{desired}
   \ENDFUNCTION{}
    \vspace{0.3cm}
   \FUNCTION{{\tt get\_job}()}{}{}
   \STATE {\small $\texttt{// Check if there is a promotable config.}$}
   \FOR{$k=\floor{\log_\redfactor(\maxresource/\minresource)} - 1,\dots,1,0$} 
   \STATE candidates $={\small \texttt{top\_k}}(\text{rung } k, |\text{rung } k|/\redfactor)$ 
   \STATE promotable $=\{t \in \text{candidates}: t \text{ not promoted} \}$ 
   \IF{\upshape $|\text{promotable}|>0$}
   \STATE \textbf{return} \upshape $\text{promotable}[0], k + 1$
   \ENDIF
   \STATE {\small \texttt{// If not, grow bottom rung.}}
   \STATE Suggest random configuration $\hp$. 
   \STATE \textbf{return} $\hp, 0$
   \ENDFOR

   \ENDFUNCTION{}

\end{algorithmic}

\end{algorithm}

\section{Experiment details}
\label{sec:experiment-details}

Statistics of different blackboxes are given in Table \ref{tab:blackbox-stat} and their configuration spaces are given in Table \ref{tab:search-space} together with the base distribution used for each hyperparameters which are used when sampling random candidates. For \LCBench{}, we run the 5 most expensive tasks among the 35 tasks available ("airlines", "albert", "covertype", "christine" and "Fashion-MNIST"). Since \LCBench{} does not contain all possible evaluations on a grid, we run evaluations using a $k$-nearest-neighbors surrogate with $k=1$. We use Yahpo implementation \cite{pfisterer2022yahpo} to get access to \NASSurr{} \cite{nas301}.

\begin{table}
\centering
\scriptsize
\caption{Tabulated benchmark statistics \label{tab:blackbox-stat}}
\begin{tabular}{lllll}
  Benchmark & \#Evaluations & \#Hyperparameters & \#Tasks & \#Fidelities \\
\toprule
\FCNet{} & 62208 & 9 & 4 & 100 \\
\LCBench{} & 2000 & 7 & 5 & 52 \\
\NASBench{} & 15625 & 6 & 3 & 200 \\
\NASSurr{} & NA & 35 & 1 & 97 \\
  \hline
\end{tabular}
\end{table}

\begin{table}
\centering
\scriptsize
\caption{Configuration spaces for all tabulated benchmarks.\label{tab:search-space}}
\begin{tabular}{llll}
  Benchmark & Hyperparameter & Configuration space & Domain \\
\toprule
\FCNet{} & activation\_1 & [tanh, relu]  & categorical \\
  &activation\_2 &  [tanh, relu] &  categorical \\
  &batch\_size &[8, 16, 32, 64] &finite-range log-space \\
  &dropout\_1  &[0.0,  0.3, 0.6] & finite-range \\
  &dropout\_2  &[0.0,  0.3, 0.6] & finite-range \\
  &init\_lr  &[0.0005, 0.001, 0.005, 0.01, 0.05, 0.1]& categorical \\
  &lr\_schedule & [cosine, const] & categorical \\
  &n\_units\_1  &[16, 32, 64, 128, 256, 512] &finite-range log-space \\
  &n\_units\_2  &[16, 32, 64, 128, 256, 512] &finite-range log-space \\
  \hline
\NASBench{}& x0  &[avg\_pool\_3x3, nor\_conv\_3x3, skip\_connect, nor\_conv\_1x1, none]  &categorical \\
  &x1  &[avg\_pool\_3x3, nor\_conv\_3x3, skip\_connect, nor\_conv\_1x1, none] & categorical \\
  &x2  &[avg\_pool\_3x3, nor\_conv\_3x3, skip\_connect, nor\_conv\_1x1, none]  &categorical \\
  &x3 & [avg\_pool\_3x3, nor\_conv\_3x3, skip\_connect, nor\_conv\_1x1, none] & categorical \\
  & x5 & [avg\_pool\_3x3, nor\_conv\_3x3, skip\_connect, nor\_conv\_1x1, none] & categorical \\
  \hline
  \LCBench{} & num\_layers & [1, 5] & uniform \\
   & max\_units & [64, 512] & log-uniform \\
   & batch\_size & [16, 512] & log-uniform \\
   & learning\_rate & [1e-4, 1e-1] & log-uniform \\
   & weight\_decay & [1e-5, 0.1] & uniform \\
   & momentum & [0.1, 0.99] & uniform \\
   & max\_dropout & [0.0, 1.0] & uniform \\
  \hline
  \NASSurr{} & edge-normal-\{0-13\} & [max\_pool\_3x3, avg\_pool\_3x3, skip\_connect, sep\_conv\_3x3, sep\_conv\_5x5, dil\_conv\_3x3, dil\_conv\_5x5] & categorical\\
    & node-normal-\{0-13\} & [max\_pool\_3x3, avg\_pool\_3x3, skip\_connect, sep\_conv\_3x3, sep\_conv\_5x5, dil\_conv\_3x3, dil\_conv\_5x5] & categorical \\
 & inputs-node-normal-\{3-5\} & [0\_1, 0\_2, 1\_2] & categorical \\
 & inputs-node-reduce-\{3-5\} & [0\_1, 0\_2, 1\_2] &  categorical \\
\end{tabular}
\end{table}

\paragraph{Schedulers details} 

We give here the list of parameters used for running schedulers in our experiments. In general, the \ST{} defaults have been used.
\begin{itemize}
  \item \REA{} is run with a population size of 10, and 5 samples are drawn to select a mutation from
  \item  \GP{} is run using a Mat\'{e}rn $\frac{5}2$ kernel with automatic relevance determination parameters. For each suggestion, the surrogate model is fit by marginal likelihood maximization, and a configuration is returned which maximizes the expected improvement acquisition function. This involves averaging over 20 samples of fantasy outcomes for pending evaluations.
  \item \TPE{} is based on a multi-variate kernel density estimator as proposed by \citet{Falkner:18} to capture interactions between hyperparameters, which is not possible with unit-variant kernel density estimator as used for the original TPE approach~\citep{Bergstra:11}. We limit the minimum bandwidth for the kernel density estimator to 0.1 to avoid that all probability mass is assigned to a single categorical value, which would eliminate extrapolation.
  \item \ASHA{} is running the stopping variant described in \cite{Klein:mobster} with grace period 1 and reduction factor 3, so that stopping trials happens after $1, 3, 9, \dots$ epochs. Configurations for new trials are sampled at random.
  \item \BOHB{} uses the same multi-variate kernel density estimator as \TPE{}, and hyperparameters are set to default values in \citet{Falkner:18}. Note that \BOHB{} uses the same asynchronous scheduling as \ASHA{} and \MOB{}, while the algorithm in \citet{Falkner:18} is synchronous.
  \item \MOB{} is running the same scheduling as \ASHA{}, but configurations for new trials are chosen as in Bayesian optimization. We deal with pending evaluations by averaging the acquisition function over 20 samples of fantasy outcomes.
  \item \BORE{} is evaluated with XGBoost as the classifier with default hyperparameters \citep{chen2016xgboost}. We use the default hyperparameters of the method, in particular, $\gamma = 1 / 4$ which means that the method maximizes the probability that a configuration is in the top 25\% of configurations.
  \item \HT{} uses the same rung levels (given by grace period and reduction factor) as \ASHA{} and fit independent Gaussian process models to data at each rung level. Compared to \MOB{}, \HT{} is using a more advanced acquisition function, which averages the rung level models with a weighting depending on the fraction of trials which flipped ranks between low and high levels. 
  \item \QR{} and \CQR{}: we estimate $\numquantiles=4$ quantiles by fitting gradient boosted trees models with quantile-losses and we use the same hyperparameter as \BORE{} for the boosted-trees. We conformalize only when more than 32 samples are available to avoid poorly estimated correction due to too little samples and use 10\% of the data for validation. We sample $\numcandidates = 2000$ candidates when performing independent Thompson Sampling to select the next candidate.
  \item \{\REA{}/\GP{}/\BORE{}/\CQR\} + MF: for these methods, we run ASHA by calling each single fidelity method when a new configuration has to be suggested using the data transformation proposed in section \ref{sec:multifidelity} to obtain the method observations $\observations = \{(x_i, z_i)\}_{i=1}^\numevaluations$.
\end{itemize}

All multi-fidelity methods based on successful-halving/\ASHA{} (e.g. all except \BOHB{}) uses a single bracket. We use implementations of all baselines provided in Syne Tune (\url{https://github.com/awslabs/syne-tune}).

\section{Performance per task}

\begin{figure*}
\includegraphics[width=0.31\textwidth]{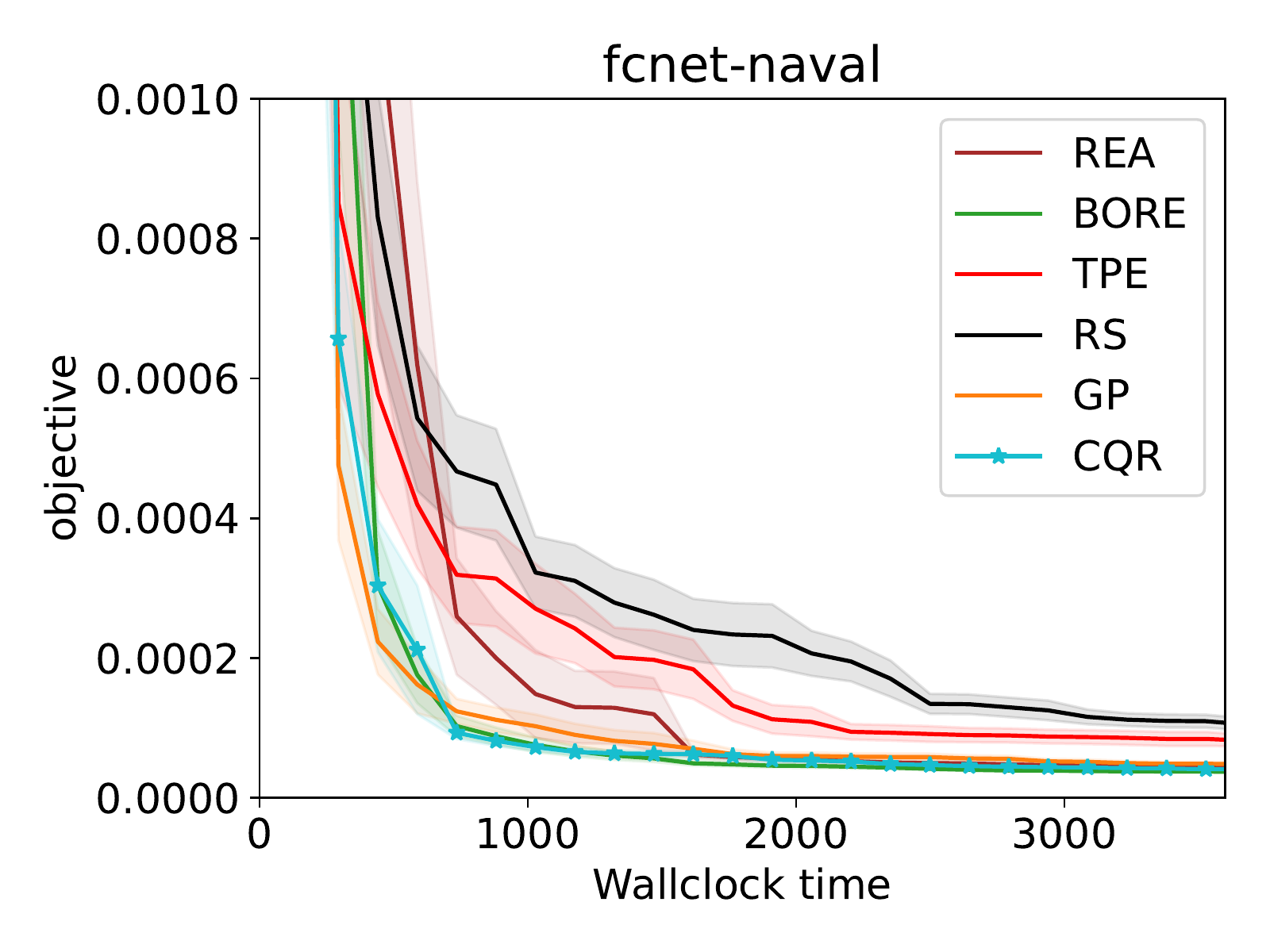}
\includegraphics[width=0.31\textwidth]{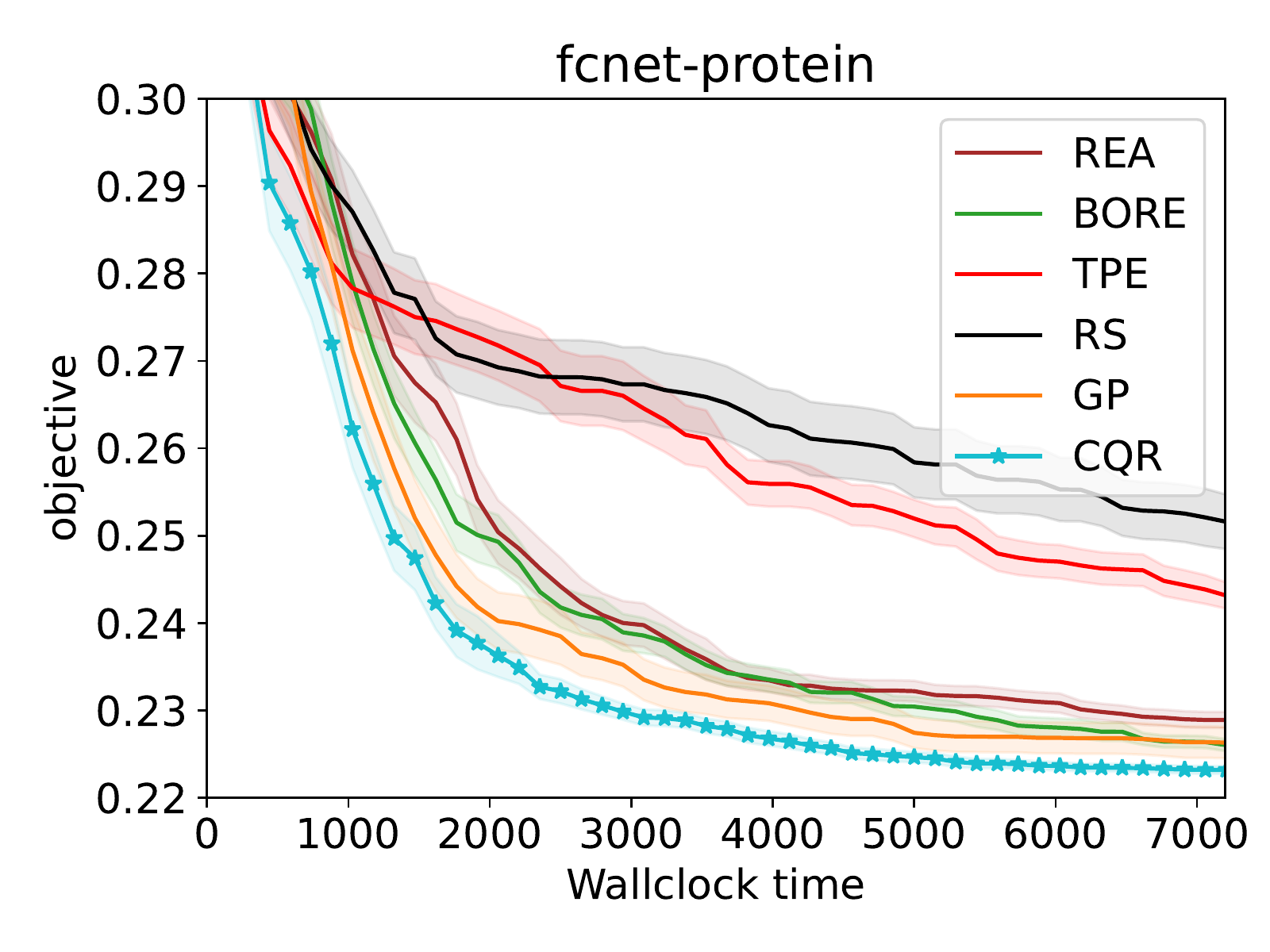}
\includegraphics[width=0.31\textwidth]{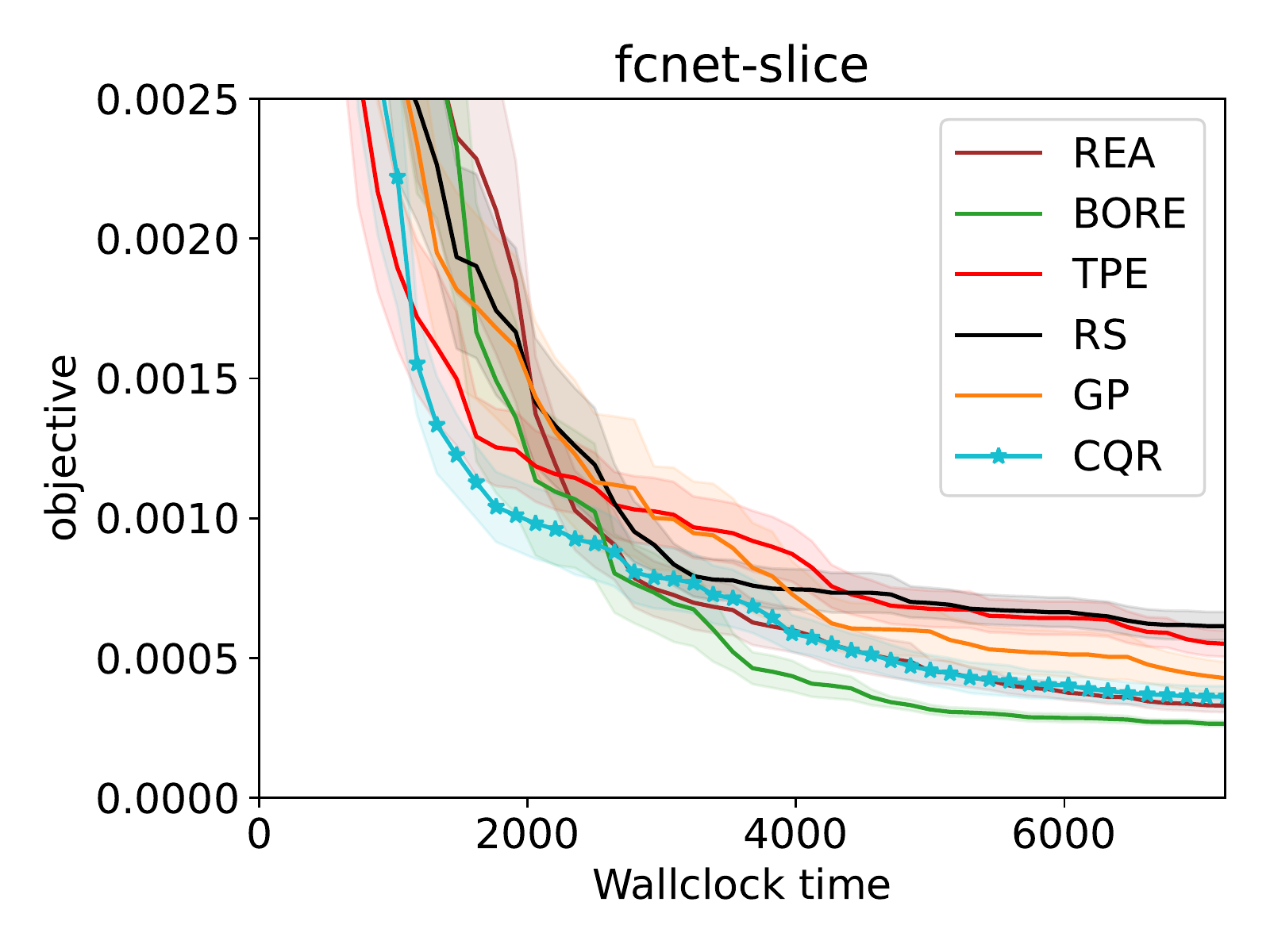} \\
\includegraphics[width=0.31\textwidth]{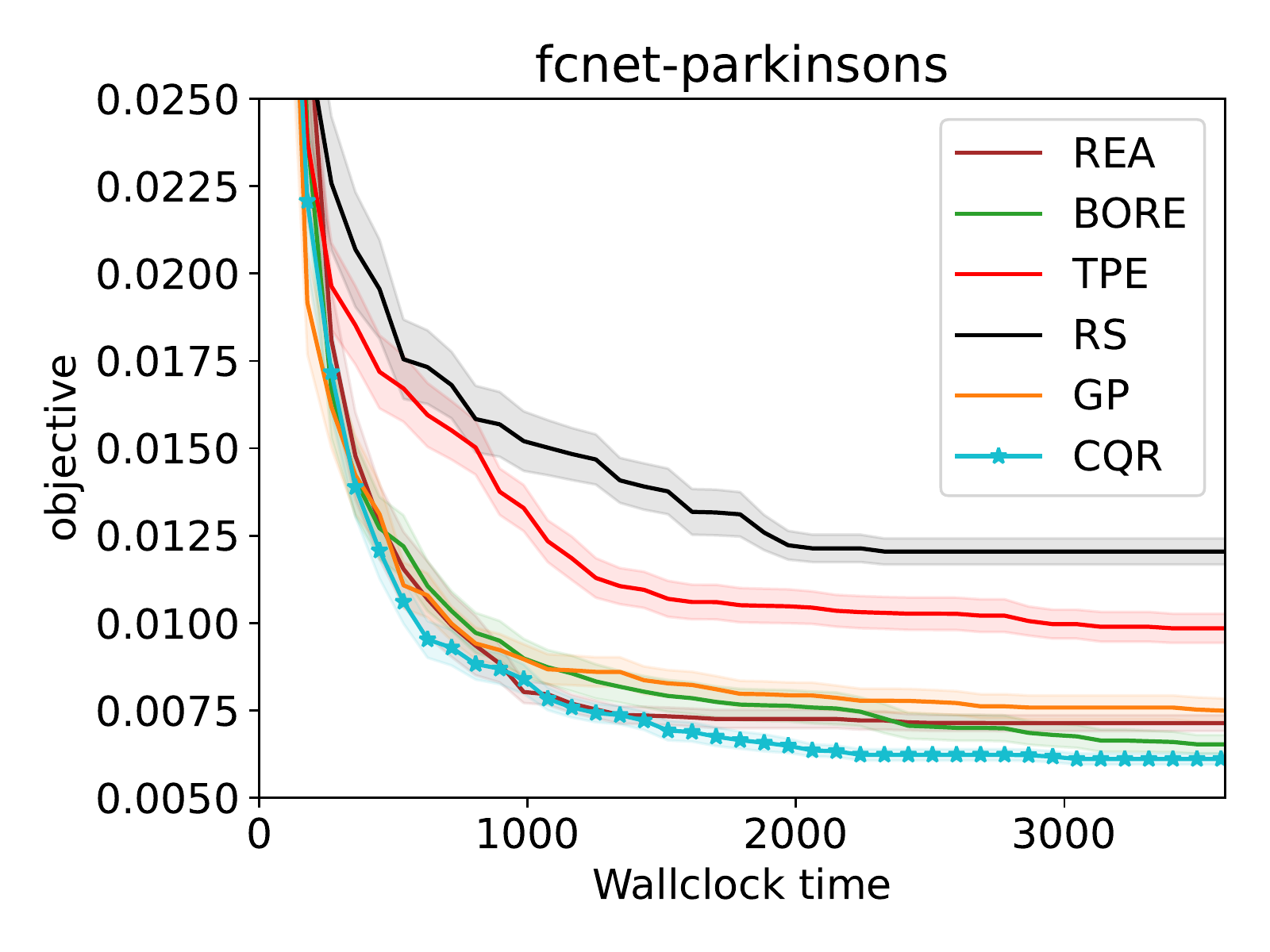} 
\includegraphics[width=0.31\textwidth]{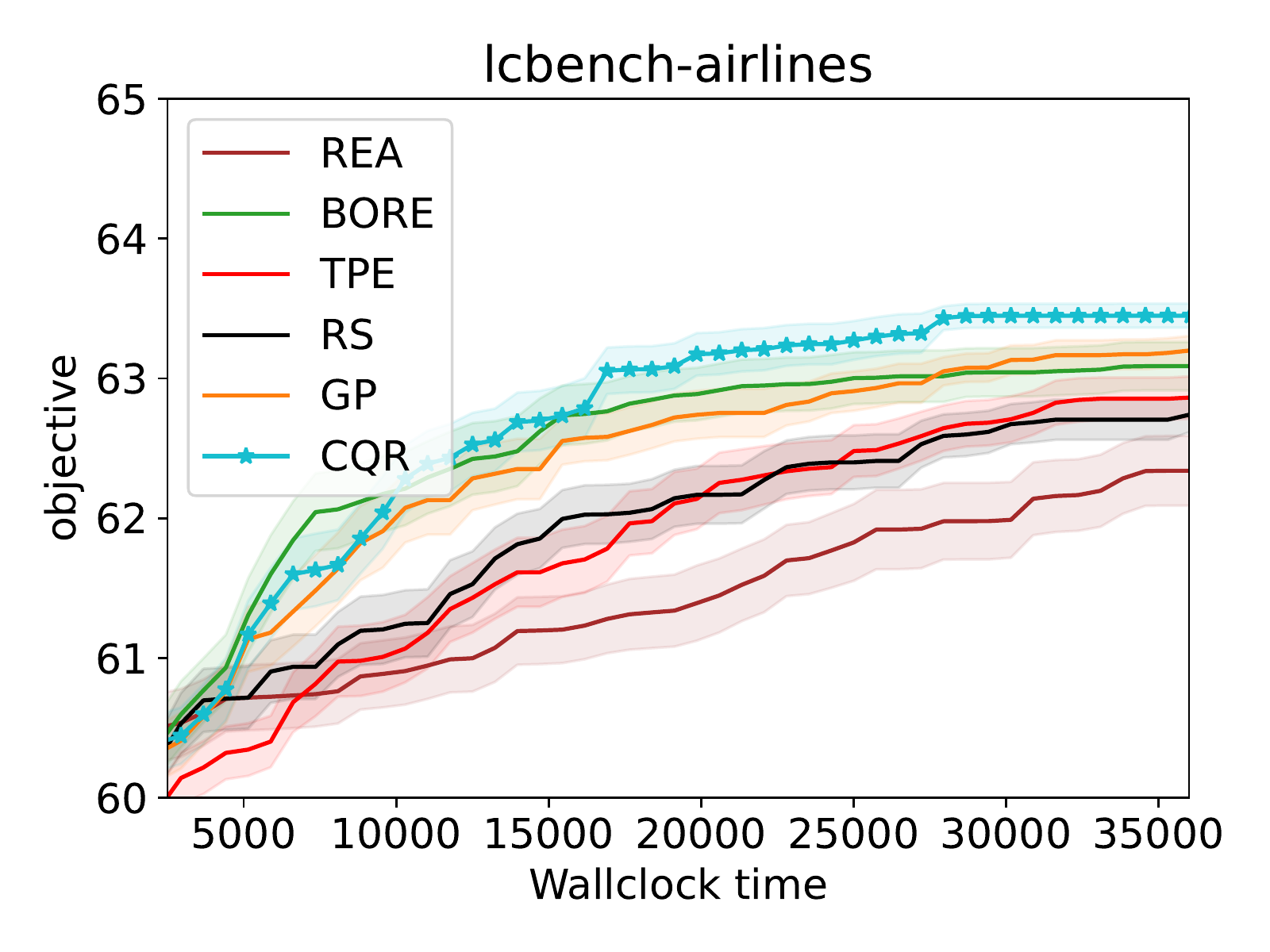}
\includegraphics[width=0.31\textwidth]{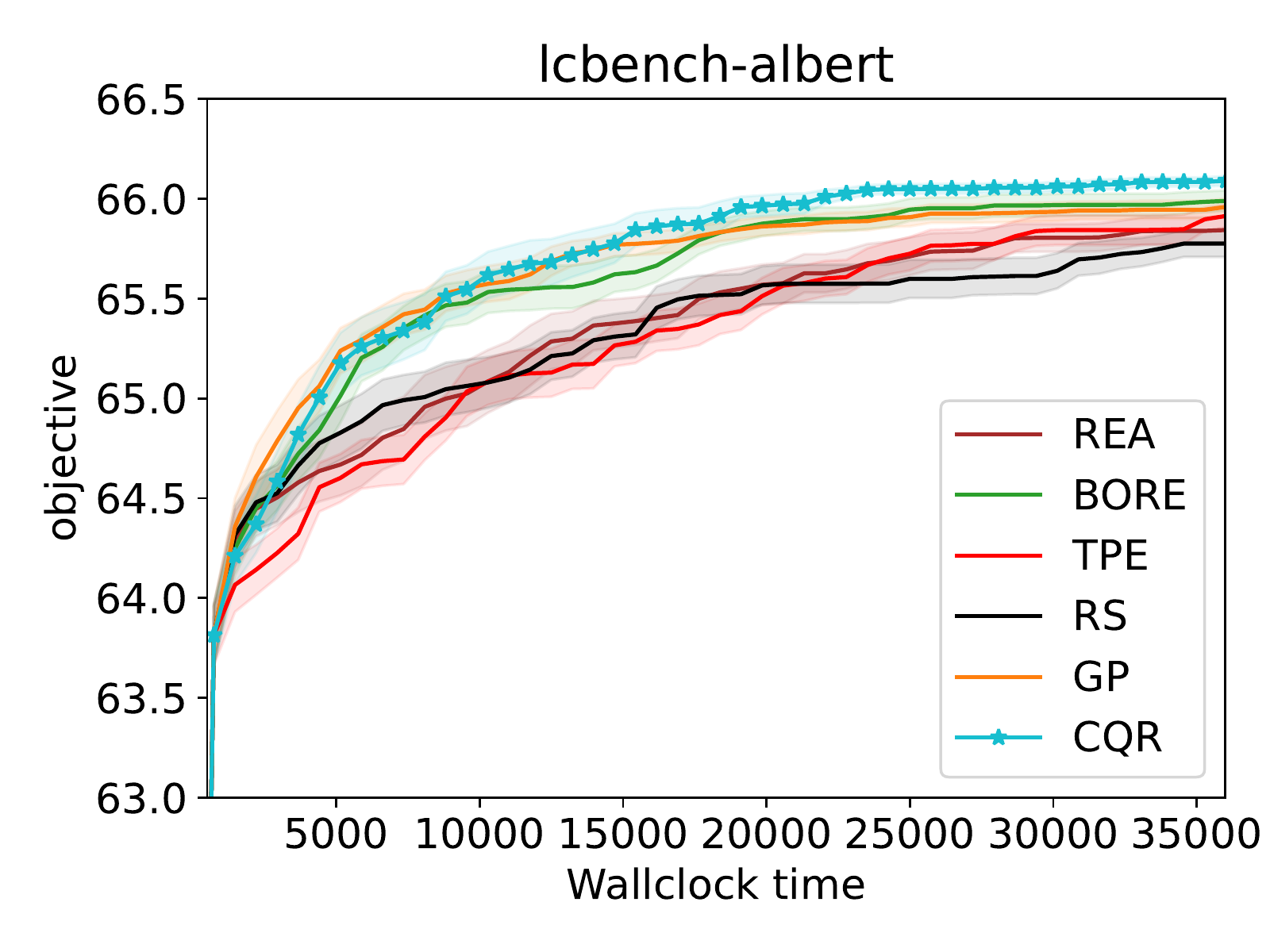} \\
\includegraphics[width=0.31\textwidth]{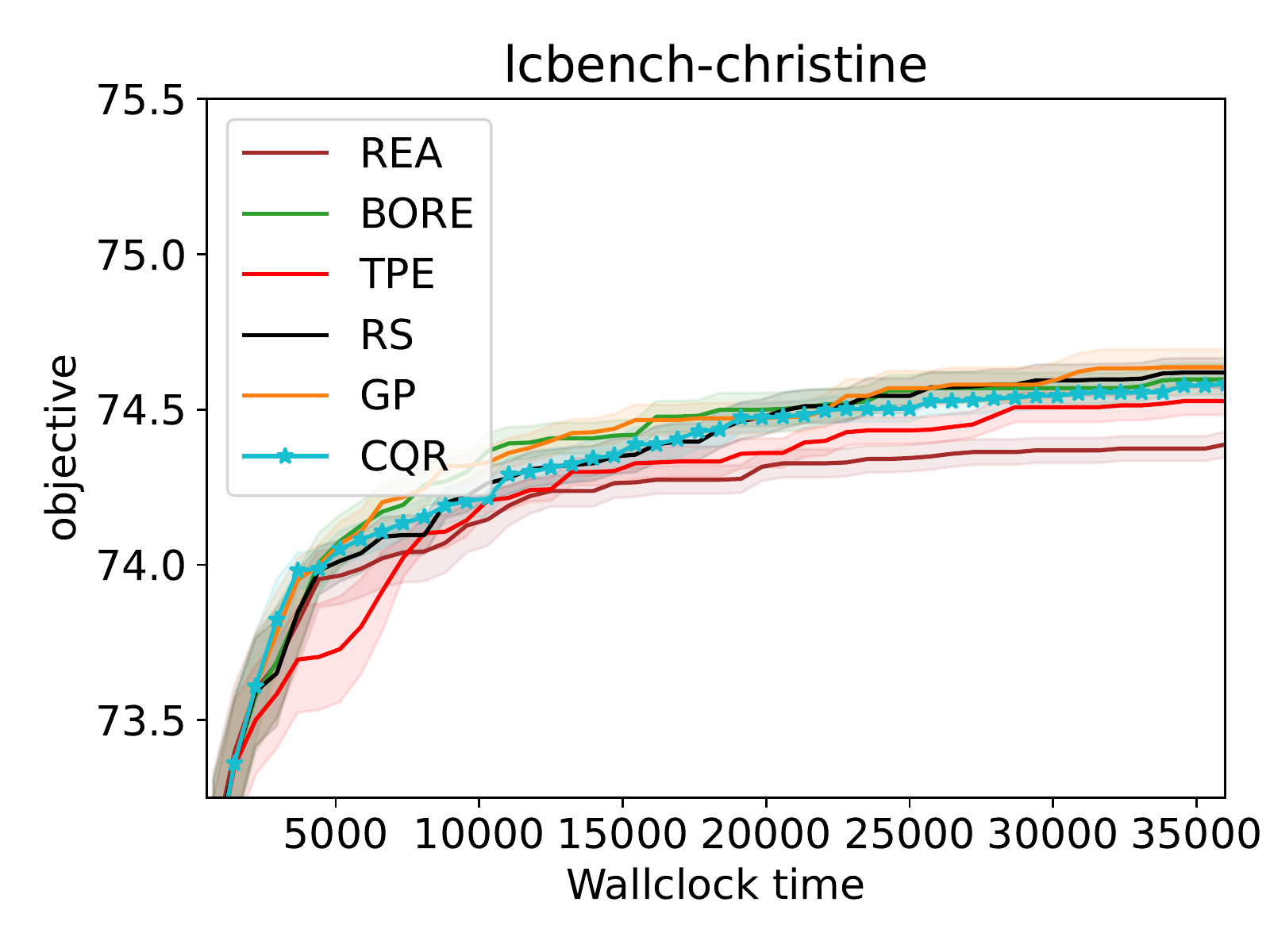}
\includegraphics[width=0.31\textwidth]{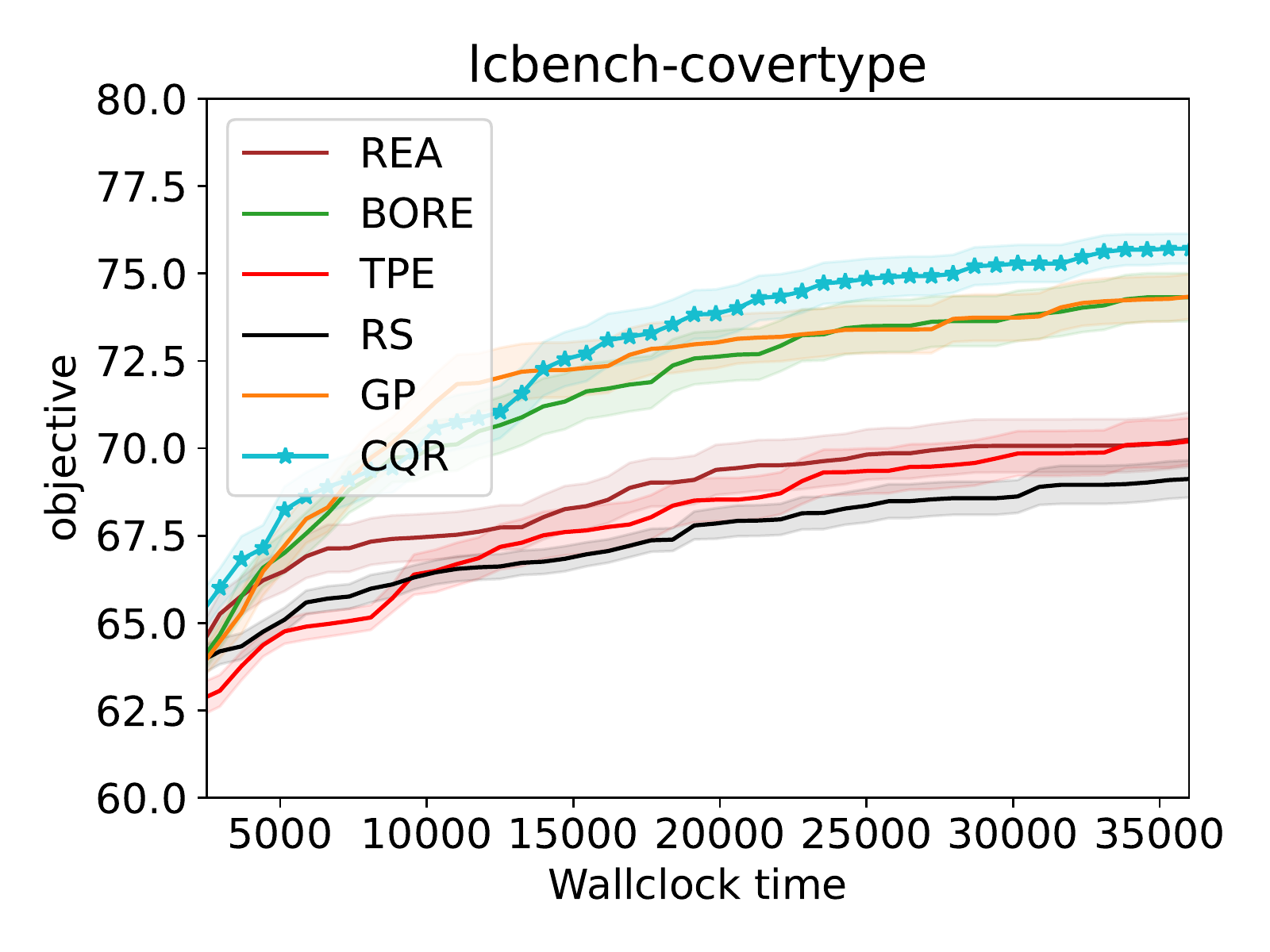} 
\includegraphics[width=0.31\textwidth]{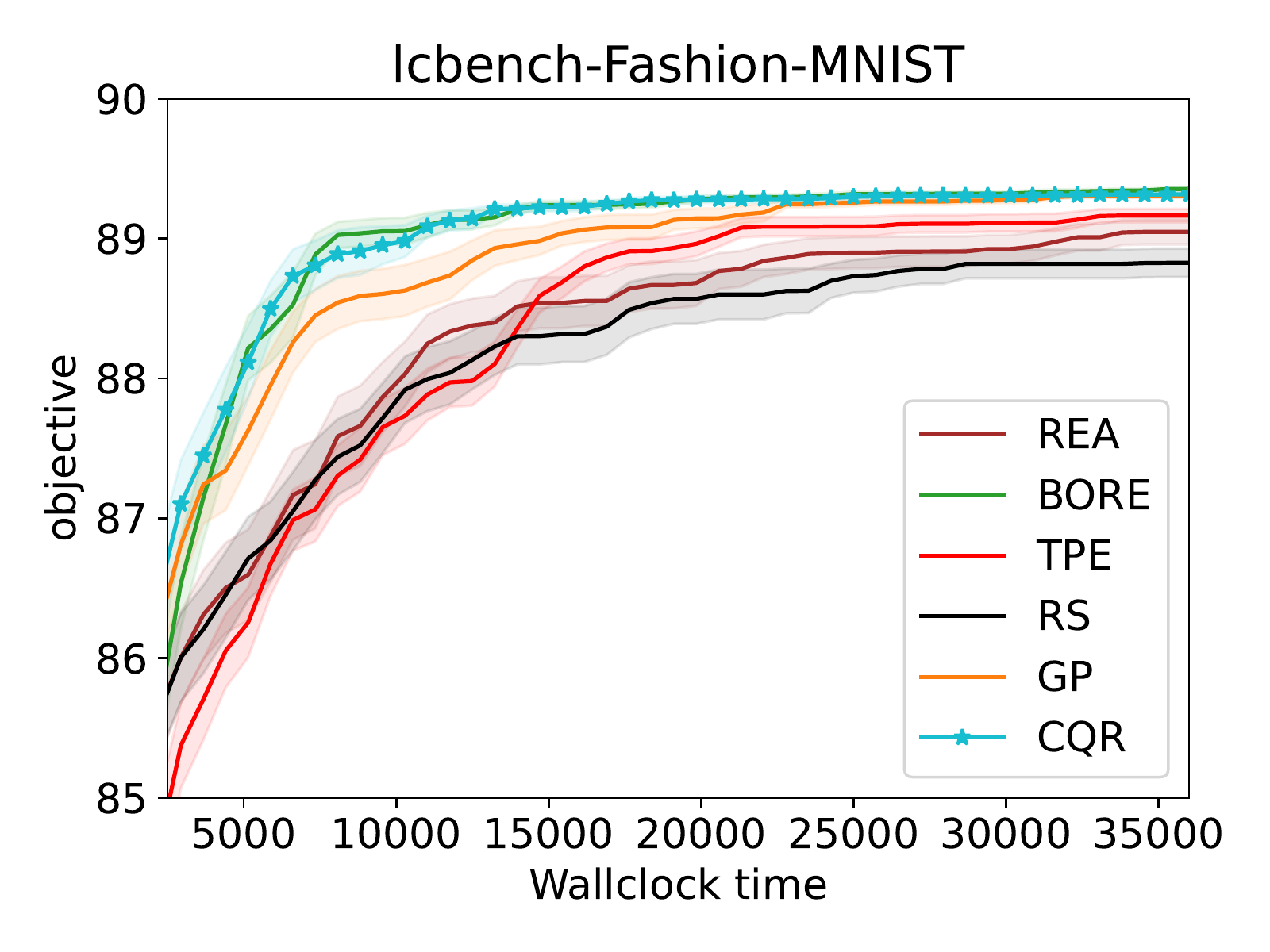} \\
\includegraphics[width=0.31\textwidth]{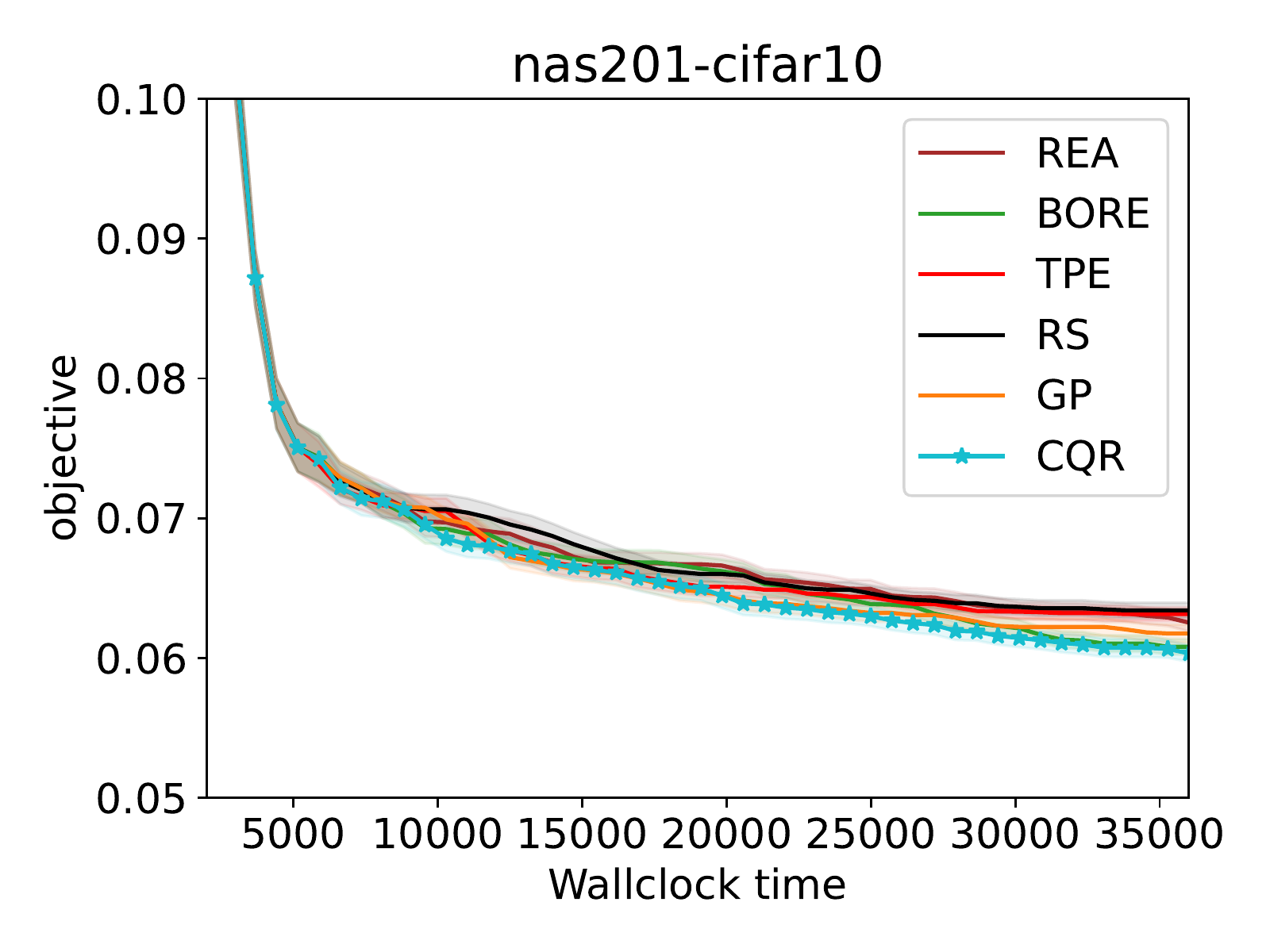}
\includegraphics[width=0.31\textwidth]{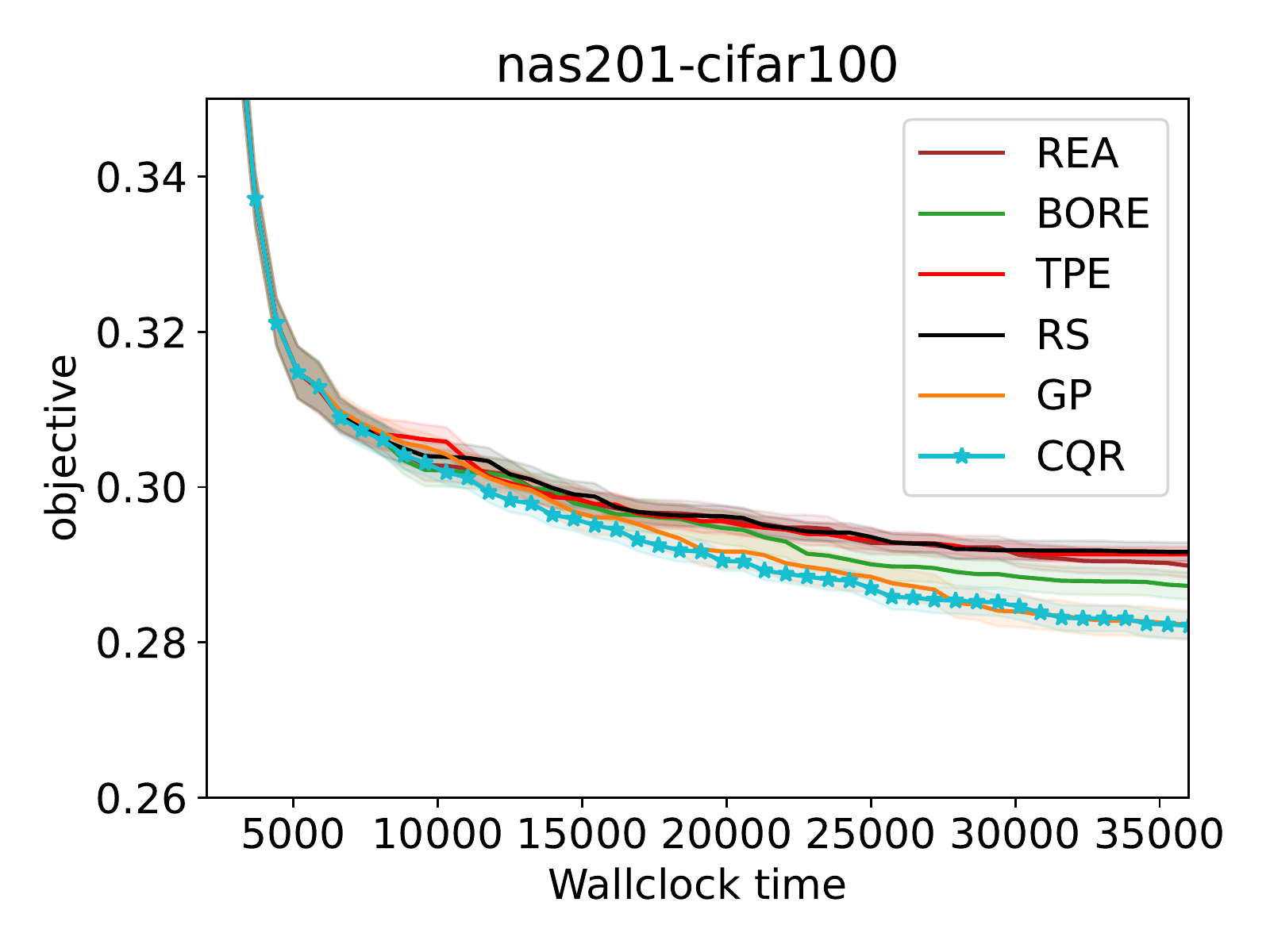}
\includegraphics[width=0.31\textwidth]{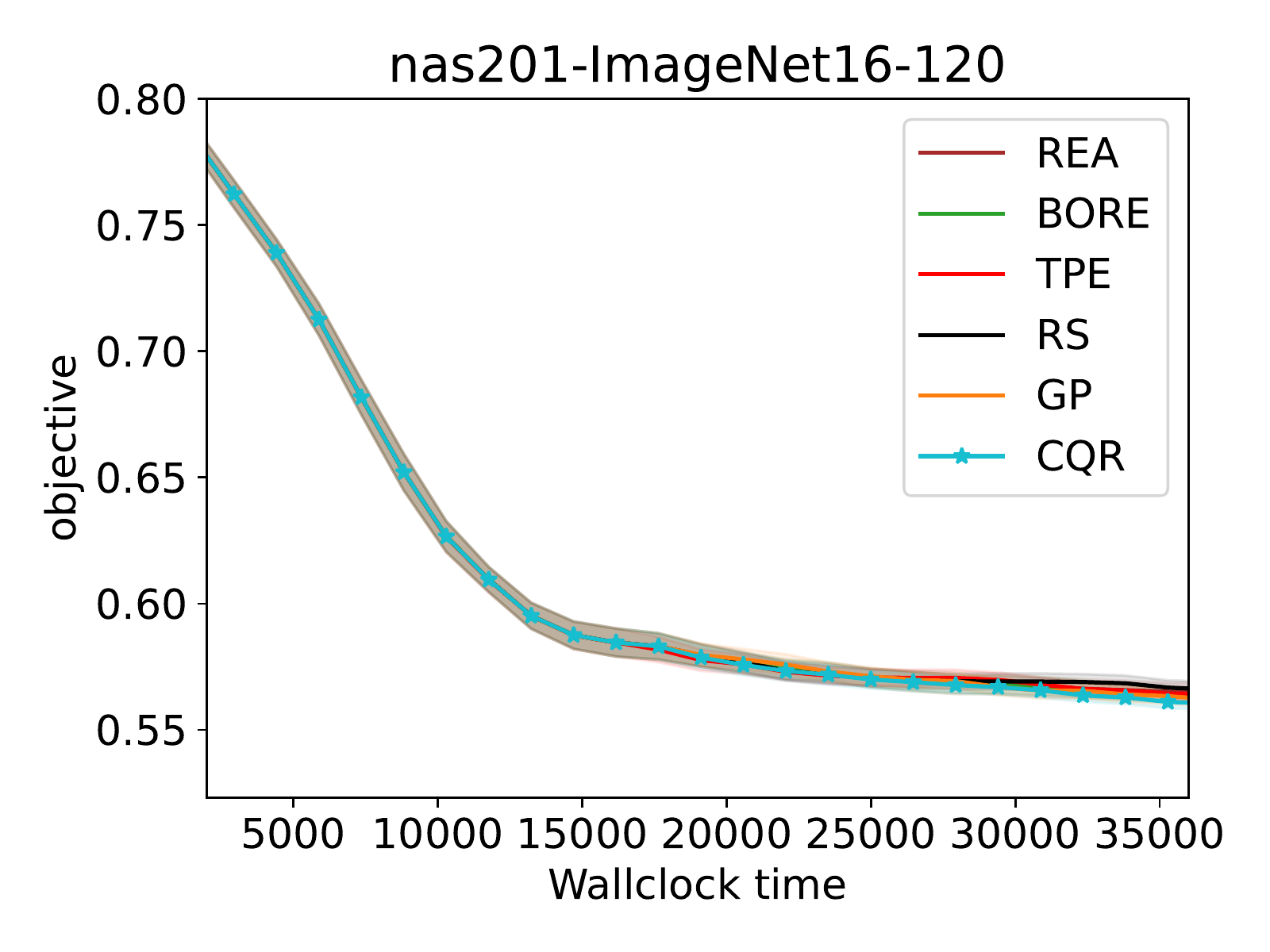} \\
\includegraphics[width=0.31\textwidth]{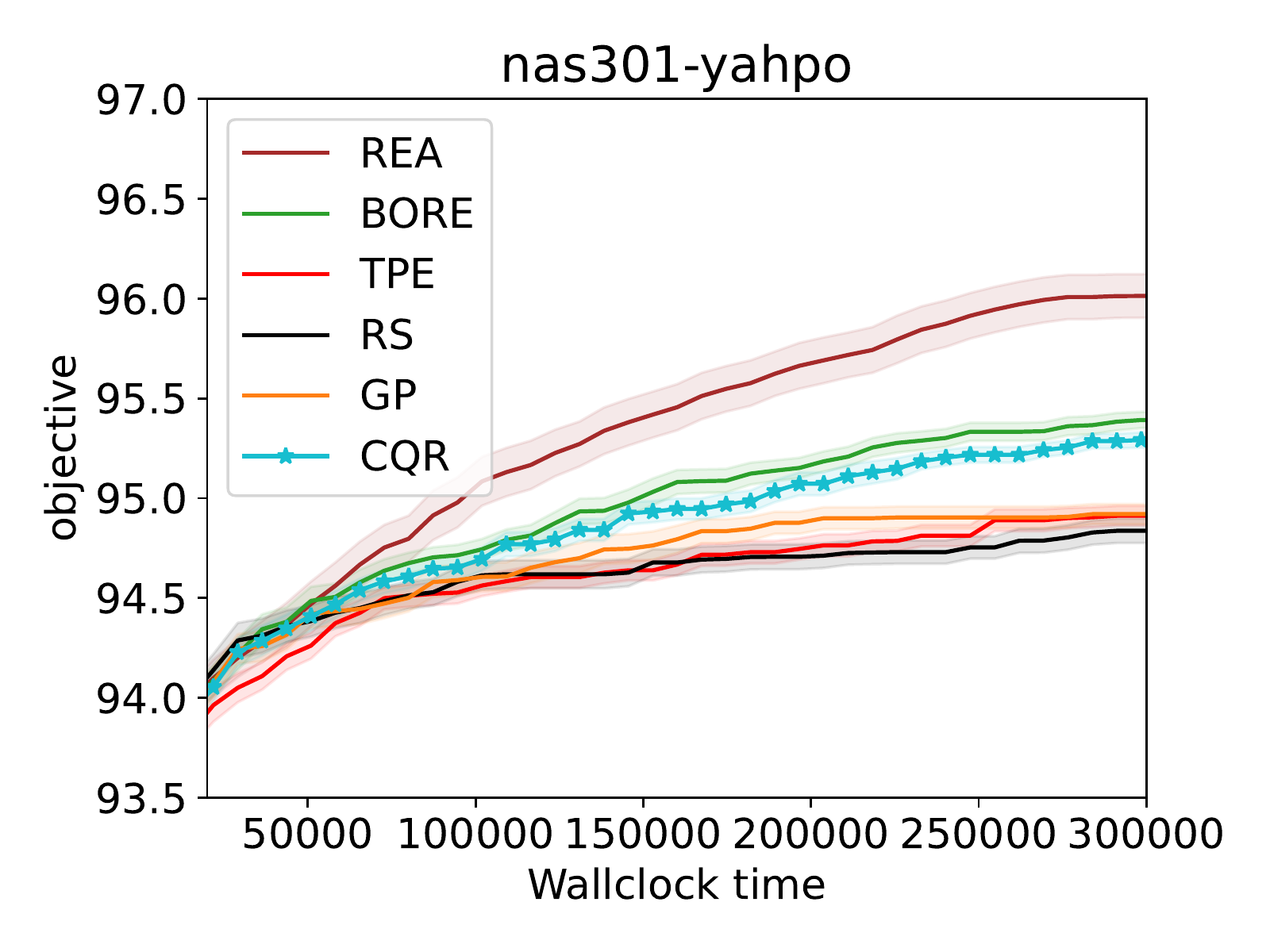}
\caption{Performance of single-fidelity methods over time on all individual tasks considered. Mean and standard errors are computed over \numseed{} seeds. \label{fig:single-fidelity-all-tasks}}
\end{figure*}

\begin{figure*}
\includegraphics[width=0.31\textwidth]{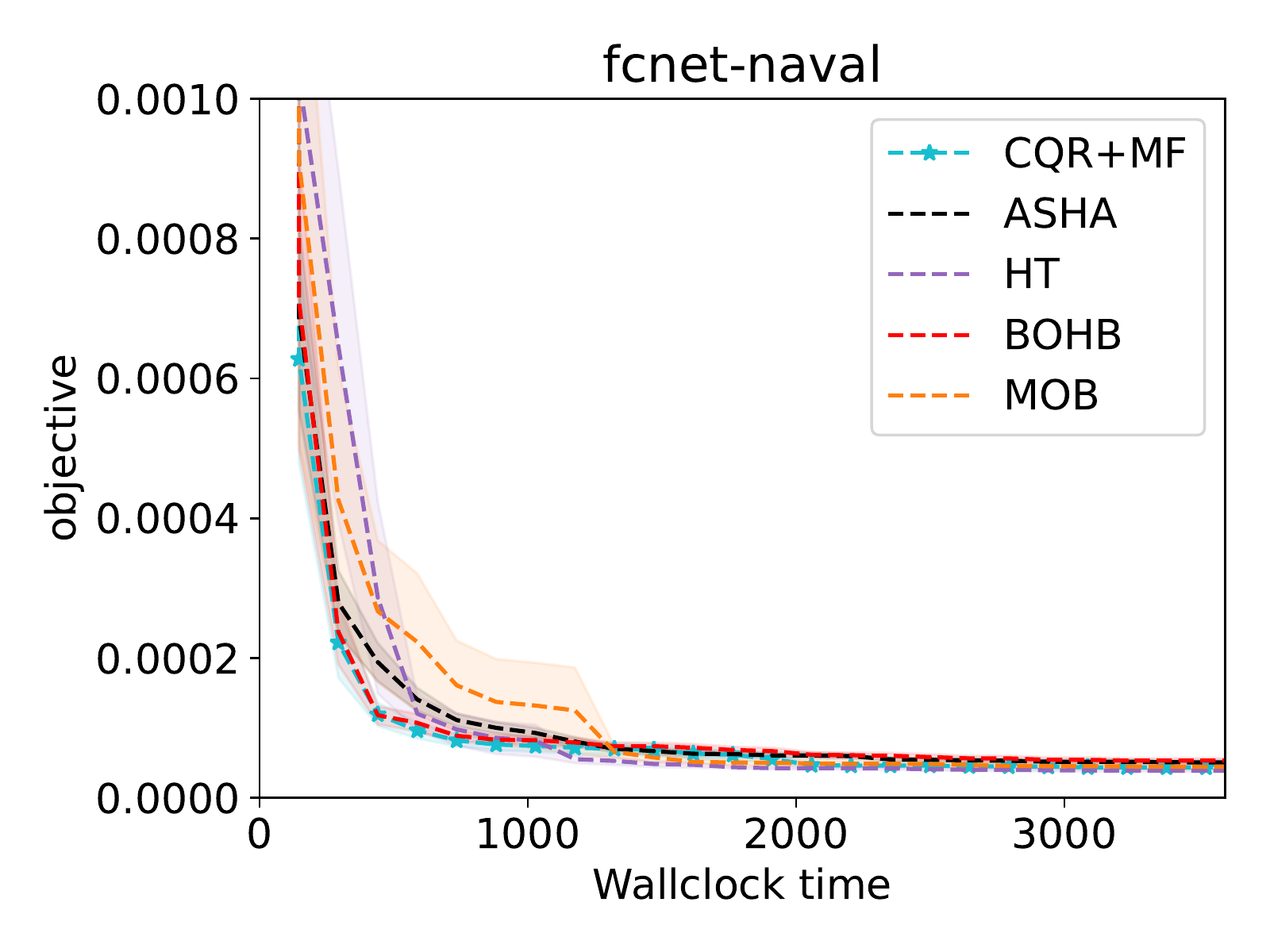}
\includegraphics[width=0.31\textwidth]{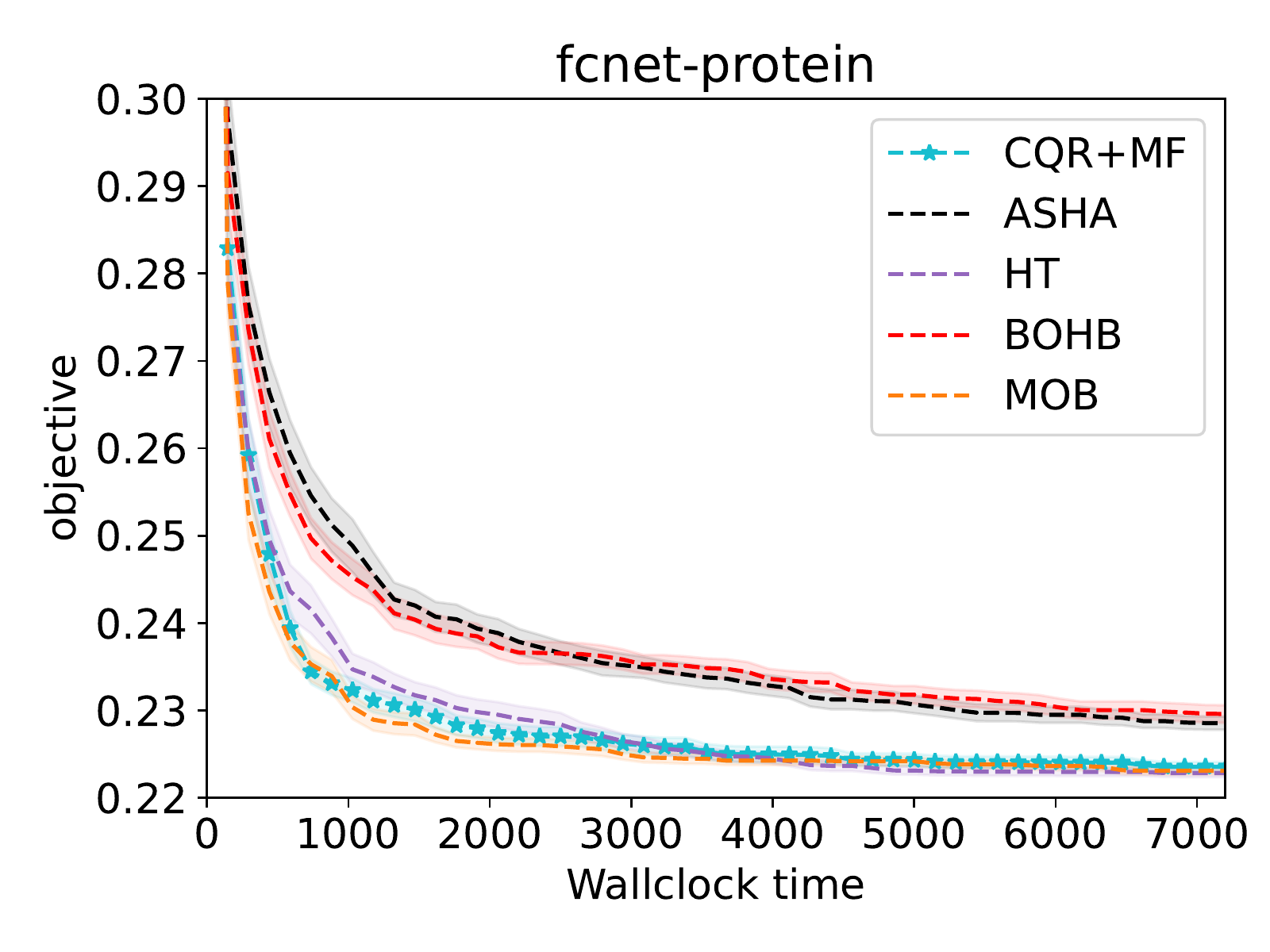}
\includegraphics[width=0.31\textwidth]{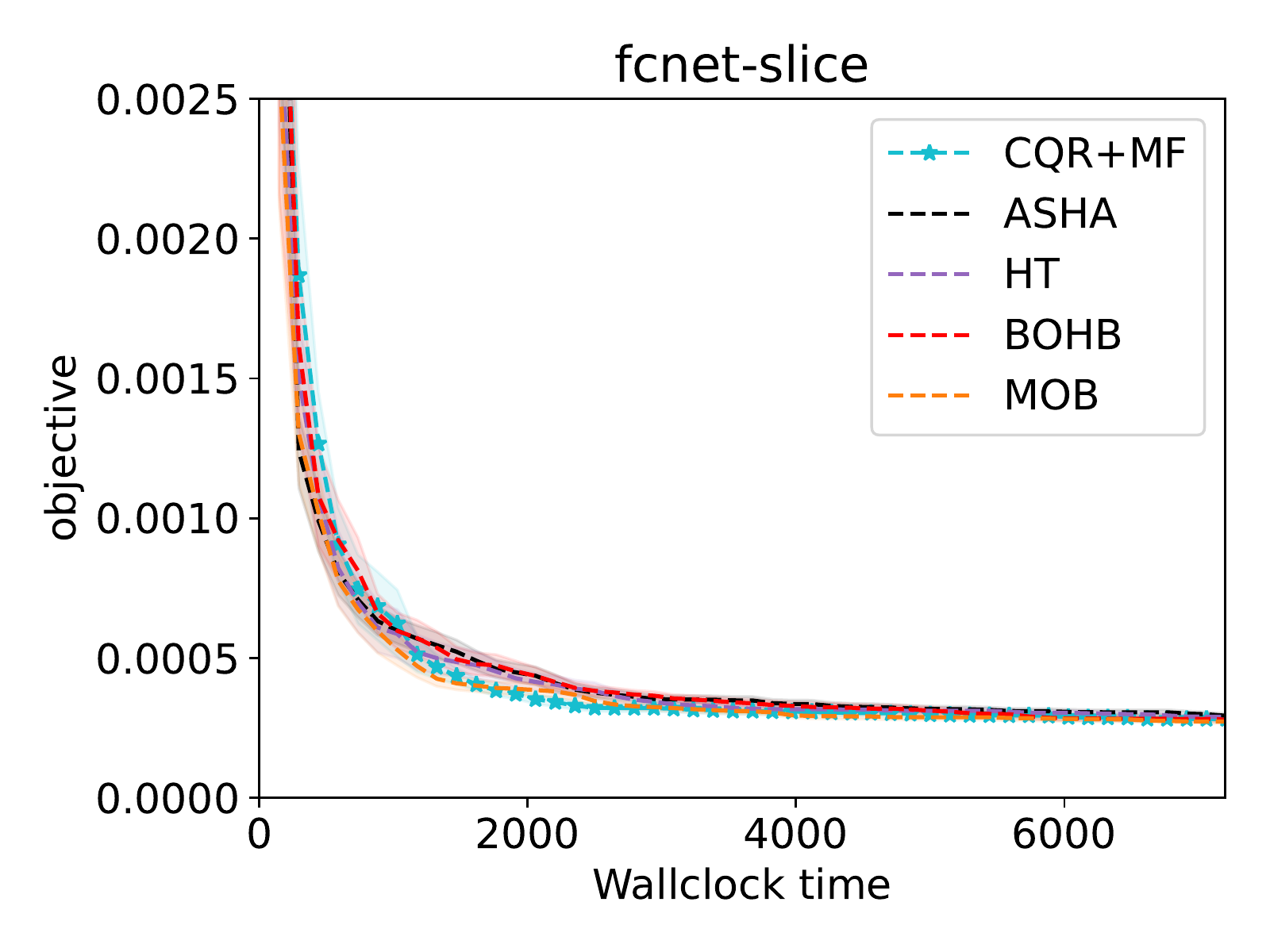} \\
\includegraphics[width=0.31\textwidth]{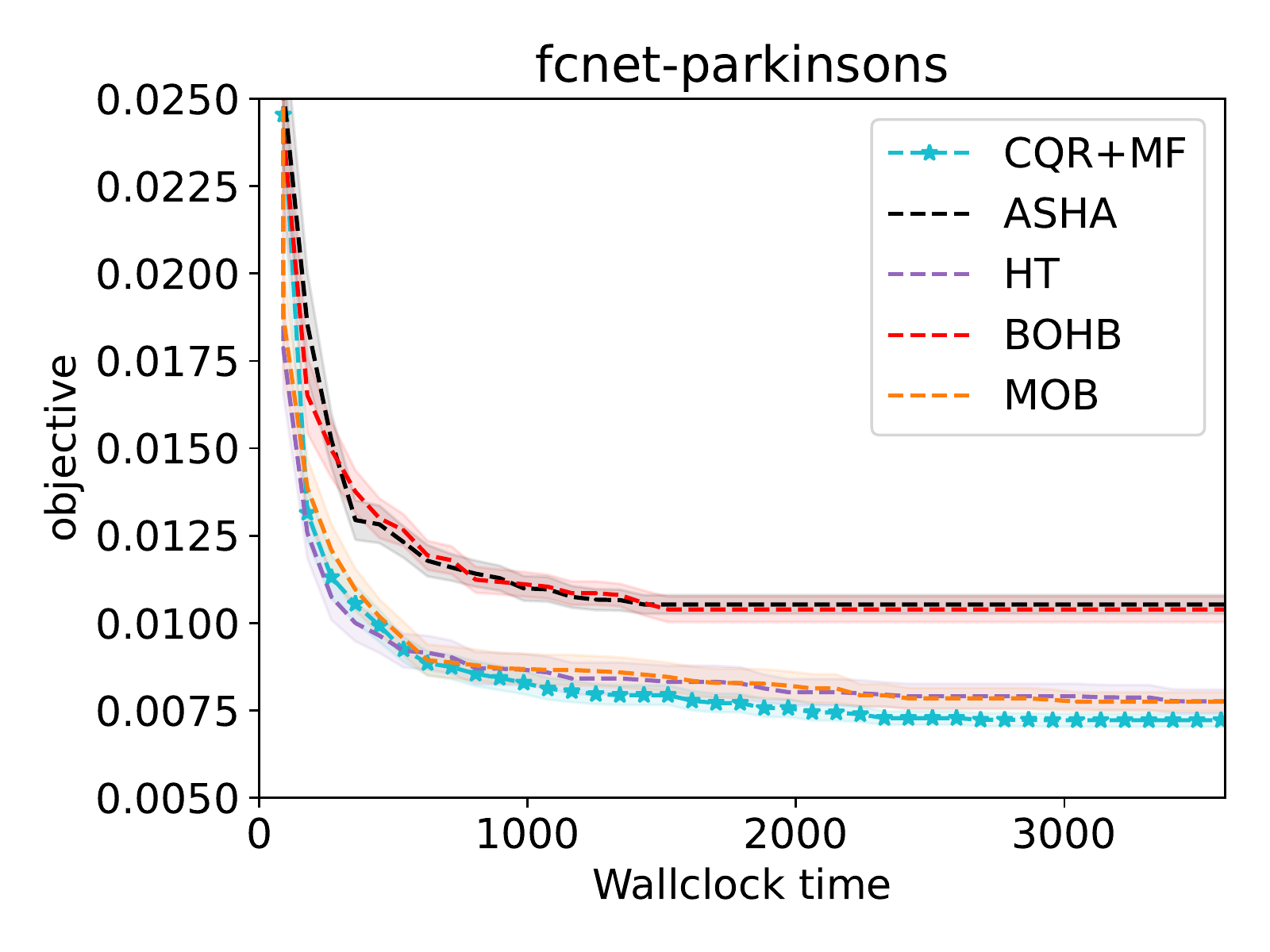} 
\includegraphics[width=0.31\textwidth]{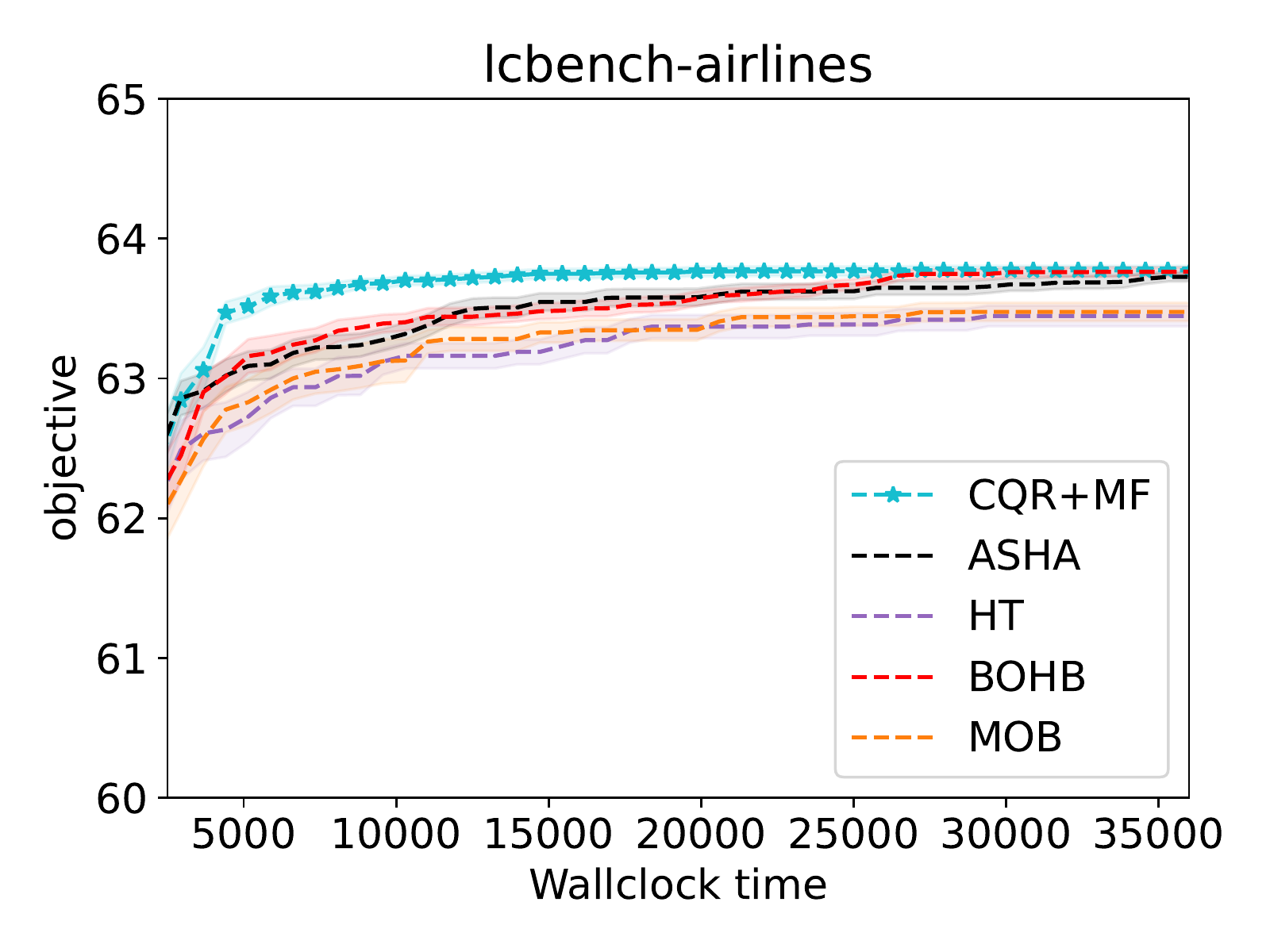}
\includegraphics[width=0.31\textwidth]{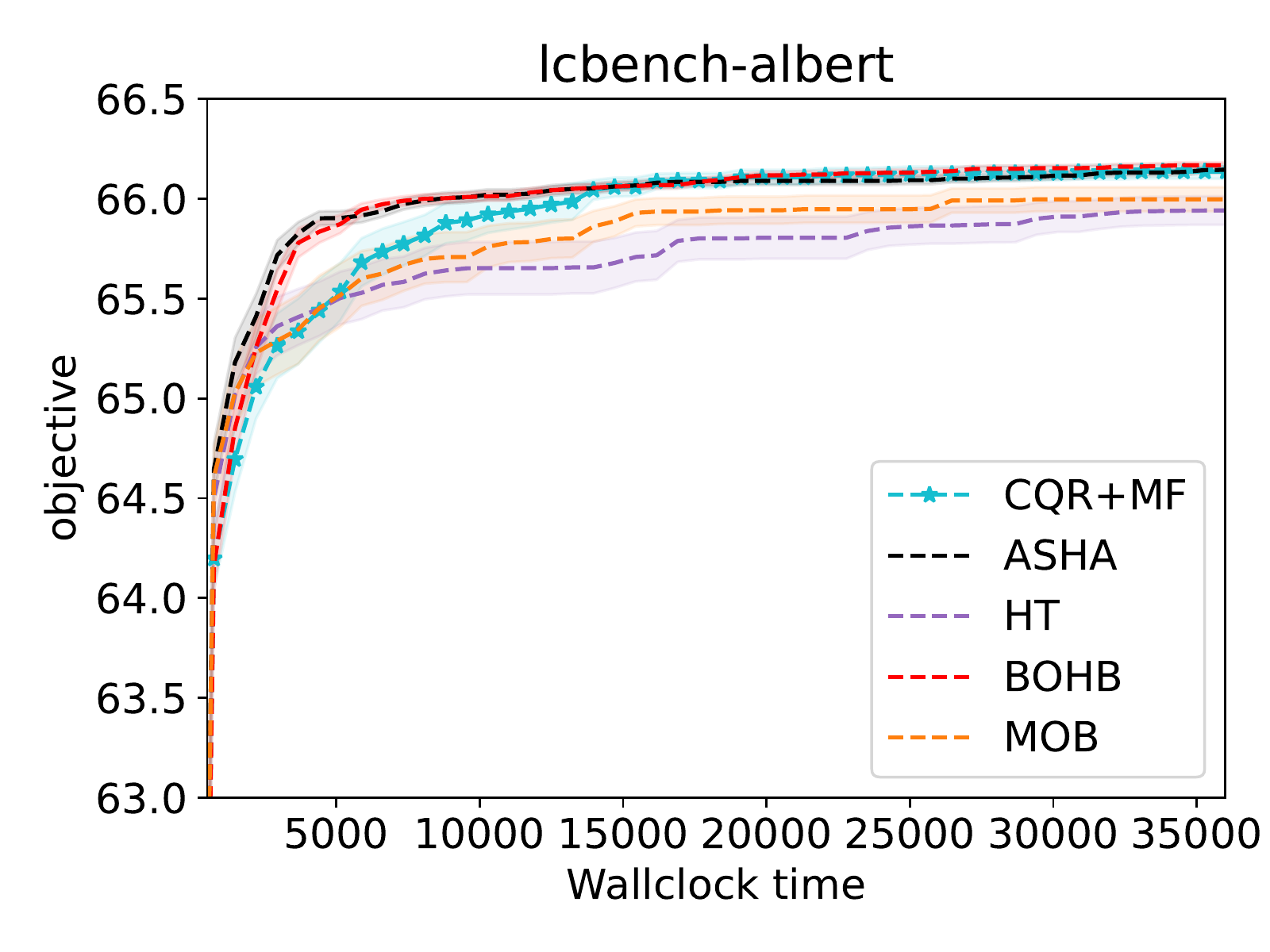} \\
\includegraphics[width=0.31\textwidth]{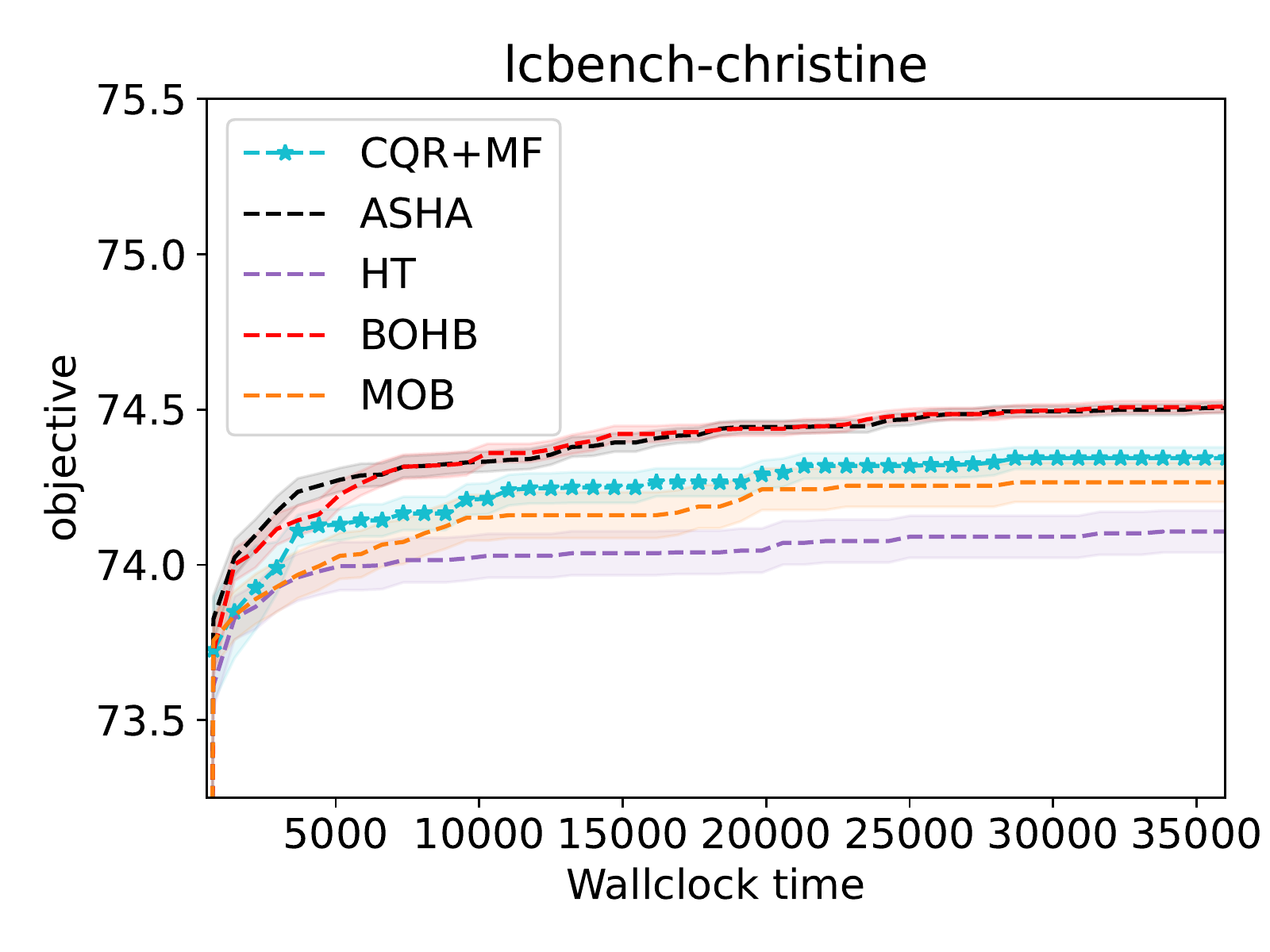}
\includegraphics[width=0.31\textwidth]{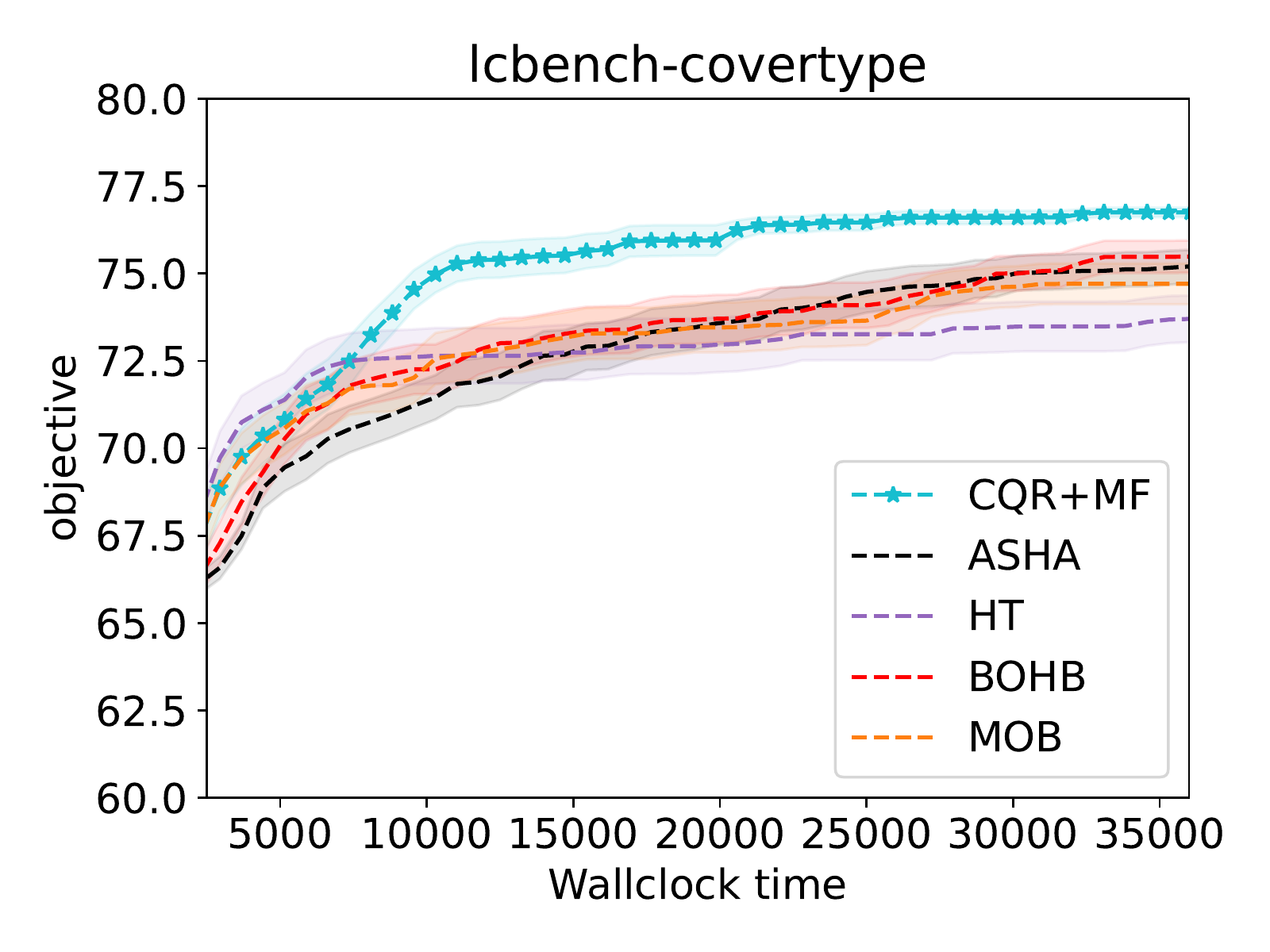} 
\includegraphics[width=0.31\textwidth]{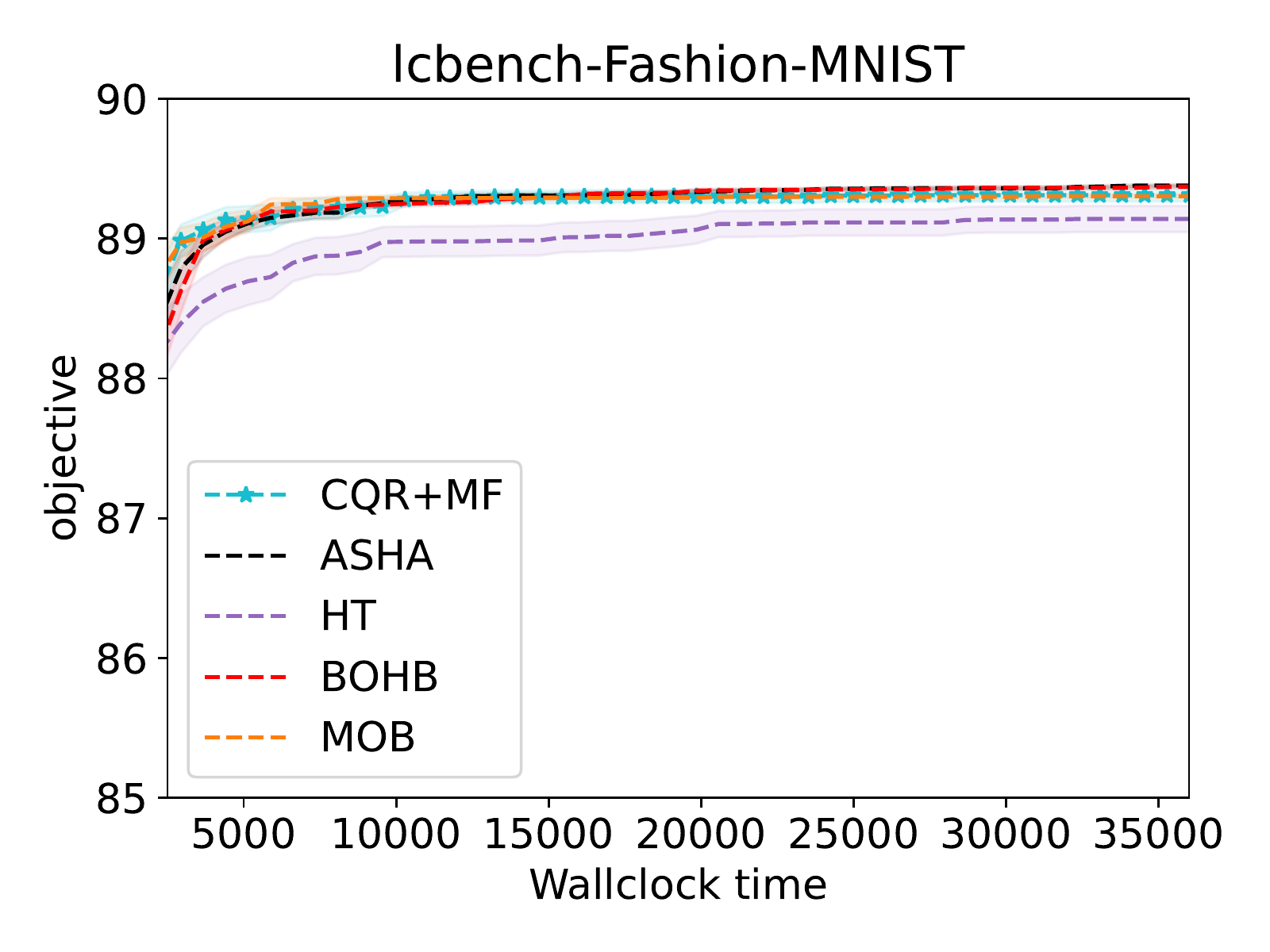} \\
\includegraphics[width=0.31\textwidth]{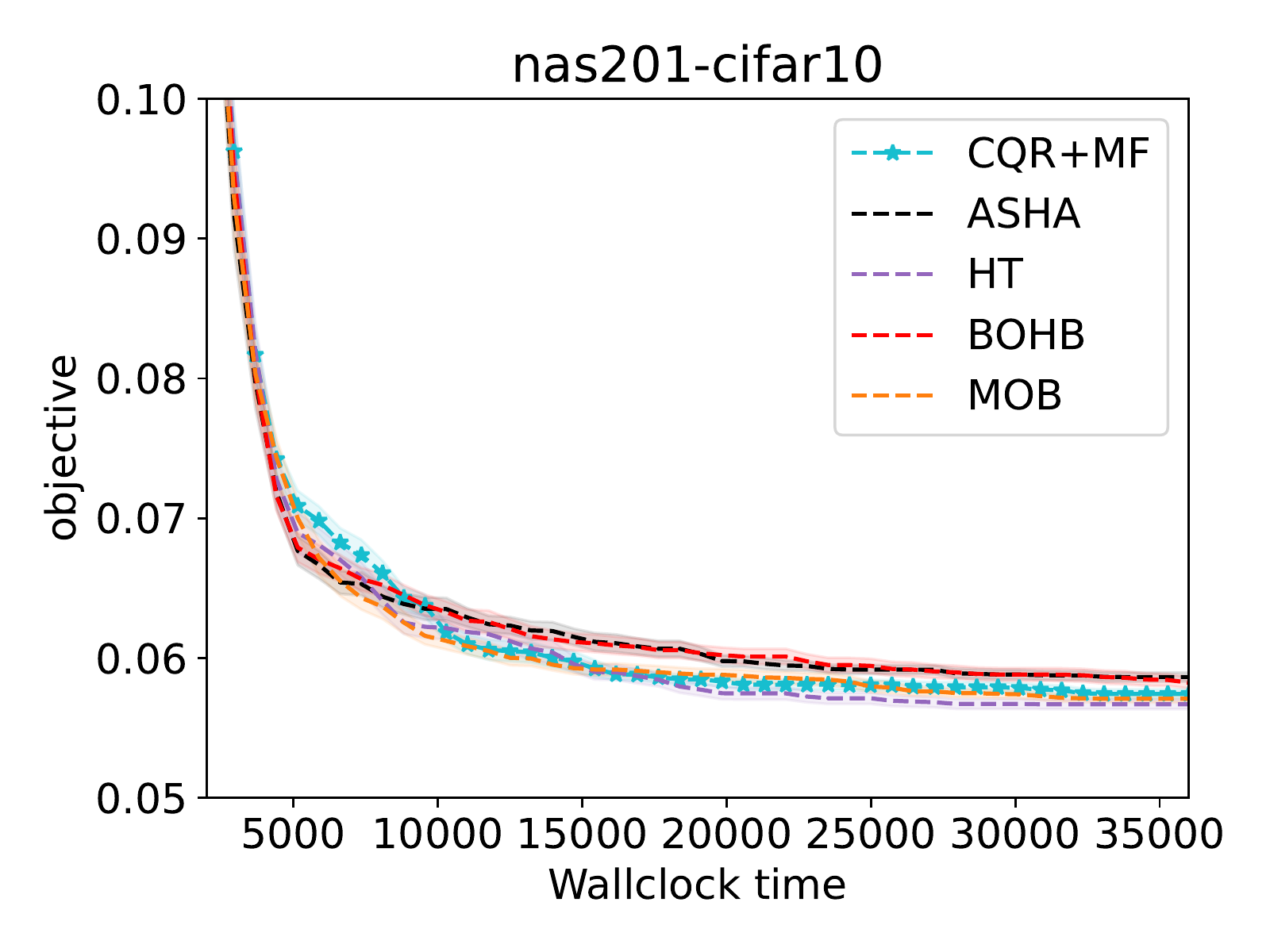}
\includegraphics[width=0.31\textwidth]{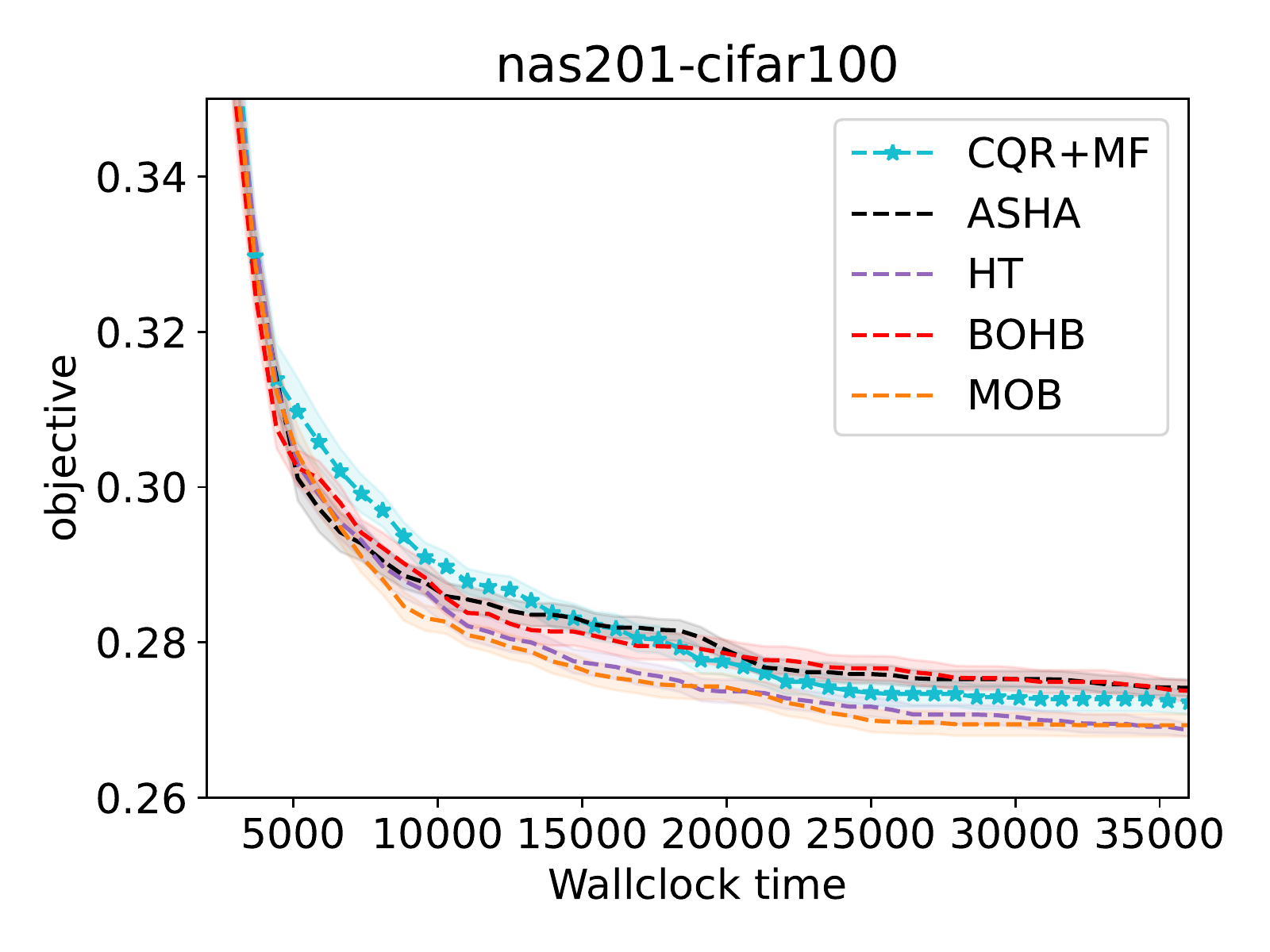}
\includegraphics[width=0.31\textwidth]{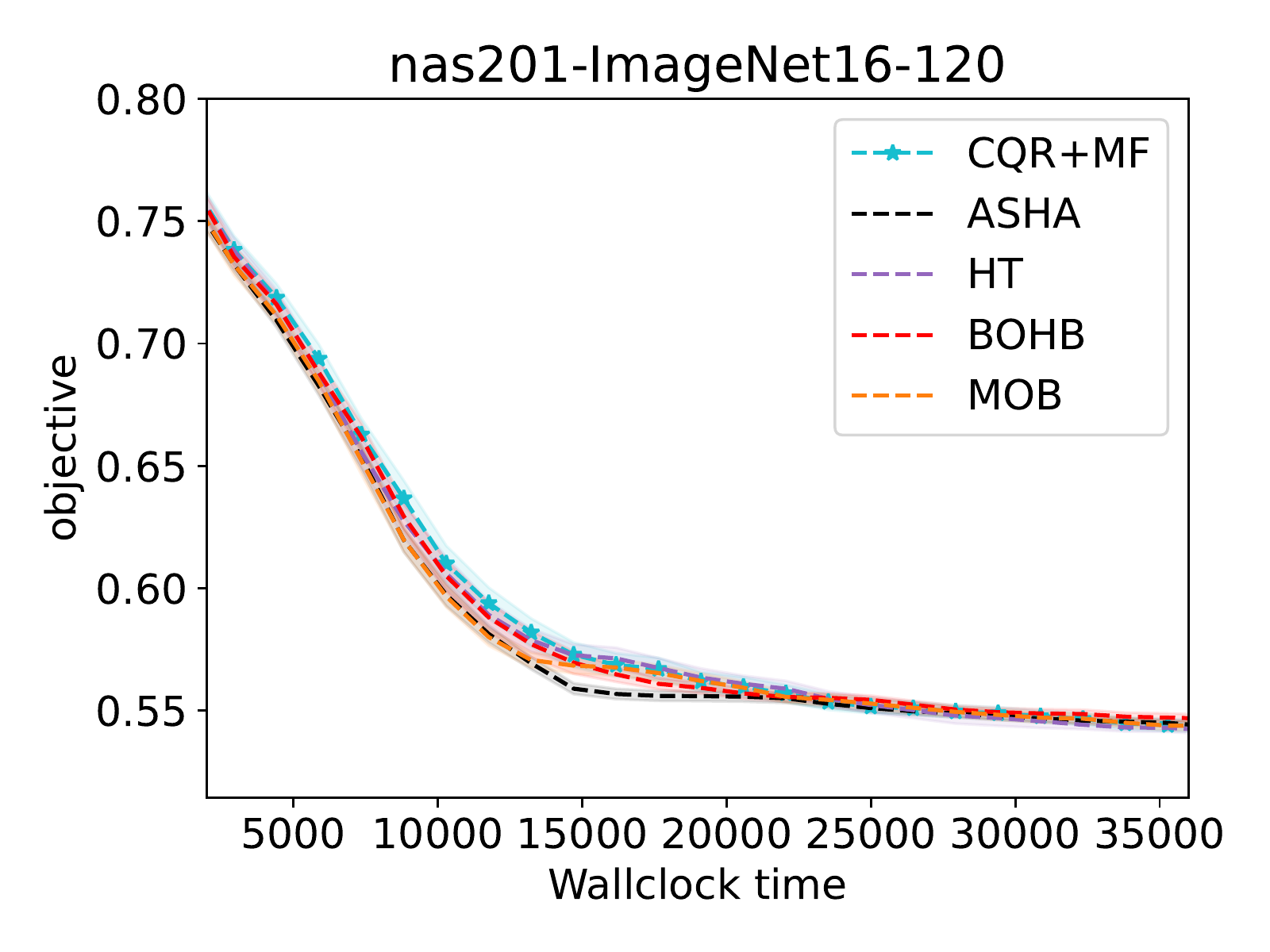} \\
\includegraphics[width=0.31\textwidth]{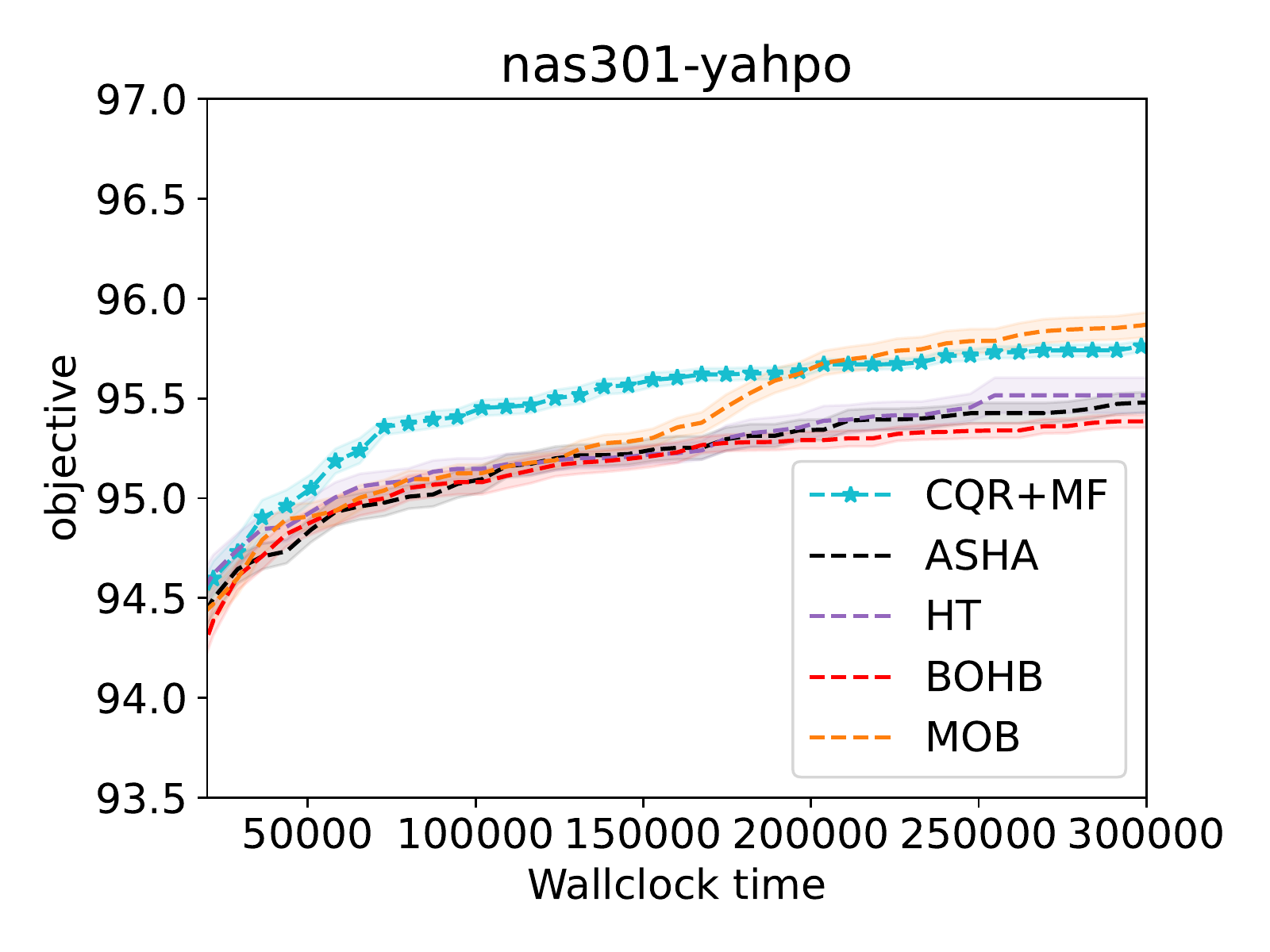}
\caption{Performance of multi-fidelity methods over time on all individual tasks considered. Mean and standard errors are computed over \numseed{} seeds. \label{fig:multi-fidelity-all-tasks}}
\end{figure*}

We plot the performance of methods on all 13 tasks for single-fidelity methods in Figure \ref{fig:single-fidelity-all-tasks} and \ref{fig:multi-fidelity-all-tasks} for multi-fidelity methods.

\begin{figure*}
\includegraphics[width=0.31\textwidth]{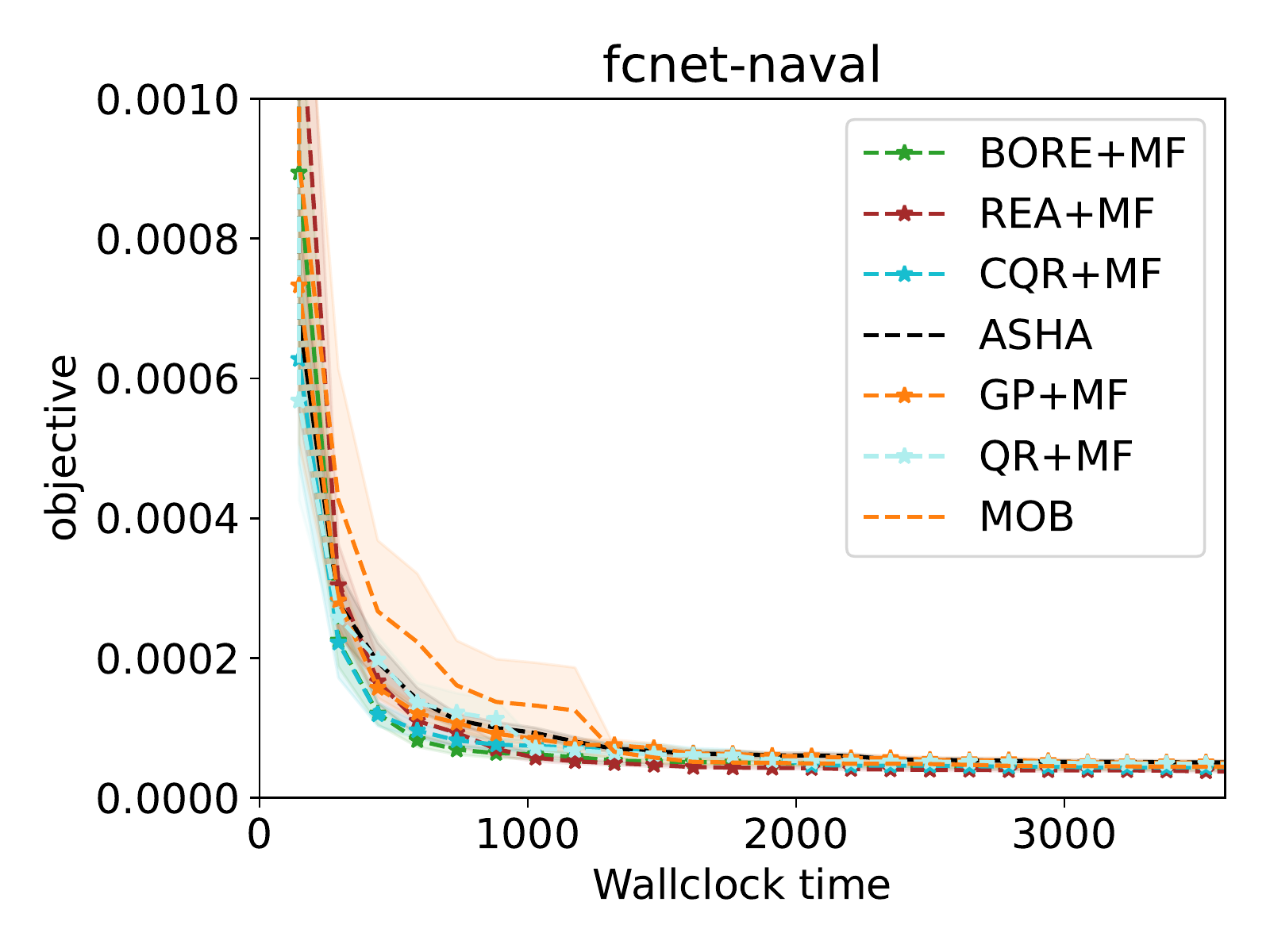}
\includegraphics[width=0.31\textwidth]{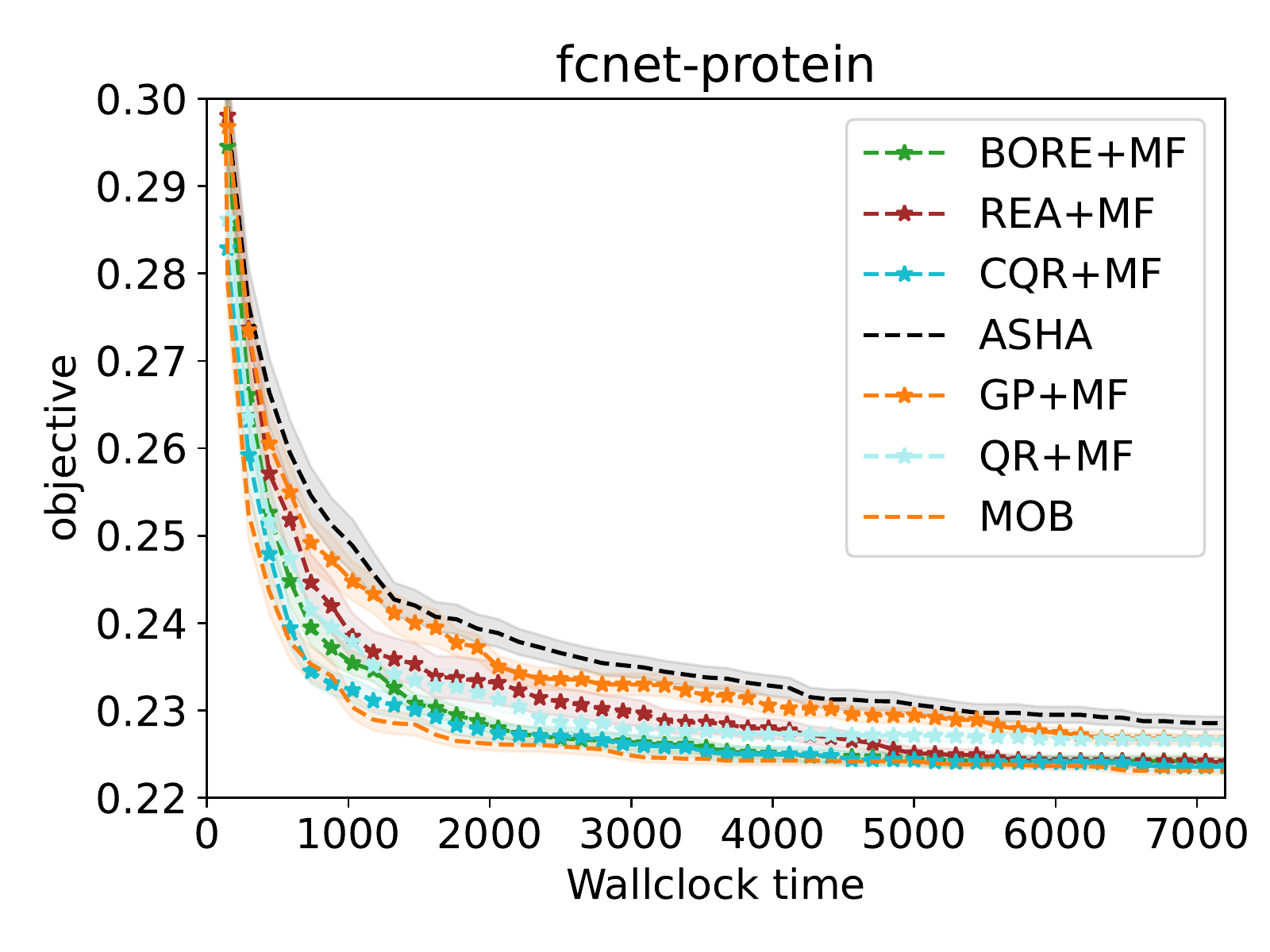}
\includegraphics[width=0.31\textwidth]{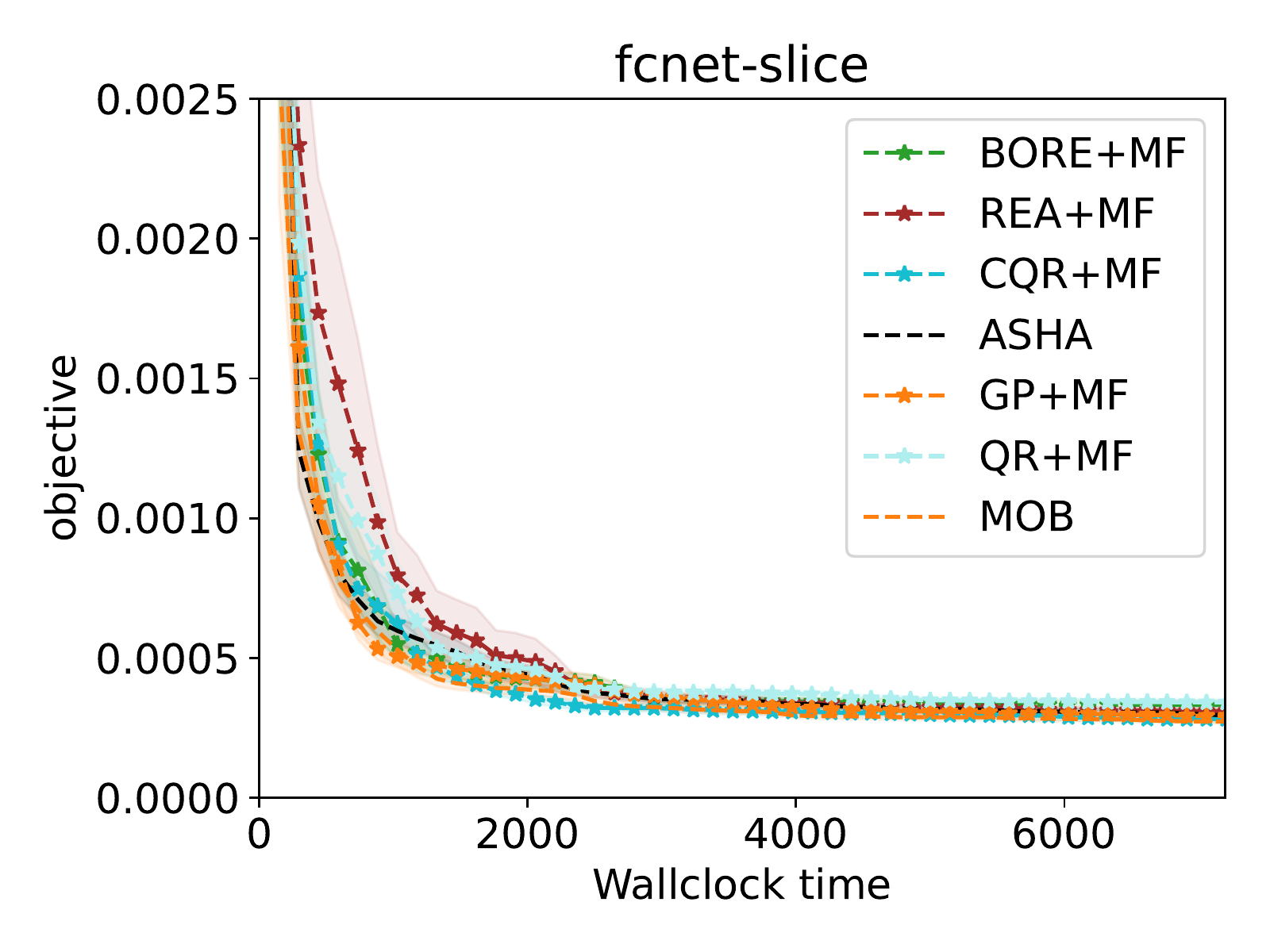} \\
\includegraphics[width=0.31\textwidth]{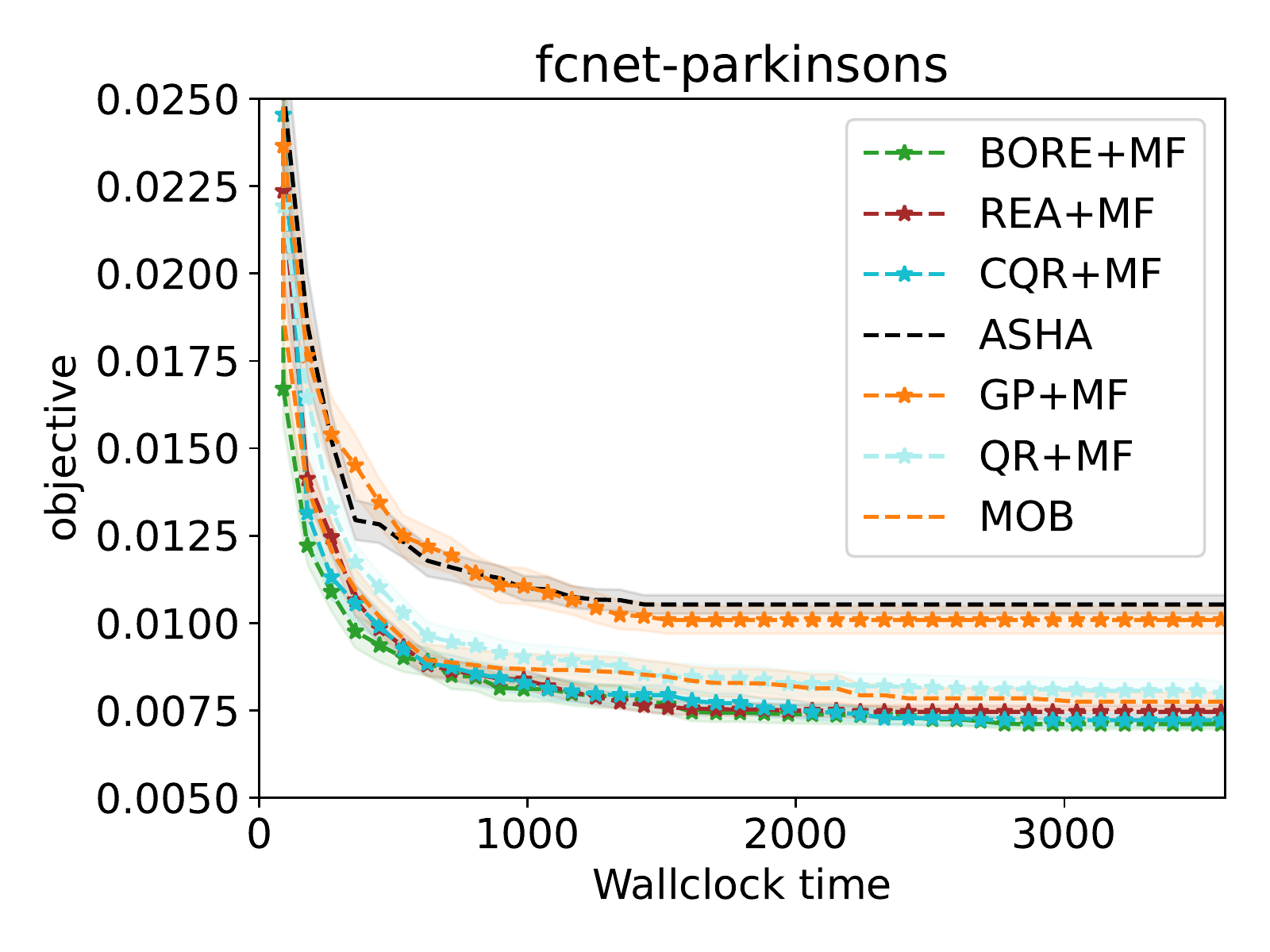} 
\includegraphics[width=0.31\textwidth]{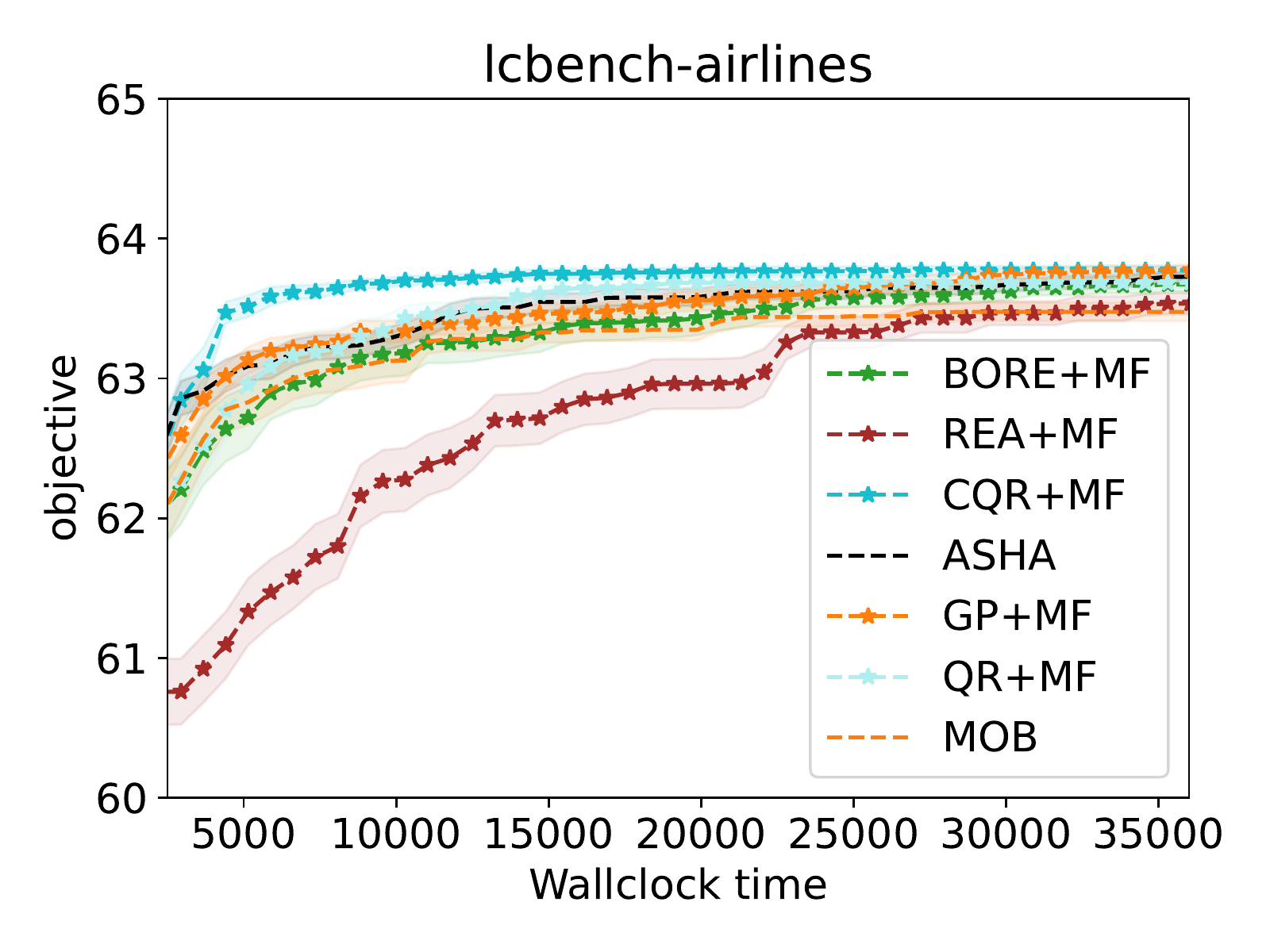}
\includegraphics[width=0.31\textwidth]{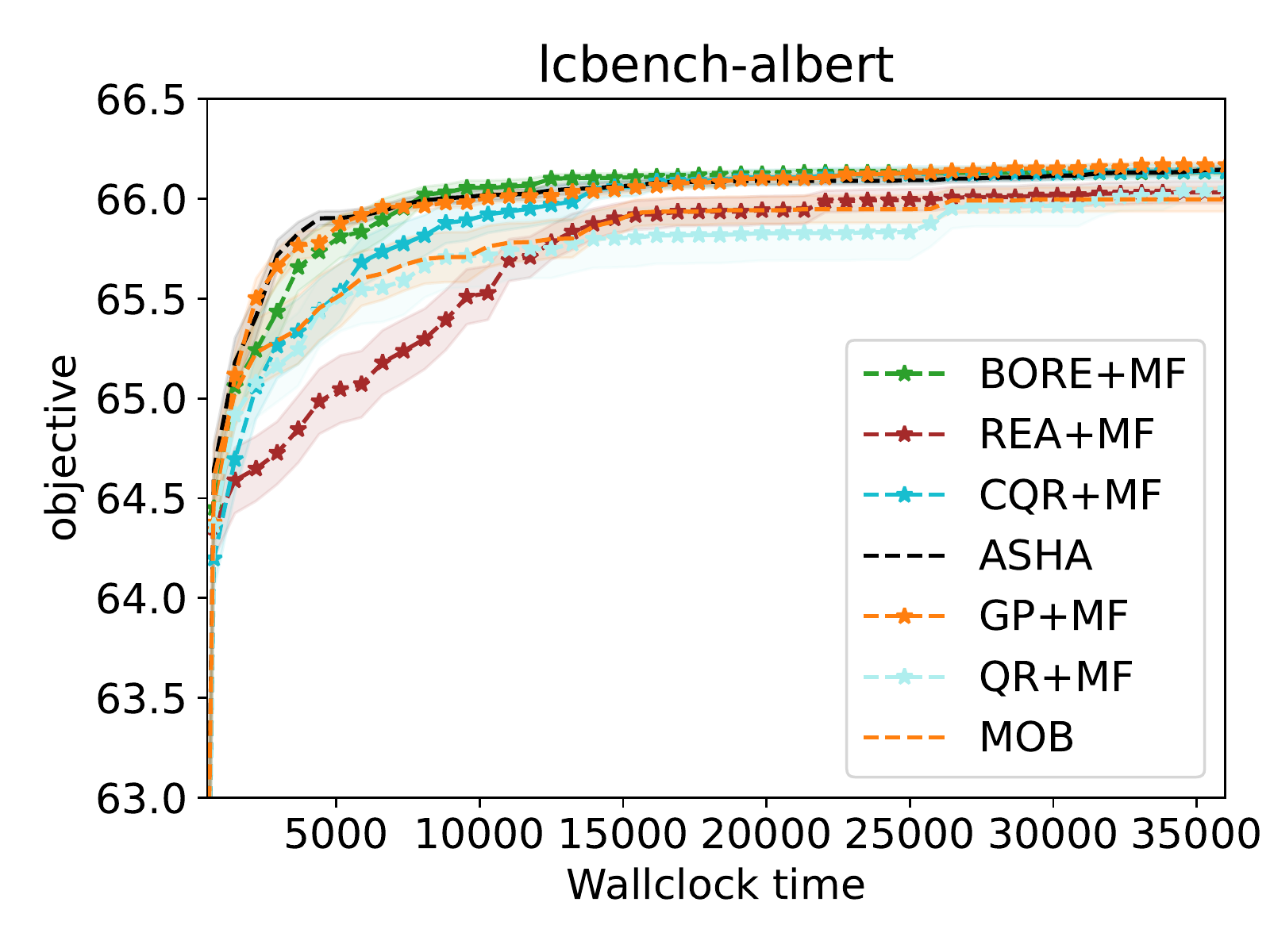} \\
\includegraphics[width=0.31\textwidth]{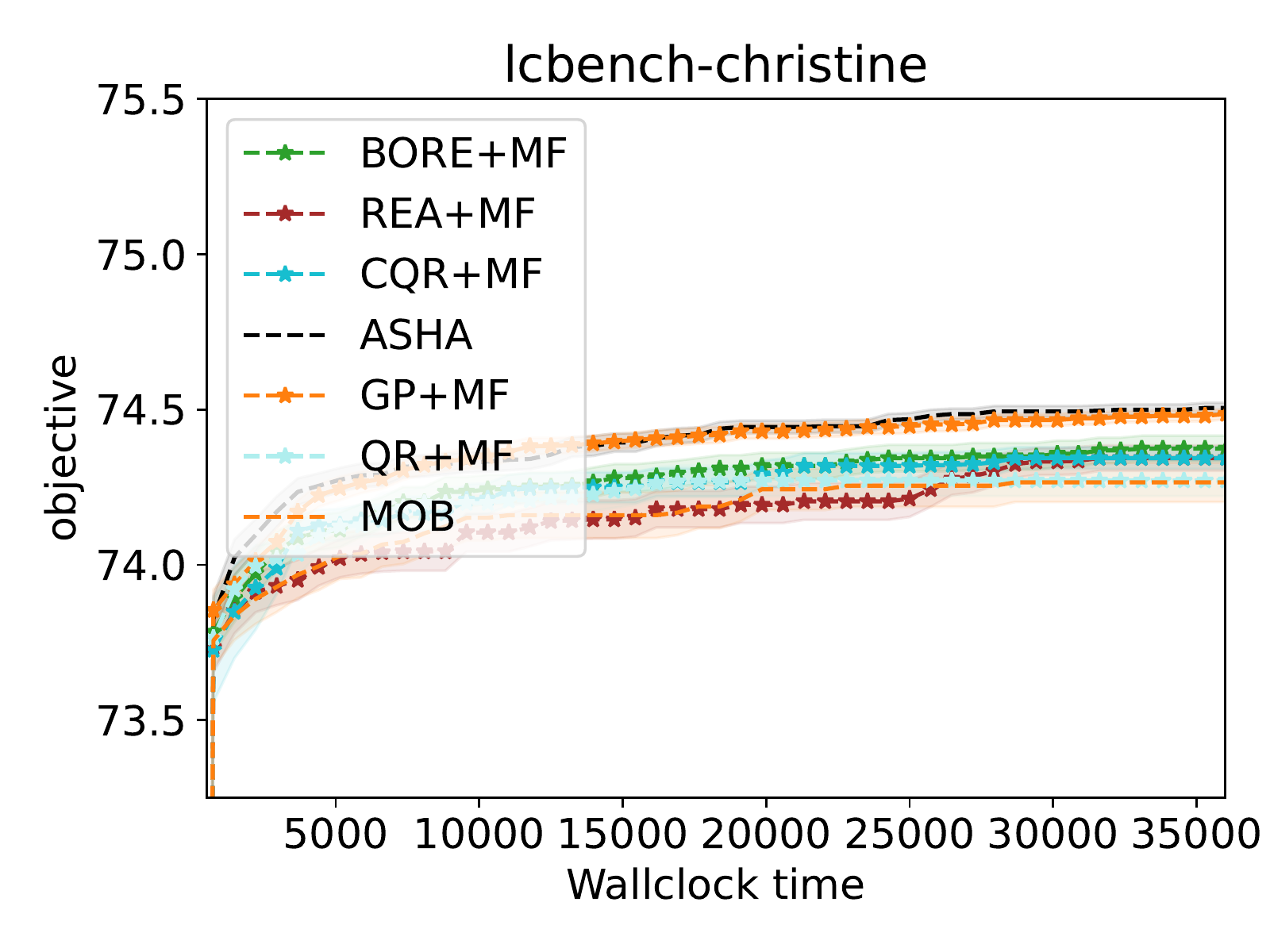}
\includegraphics[width=0.31\textwidth]{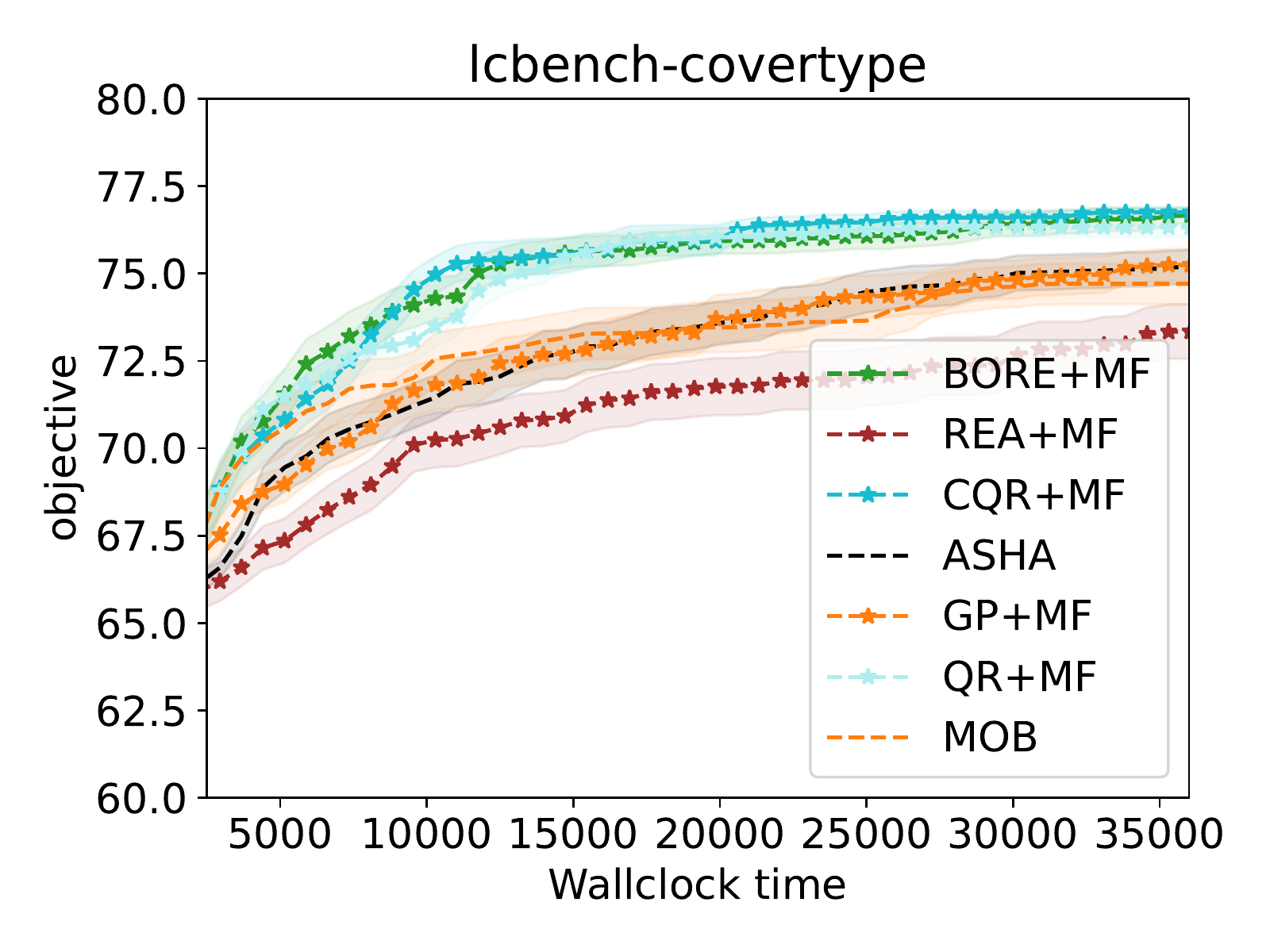} 
\includegraphics[width=0.31\textwidth]{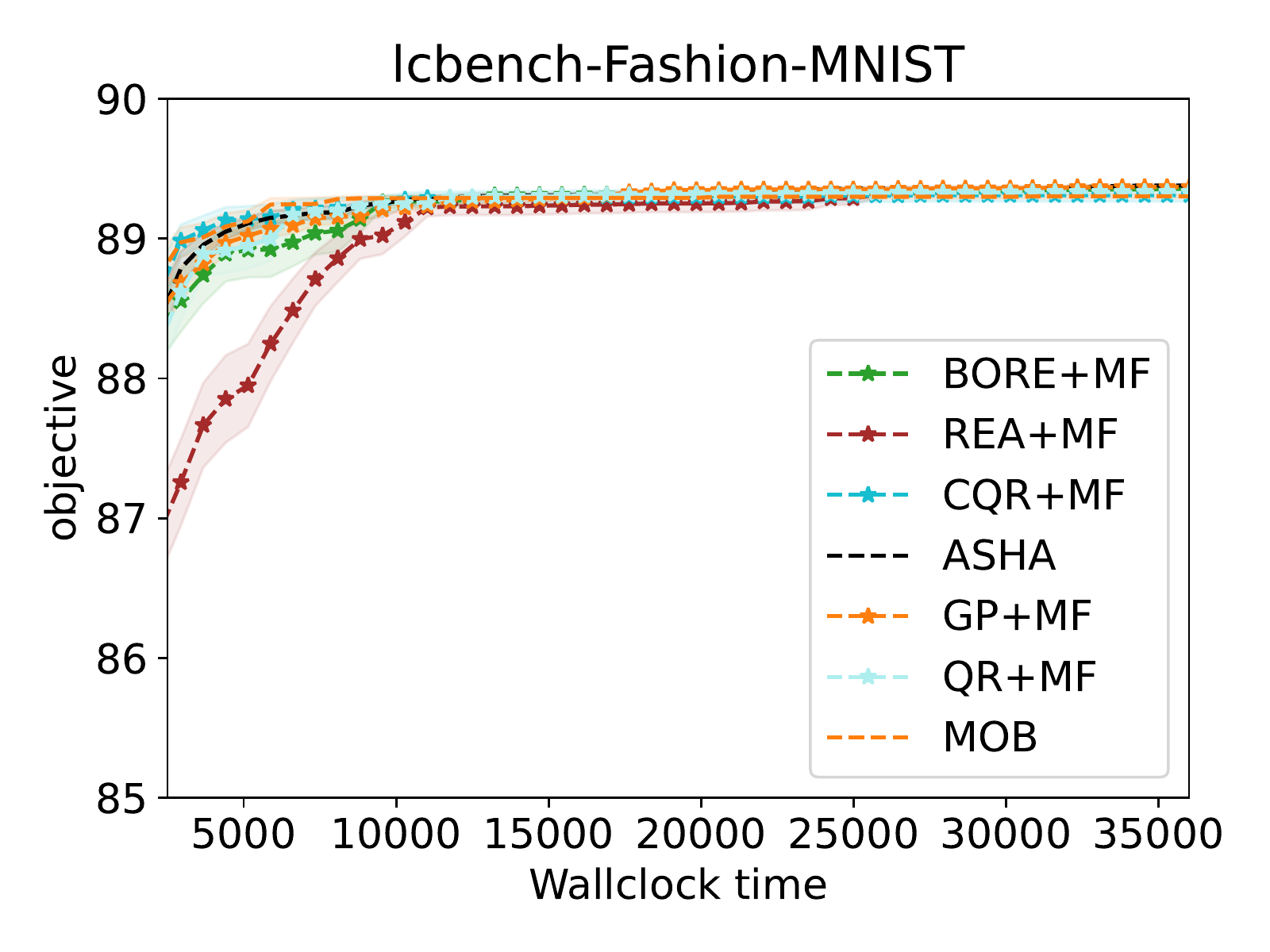} \\
\includegraphics[width=0.31\textwidth]{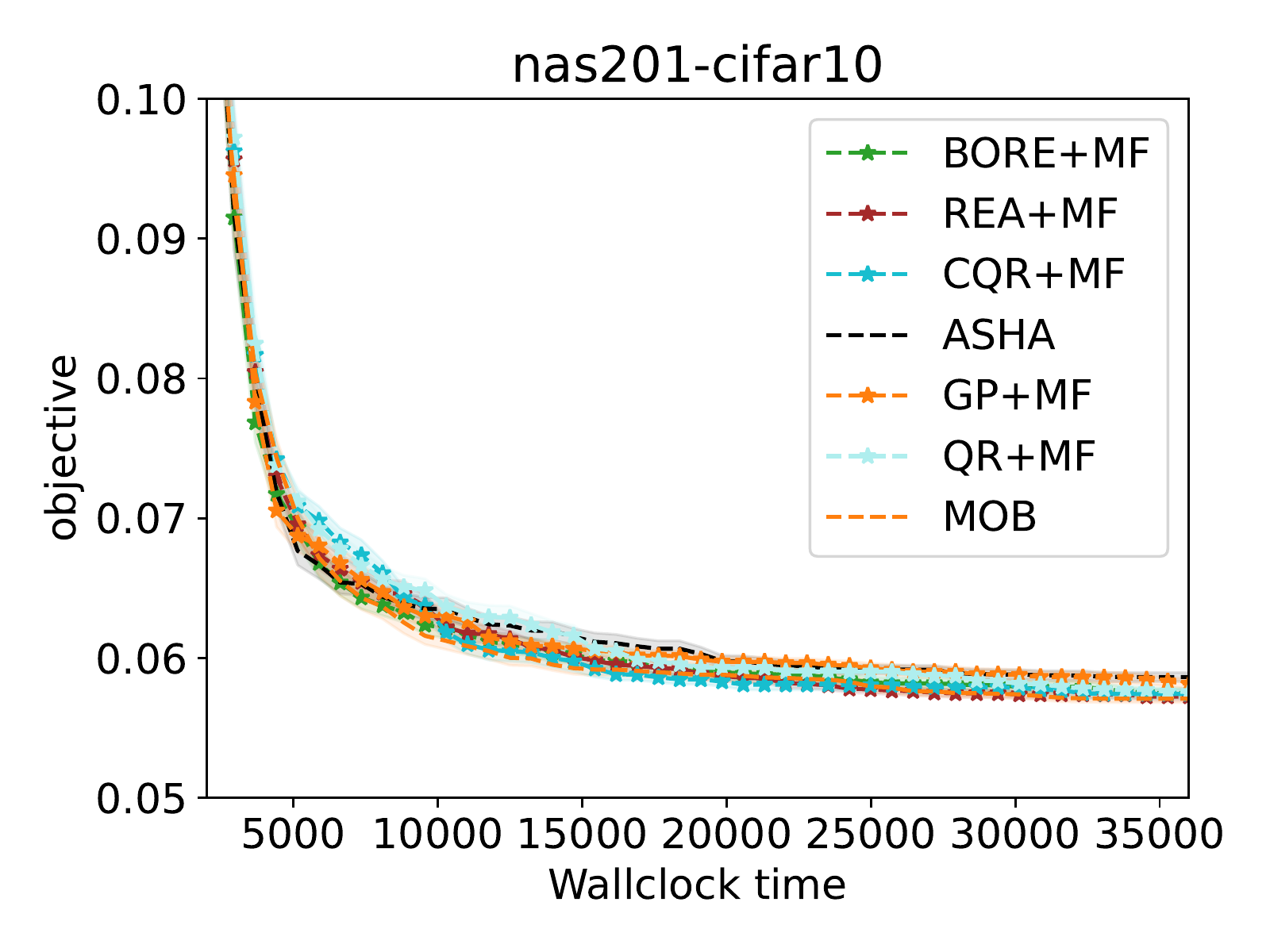}
\includegraphics[width=0.31\textwidth]{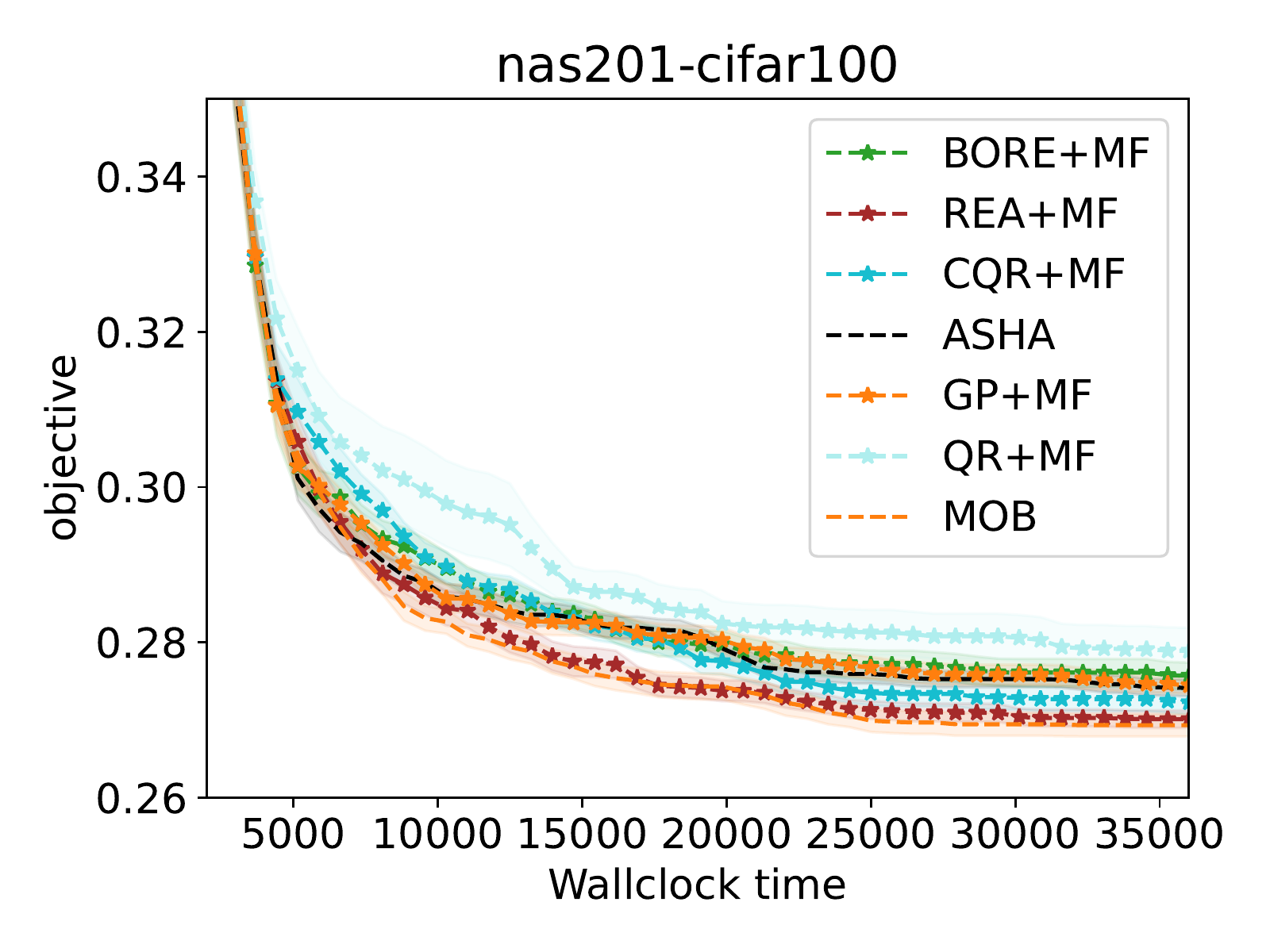}
\includegraphics[width=0.31\textwidth]{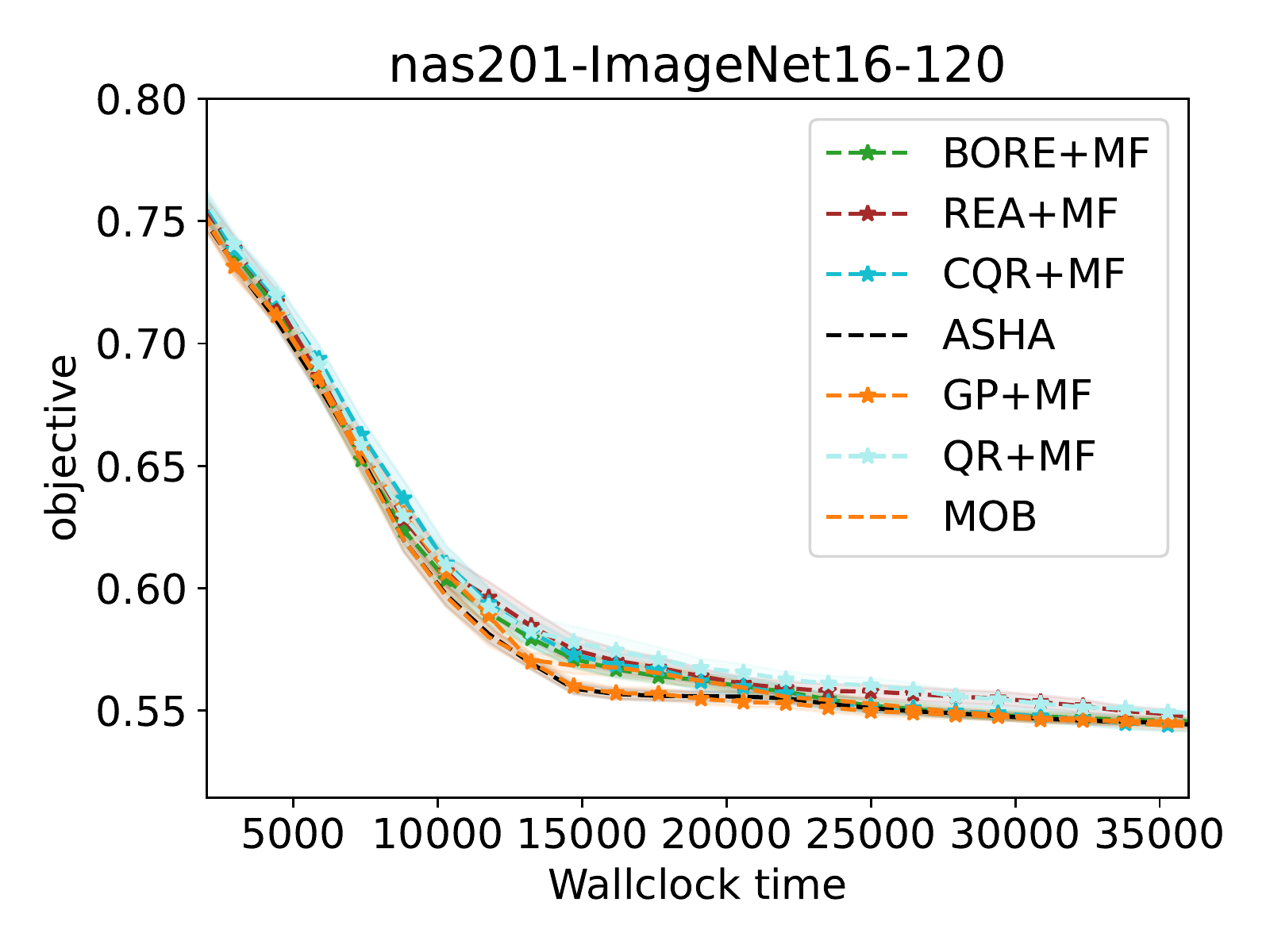} \\
\includegraphics[width=0.31\textwidth]{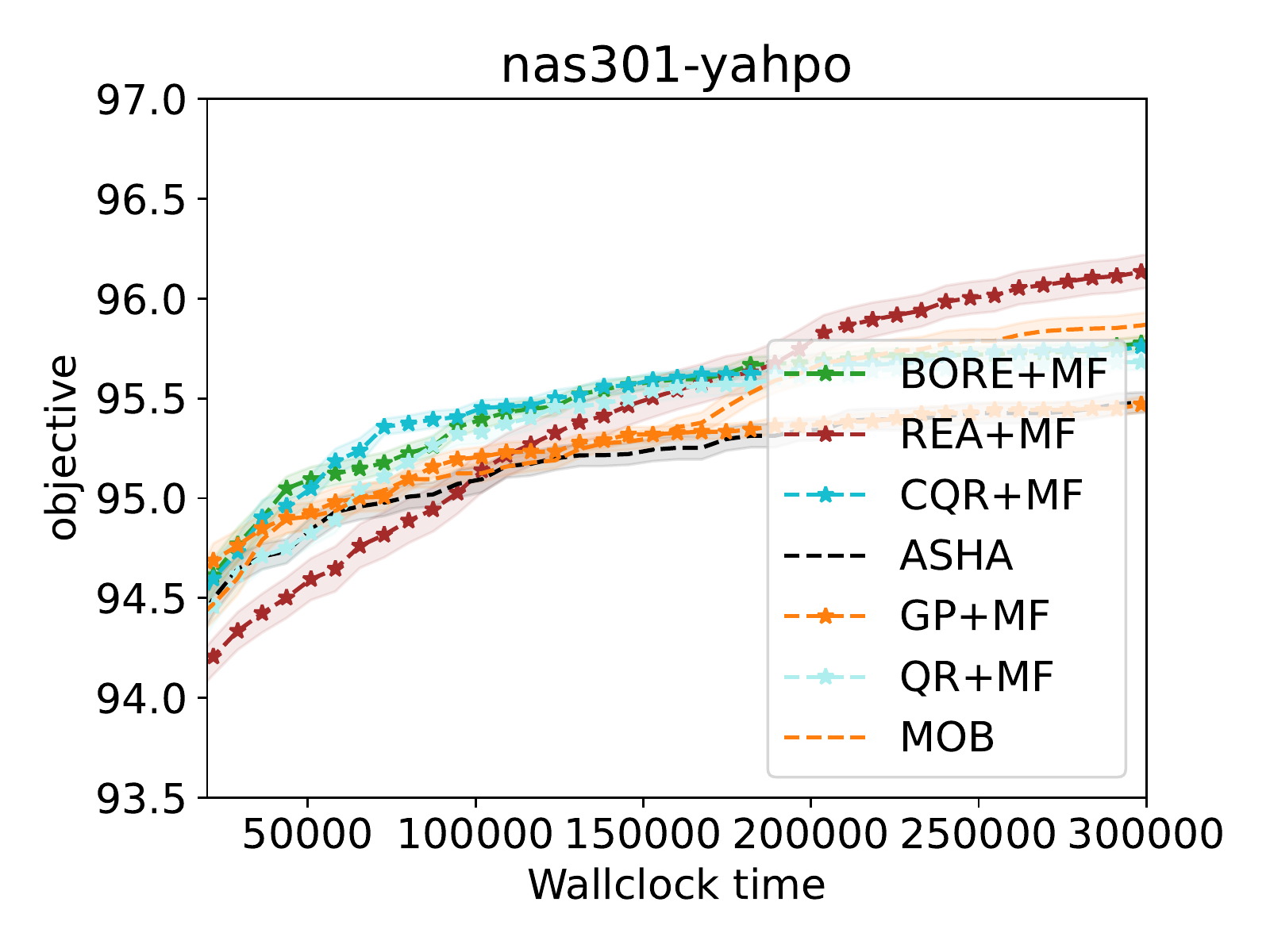}
\caption{Performance of multi-fidelity ablation variants over time on all individual tasks considered. Mean and standard errors are computed over \numseed{} seeds. \label{fig:multi-fidelity-ablation-all-tasks}}
\end{figure*}

\section{Surrogate performance}

\paragraph{RMSE.} 
To compute RMSE for quantile predictions, we first compute the mean by taking the average of predicted quantiles.

\paragraph{Calibration error.} A calibrated estimator of the conditional distribution $\quantilemodel{\alpha}(x)$ should exceed the target $\alpha$ percent of the time asymptotically. 
For instance, we expect the predicted P90-th percentile to exceed the target 90\% of the time for a calibrated estimator. Formally, we expect the following property to hold for a calibrated estimator for all $\alpha\in[0, 1]$:
$$\mathbb{E}_{(\hp, y)\in \observations}[ \bm{1}_{y < \quantilemodel{\alpha}(x)}] = \alpha$$
where the mean is taken over examples on a validation set $(\hp, y)\in \valset$.

The calibration error measures the gap shown in predictions compared to this expected property \cite{Kuleshov2018}. 
Let us denote $P(\alpha) = \mathbb{E}_{(\hp, y)\in \observations}[ \bm{1}_{y < \quantilemodel{\alpha}(x)}]$ the average number of times the prediction $\quantilemodel{\alpha}(x)$ exceeds the target, then the calibration error is defined over a set of quantiles $\alpha_1, \dots, \alpha_k$ as $ \sqrt{\sum_{j=1}^k (P(\alpha_j) - \alpha_j)^2}$. We report this error on 5 equally-spaced quantiles in surrogate experiments. For models predicting quantiles (\QR{} and \CQR{}), we evaluate directly the calibration on the predicted quantiles and for models predicting normal distribution (\GP{}), we compute quantiles predictions using the Gaussian inverse CDF.

\begin{table}
\caption{Calibration error, Pearson correlation and runtime for different surrogates when increasing the number of samples. Results are averaged over \numseed{} different seeds. \label{tab:surr-perf-per-task}}
\footnotesize
\center
\addtolength{\tabcolsep}{-2.5pt}    
\begin{tabular}{ll|rrr|rrr|rrr|rrr}
\toprule
 &  & \multicolumn{3}{c}{RMSE $\downarrow$} & \multicolumn{3}{c}{Calibration error $\downarrow$} & \multicolumn{3}{c}{Runtime $\downarrow$} \\
 & model & GP & QR & CQR & GP & QR & CQR & GP & QR & CQR \\
task & $n$ &  &  &  &  &  &  &  &  &  \\
\midrule
\multirow[c]{4}{*}{airlines} & 16 & 1.44 & 0.89 & 0.88 & 0.17 & 0.13 & 0.16 & 1.92 & 0.95 & 0.87 \\
 & 64 & 1.19 & 0.68 & 0.67 & 0.11 & 0.09 & 0.08 & 3.11 & 1.36 & 1.27 \\
 & 256 & 1.10 & 0.51 & 0.54 & 0.14 & 0.07 & 0.04 & 3.90 & 1.67 & 1.59 \\
 & 1024 & 0.92 & 0.45 & 0.45 & 0.17 & 0.07 & 0.05 & 19.02 & 2.45 & 2.34 \\
\cline{1-11}
\multirow[c]{4}{*}{cifar10} & 16 & 0.79 & 0.74 & 0.81 & 0.01 & 0.14 & 0.13 & 0.54 & 1.20 & 1.14 \\
 & 64 & 0.78 & 0.53 & 0.54 & 0.01 & 0.09 & 0.06 & 0.74 & 1.49 & 1.41 \\
 & 256 & 0.72 & 0.41 & 0.42 & 0.02 & 0.06 & 0.04 & 0.97 & 1.71 & 1.73 \\
 & 1024 & 0.22 & 0.35 & 0.35 & 0.05 & 0.03 & 0.03 & 19.39 & 2.11 & 2.04 \\
\cline{1-11}
\multirow[c]{4}{*}{parkinsons} & 16 & 0.79 & 0.71 & 0.72 & 0.01 & 0.13 & 0.11 & 0.88 & 1.22 & 1.17 \\
 & 64 & 0.80 & 0.51 & 0.53 & 0.01 & 0.11 & 0.09 & 1.28 & 1.60 & 1.60 \\
 & 256 & 0.75 & 0.37 & 0.37 & 0.03 & 0.04 & 0.04 & 1.96 & 1.87 & 1.81 \\
 & 1024 & 0.58 & 0.31 & 0.31 & 0.12 & 0.02 & 0.02 & 21.68 & 2.14 & 2.10 \\
\cline{1-11}
\bottomrule
\end{tabular}

\addtolength{\tabcolsep}{2.5pt}
\end{table}

\end{document}